\RequirePackage{fix-cm}
\documentclass{article}

\usepackage{microtype}
\usepackage{graphicx}
\usepackage{subcaption}
\usepackage{booktabs}
\usepackage[hyperfootnotes=false]{hyperref}

\usepackage[preprint]{icml2026}
\usepackage{placeins}
\hypersetup{pdftitle={Library of Hands: Fix the Hands, Not the World Model}, pdfauthor={Hasaan Ahmad, Qizhen Ying, Yuxuan Sun, Tianjin Huang, Guangming Wang}, pdfsubject={}}

\usepackage{amsmath}
\usepackage{amssymb}
\usepackage{mathtools}
\usepackage{amsthm}
\usepackage[capitalize,noabbrev]{cleveref}
\usepackage{algorithm}
\usepackage{algorithmic}
\usepackage{xcolor}
\usepackage{multirow}
\usepackage{longtable}
\usepackage{enumitem}
\setlist{leftmargin=*, itemsep=1pt, topsep=2pt, parsep=0pt}
\usepackage{pgfplots}
\pgfplotsset{compat=1.17}
\usetikzlibrary{arrows.meta,positioning,calc,fit,backgrounds,shapes.geometric,shadows,decorations.pathreplacing}
\usepackage{pifont}
\usepackage{colortbl}
\usepackage[most]{tcolorbox}

\theoremstyle{plain}
\newtheorem{theorem}{Theorem}[section]
\newtheorem{proposition}[theorem]{Proposition}
\newtheorem{lemma}[theorem]{Lemma}
\newtheorem{corollary}[theorem]{Corollary}
\theoremstyle{definition}

\theoremstyle{remark}

\newcommand{\method}{\texttt{LoH}}
\newcommand{\cmark}{\textcolor{cThree}{\ding{51}}}
\newcommand{\xmark}{\textcolor{cTwo}{\ding{55}}}
\newcommand{\pmark}{\textcolor{cGrey}{$\circ$}}
\newcommand{\obs}[1]{\ding{\numexpr181+#1\relax}}
\newtcolorbox{remarkbox}{enhanced, colback=cOurs!7, colframe=cOurs!7, boxrule=0pt, arc=1.5pt,
  left=4pt, right=4pt, top=3pt, bottom=3pt, fontupper=\small}
\newcommand{\best}[1]{\textbf{#1}}
\newcommand{\second}[1]{\underline{#1}}
\newcommand{\oursrow}{\rowcolor{cOurs!9}}
\newcommand{\degr}{\ensuremath{^\circ}}
\newcommand{\R}{\mathbb{R}}
\newcommand{\E}{\mathbb{E}}
\DeclareMathOperator*{\argmin}{arg\,min}

\definecolor{cOurs}{HTML}{5B5BD6}
\definecolor{cOursLight}{HTML}{9A9CF2}
\definecolor{cOursDeep}{HTML}{3E3BB0}
\definecolor{cTwo}{HTML}{F2625A}
\colorlet{cHand}{cTwo}
\definecolor{cThree}{HTML}{12A386}
\definecolor{cGrey}{HTML}{9B9DB0}
\definecolor{cBar}{HTML}{DDA11C}
\definecolor{lohShadeA}{HTML}{ECEBF6}
\definecolor{lohShadeD}{HTML}{F5F5FA}

\newcommand{\plotfont}{\fontsize{7.5}{9}\selectfont}
\pgfplotsset{
  lohplot/.style={
    scale only axis,
    axis lines*=left,
    axis line style={-, black!80, line width=0.4pt},
    tick style={black!80, line width=0.4pt},
    tick align=outside,
    major tick length=2pt,
    tick label style={font=\plotfont, text=black!85},
    /pgf/number format/assume math mode=true,
    label style={font=\plotfont},
    legend style={font=\plotfont, draw=none, fill=none, inner sep=1pt, row sep=-1pt},
    legend cell align=left,
    grid style={black!10, line width=0.35pt},
    clip=false,
  },
}

\icmltitlerunning{Library of Hands: Fix the Hands, Not the World Model}

\begin{document}

\twocolumn[
  \icmltitle{Library of Hands: Fix the Hands, Not the World Model}

  \begin{icmlauthorlist}
    \icmlauthor{Hasaan Ahmad}{uwe}
    \icmlauthor{Qizhen Ying}{oxf}
    \icmlauthor{Yuxuan Sun}{imp}
    \icmlauthor{Tianjin Huang}{exe}
    \icmlauthor{Guangming Wang}{cam}
  \end{icmlauthorlist}

  \icmlaffiliation{uwe}{University of the West of England, Bristol, UK}
  \icmlaffiliation{oxf}{University of Oxford, Oxford, UK}
  \icmlaffiliation{imp}{Imperial College London, London, UK}
  \icmlaffiliation{exe}{University of Exeter, Exeter, UK}
  \icmlaffiliation{cam}{University of Cambridge, Cambridge, UK}
  \icmlcorrespondingauthor{Hasaan Ahmad}{hasaan2.ahmad@live.uwe.ac.uk}

  \icmlkeywords{world models, model-based planning, test-time adaptation, system identification, continual learning, robot learning}

  \vskip 0.3in
]

\printAffiliationsAndNotice{}

\begin{abstract}
A world model predicts how a scene changes after an action, and a planner uses it to search for actions that reach a goal. If the camera moves or the controller changes, the same command moves the robot differently: when its commands are turned by 90\degr{}, a frozen planner with the LeWM world model that succeeds in 77.5\% of episodes succeeds in only 7.8\%. Current fixes retrain the model, need training data, or forget a set-up when it returns. We propose Library of Hands (\method{}), a training-free method that keeps the world model frozen and corrects only the map from commands to applied actions, which we call the \emph{hand}. \method{} reads the hand through the frozen model, with no extra sensor: it finds the action that best explains the change between two camera images (a \emph{reading}). It stores every hand it identifies, so a returning set-up is recognised rather than re-learnt. In closed-loop tests, \method{} more than doubles success on three simulated tasks, from 28.6\% to 62.3\%, and loses only 0.6 percentage points when nothing changes. It also works, unchanged, with two other world models and on two of three models pretrained on real robot data, and its gain grows with the quality of its readings.
\end{abstract}

\section{Introduction}
\label{sec:intro}

\textbf{A world model can be right about the world and still plan with the wrong actions.} Latent world models, which predict in a learned compact representation (a latent space) instead of in pixels, now plan robot actions directly from camera images \citep{lewm,dinowm,vjepa2,jepawms}. A learned predictor models how the scene evolves in response to an action, and a planner searches for the action sequence whose predicted outcome reaches a goal \citep{cem}. The predictor implicitly assumes that every action follows the convention of its training data. This assumption fails when the camera is moved or the controller is replaced: the same command then moves the robot in a different direction, by a different magnitude (a changed gain), or only after a delay. The situation resembles a joystick mounted with a 90\degr{} offset, where a forward command moves the robot to the left. Even when the predictor remains an accurate model of the world, every plan is then expressed in the wrong action convention, and nothing in the model reveals the cause. With the LeWM world model \citep{lewm}, a frozen planner succeeds in 77.5\% of episodes with its original convention, and in 7.8\% when its commands are turned by 90\degr{}.

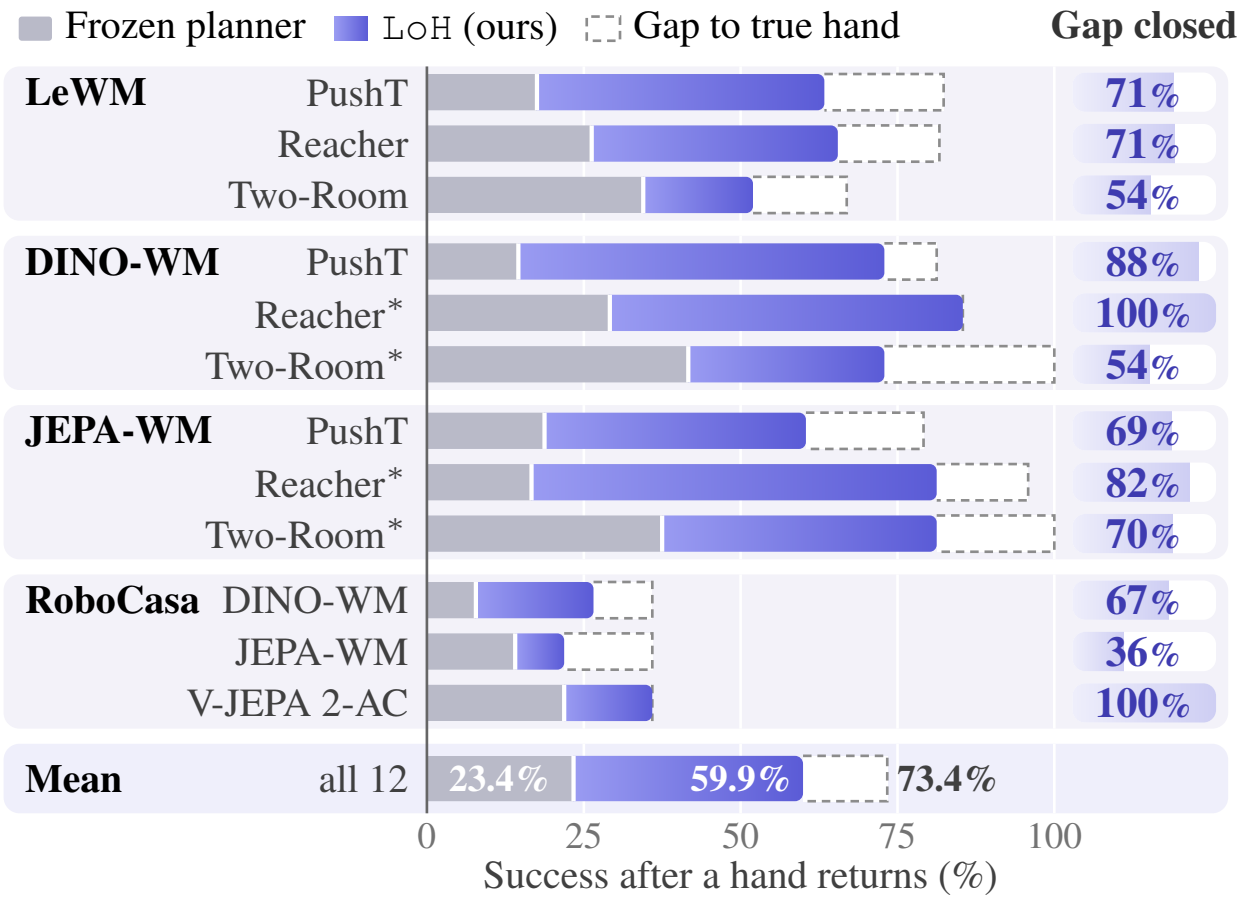
\begin{figure}[t]
  \centering
\begin{tikzpicture}[x=1cm, y=1cm, font=\plotfont]
  \fill[lohShadeA!65, rounded corners=3pt] (-0.72,-0.85) rectangle (7.38,0.17);
  \draw[white, line width=0.5pt] (3.118,-0.85) -- (3.118,0.17);
  \draw[white, line width=0.5pt] (4.155,-0.85) -- (4.155,0.17);
  \draw[white, line width=0.5pt] (5.193,-0.85) -- (5.193,0.17);
  \draw[white, line width=0.5pt] (6.23,-0.85) -- (6.23,0.17);
  \node[anchor=west, inner sep=1pt, font=\plotfont\bfseries] at (-0.62,0) {LeWM};
  \fill[lohShadeA!65, rounded corners=3pt] (-0.72,-1.97) rectangle (7.38,-0.95);
  \draw[white, line width=0.5pt] (3.118,-1.97) -- (3.118,-0.95);
  \draw[white, line width=0.5pt] (4.155,-1.97) -- (4.155,-0.95);
  \draw[white, line width=0.5pt] (5.193,-1.97) -- (5.193,-0.95);
  \draw[white, line width=0.5pt] (6.23,-1.97) -- (6.23,-0.95);
  \node[anchor=west, inner sep=1pt, font=\plotfont\bfseries] at (-0.62,-1.12) {DINO-WM};
  \fill[lohShadeA!65, rounded corners=3pt] (-0.72,-3.09) rectangle (7.38,-2.07);
  \draw[white, line width=0.5pt] (3.118,-3.09) -- (3.118,-2.07);
  \draw[white, line width=0.5pt] (4.155,-3.09) -- (4.155,-2.07);
  \draw[white, line width=0.5pt] (5.193,-3.09) -- (5.193,-2.07);
  \draw[white, line width=0.5pt] (6.23,-3.09) -- (6.23,-2.07);
  \node[anchor=west, inner sep=1pt, font=\plotfont\bfseries] at (-0.62,-2.24) {JEPA-WM};
  \fill[lohShadeA!65, rounded corners=3pt] (-0.72,-4.21) rectangle (7.38,-3.19);
  \draw[white, line width=0.5pt] (3.118,-4.21) -- (3.118,-3.19);
  \draw[white, line width=0.5pt] (4.155,-4.21) -- (4.155,-3.19);
  \draw[white, line width=0.5pt] (5.193,-4.21) -- (5.193,-3.19);
  \draw[white, line width=0.5pt] (6.23,-4.21) -- (6.23,-3.19);
  \node[anchor=west, inner sep=1pt, font=\plotfont\bfseries] at (-0.62,-3.36) {RoboCasa};
  \node[anchor=north, inner sep=1.5pt, font=\scriptsize, text=black!55] at (2.08,-4.77) {0};
  \node[anchor=north, inner sep=1.5pt, font=\scriptsize, text=black!55] at (3.118,-4.77) {25};
  \node[anchor=north, inner sep=1.5pt, font=\scriptsize, text=black!55] at (4.155,-4.77) {50};
  \node[anchor=north, inner sep=1.5pt, font=\scriptsize, text=black!55] at (5.193,-4.77) {75};
  \node[anchor=north, inner sep=1.5pt, font=\scriptsize, text=black!55] at (6.23,-4.77) {100};
  \node[anchor=north, inner sep=1pt, text=black!75] at (4.155,-5.04) {Success after a hand returns (\%)};
  \fill[white] (4.709,-0.12) rectangle (5.498,0.12); \draw[black!45, line width=0.5pt, dash pattern=on 1.6pt off 1.1pt] (4.709,0.12) -- (5.498,0.12) -- (5.498,-0.12) -- (4.709,-0.12);
  \begin{scope}\clip (2.08,-0.12) -- (2.08,0.12) [rounded corners=1.3pt] -- (4.709,0.12) -- (4.709,-0.12) [sharp corners] -- cycle; \shade[left color=cOursLight, right color=cOurs] (2.808,-0.12) rectangle (4.709,0.12); \fill[cGrey!70] (2.08,-0.12) rectangle (2.808,0.12); \end{scope}
  \draw[white, line width=0.8pt] (2.808,-0.12) -- (2.808,0.12);
  \begin{scope}\clip[rounded corners=2.4pt] (6.35,-0.13) rectangle (7.3,0.13); \fill[white] (6.35,-0.13) rectangle (7.3,0.13); \shade[left color=cOurs!10, right color=cOurs!30] (6.35,-0.13) rectangle (7.021,0.13); \end{scope}
  \node[inner sep=0pt, font=\fontsize{8}{10}\selectfont\bfseries, text=cOursDeep] at (6.825,0) {71{\fontsize{6.5}{8}\selectfont\%}};
  \node[anchor=east, inner sep=1pt, text=black!75] at (2,0) {PushT};
  \fill[white] (4.798,-0.46) rectangle (5.47,-0.22); \draw[black!45, line width=0.5pt, dash pattern=on 1.6pt off 1.1pt] (4.798,-0.22) -- (5.47,-0.22) -- (5.47,-0.46) -- (4.798,-0.46);
  \begin{scope}\clip (2.08,-0.46) -- (2.08,-0.22) [rounded corners=1.3pt] -- (4.798,-0.22) -- (4.798,-0.46) [sharp corners] -- cycle; \shade[left color=cOursLight, right color=cOurs] (3.170,-0.46) rectangle (4.798,-0.22); \fill[cGrey!70] (2.08,-0.46) rectangle (3.170,-0.22); \end{scope}
  \draw[white, line width=0.8pt] (3.170,-0.46) -- (3.170,-0.22);
  \begin{scope}\clip[rounded corners=2.4pt] (6.35,-0.47) rectangle (7.3,-0.21); \fill[white] (6.35,-0.47) rectangle (7.3,-0.21); \shade[left color=cOurs!10, right color=cOurs!30] (6.35,-0.47) rectangle (7.023,-0.21); \end{scope}
  \node[inner sep=0pt, font=\fontsize{8}{10}\selectfont\bfseries, text=cOursDeep] at (6.825,-0.34) {71{\fontsize{6.5}{8}\selectfont\%}};
  \node[anchor=east, inner sep=1pt, text=black!75] at (2,-0.34) {Reacher};
  \fill[white] (4.237,-0.8) rectangle (4.857,-0.56); \draw[black!45, line width=0.5pt, dash pattern=on 1.6pt off 1.1pt] (4.237,-0.56) -- (4.857,-0.56) -- (4.857,-0.8) -- (4.237,-0.8);
  \begin{scope}\clip (2.08,-0.8) -- (2.08,-0.56) [rounded corners=1.3pt] -- (4.237,-0.56) -- (4.237,-0.8) [sharp corners] -- cycle; \shade[left color=cOursLight, right color=cOurs] (3.511,-0.8) rectangle (4.237,-0.56); \fill[cGrey!70] (2.08,-0.8) rectangle (3.511,-0.56); \end{scope}
  \draw[white, line width=0.8pt] (3.511,-0.8) -- (3.511,-0.56);
  \begin{scope}\clip[rounded corners=2.4pt] (6.35,-0.81) rectangle (7.3,-0.55); \fill[white] (6.35,-0.81) rectangle (7.3,-0.55); \shade[left color=cOurs!10, right color=cOurs!30] (6.35,-0.81) rectangle (6.863,-0.55); \end{scope}
  \node[inner sep=0pt, font=\fontsize{8}{10}\selectfont\bfseries, text=cOursDeep] at (6.825,-0.68) {54{\fontsize{6.5}{8}\selectfont\%}};
  \node[anchor=east, inner sep=1pt, text=black!75] at (2,-0.68) {Two-Room};
  \fill[white] (5.106,-1.24) rectangle (5.452,-1); \draw[black!45, line width=0.5pt, dash pattern=on 1.6pt off 1.1pt] (5.106,-1) -- (5.452,-1) -- (5.452,-1.24) -- (5.106,-1.24);
  \begin{scope}\clip (2.08,-1.24) -- (2.08,-1) [rounded corners=1.3pt] -- (5.106,-1) -- (5.106,-1.24) [sharp corners] -- cycle; \shade[left color=cOursLight, right color=cOurs] (2.685,-1.24) rectangle (5.106,-1); \fill[cGrey!70] (2.08,-1.24) rectangle (2.685,-1); \end{scope}
  \draw[white, line width=0.8pt] (2.685,-1.24) -- (2.685,-1);
  \begin{scope}\clip[rounded corners=2.4pt] (6.35,-1.25) rectangle (7.3,-0.99); \fill[white] (6.35,-1.25) rectangle (7.3,-0.99); \shade[left color=cOurs!10, right color=cOurs!30] (6.35,-1.25) rectangle (7.181,-0.99); \end{scope}
  \node[inner sep=0pt, font=\fontsize{8}{10}\selectfont\bfseries, text=cOursDeep] at (6.825,-1.12) {88{\fontsize{6.5}{8}\selectfont\%}};
  \node[anchor=east, inner sep=1pt, text=black!75] at (2,-1.12) {PushT};
  \fill[white] (5.625,-1.58) rectangle (5.625,-1.34); \draw[black!45, line width=0.5pt, dash pattern=on 1.6pt off 1.1pt] (5.625,-1.34) -- (5.625,-1.34) -- (5.625,-1.58) -- (5.625,-1.58);
  \begin{scope}\clip (2.08,-1.58) -- (2.08,-1.34) [rounded corners=1.3pt] -- (5.625,-1.34) -- (5.625,-1.58) [sharp corners] -- cycle; \shade[left color=cOursLight, right color=cOurs] (3.29,-1.58) rectangle (5.625,-1.34); \fill[cGrey!70] (2.08,-1.58) rectangle (3.29,-1.34); \end{scope}
  \draw[white, line width=0.8pt] (3.29,-1.58) -- (3.29,-1.34);
  \begin{scope}\clip[rounded corners=2.4pt] (6.35,-1.59) rectangle (7.3,-1.33); \fill[white] (6.35,-1.59) rectangle (7.3,-1.33); \shade[left color=cOurs!10, right color=cOurs!30] (6.35,-1.59) rectangle (7.3,-1.33); \end{scope}
  \node[inner sep=0pt, font=\fontsize{8}{10}\selectfont\bfseries, text=cOursDeep] at (6.825,-1.46) {100{\fontsize{6.5}{8}\selectfont\%}};
  \node[anchor=east, inner sep=1pt, text=black!75] at (2,-1.46) {Reacher$^{\ast}$};
  \fill[white] (5.106,-1.92) rectangle (6.23,-1.68); \draw[black!45, line width=0.5pt, dash pattern=on 1.6pt off 1.1pt] (5.106,-1.68) -- (6.23,-1.68) -- (6.23,-1.92) -- (5.106,-1.92);
  \begin{scope}\clip (2.08,-1.92) -- (2.08,-1.68) [rounded corners=1.3pt] -- (5.106,-1.68) -- (5.106,-1.92) [sharp corners] -- cycle; \shade[left color=cOursLight, right color=cOurs] (3.809,-1.92) rectangle (5.106,-1.68); \fill[cGrey!70] (2.08,-1.92) rectangle (3.809,-1.68); \end{scope}
  \draw[white, line width=0.8pt] (3.809,-1.92) -- (3.809,-1.68);
  \begin{scope}\clip[rounded corners=2.4pt] (6.35,-1.93) rectangle (7.3,-1.67); \fill[white] (6.35,-1.93) rectangle (7.3,-1.67); \shade[left color=cOurs!10, right color=cOurs!30] (6.35,-1.93) rectangle (6.859,-1.67); \end{scope}
  \node[inner sep=0pt, font=\fontsize{8}{10}\selectfont\bfseries, text=cOursDeep] at (6.825,-1.8) {54{\fontsize{6.5}{8}\selectfont\%}};
  \node[anchor=east, inner sep=1pt, text=black!75] at (2,-1.8) {Two-Room$^{\ast}$};
  \fill[white] (4.587,-2.36) rectangle (5.365,-2.12); \draw[black!45, line width=0.5pt, dash pattern=on 1.6pt off 1.1pt] (4.587,-2.12) -- (5.365,-2.12) -- (5.365,-2.36) -- (4.587,-2.36);
  \begin{scope}\clip (2.08,-2.36) -- (2.08,-2.12) [rounded corners=1.3pt] -- (4.587,-2.12) -- (4.587,-2.36) [sharp corners] -- cycle; \shade[left color=cOursLight, right color=cOurs] (2.858,-2.36) rectangle (4.587,-2.12); \fill[cGrey!70] (2.08,-2.36) rectangle (2.858,-2.12); \end{scope}
  \draw[white, line width=0.8pt] (2.858,-2.36) -- (2.858,-2.12);
  \begin{scope}\clip[rounded corners=2.4pt] (6.35,-2.37) rectangle (7.3,-2.11); \fill[white] (6.35,-2.37) rectangle (7.3,-2.11); \shade[left color=cOurs!10, right color=cOurs!30] (6.35,-2.37) rectangle (7.005,-2.11); \end{scope}
  \node[inner sep=0pt, font=\fontsize{8}{10}\selectfont\bfseries, text=cOursDeep] at (6.825,-2.24) {69{\fontsize{6.5}{8}\selectfont\%}};
  \node[anchor=east, inner sep=1pt, text=black!75] at (2,-2.24) {PushT};
  \fill[white] (5.452,-2.7) rectangle (6.057,-2.46); \draw[black!45, line width=0.5pt, dash pattern=on 1.6pt off 1.1pt] (5.452,-2.46) -- (6.057,-2.46) -- (6.057,-2.7) -- (5.452,-2.7);
  \begin{scope}\clip (2.08,-2.7) -- (2.08,-2.46) [rounded corners=1.3pt] -- (5.452,-2.46) -- (5.452,-2.7) [sharp corners] -- cycle; \shade[left color=cOursLight, right color=cOurs] (2.772,-2.7) rectangle (5.452,-2.46); \fill[cGrey!70] (2.08,-2.7) rectangle (2.772,-2.46); \end{scope}
  \draw[white, line width=0.8pt] (2.772,-2.7) -- (2.772,-2.46);
  \begin{scope}\clip[rounded corners=2.4pt] (6.35,-2.71) rectangle (7.3,-2.45); \fill[white] (6.35,-2.71) rectangle (7.3,-2.45); \shade[left color=cOurs!10, right color=cOurs!30] (6.35,-2.71) rectangle (7.125,-2.45); \end{scope}
  \node[inner sep=0pt, font=\fontsize{8}{10}\selectfont\bfseries, text=cOursDeep] at (6.825,-2.58) {82{\fontsize{6.5}{8}\selectfont\%}};
  \node[anchor=east, inner sep=1pt, text=black!75] at (2,-2.58) {Reacher$^{\ast}$};
  \fill[white] (5.452,-3.04) rectangle (6.23,-2.8); \draw[black!45, line width=0.5pt, dash pattern=on 1.6pt off 1.1pt] (5.452,-2.8) -- (6.23,-2.8) -- (6.23,-3.04) -- (5.452,-3.04);
  \begin{scope}\clip (2.08,-3.04) -- (2.08,-2.8) [rounded corners=1.3pt] -- (5.452,-2.8) -- (5.452,-3.04) [sharp corners] -- cycle; \shade[left color=cOursLight, right color=cOurs] (3.636,-3.04) rectangle (5.452,-2.8); \fill[cGrey!70] (2.08,-3.04) rectangle (3.636,-2.8); \end{scope}
  \draw[white, line width=0.8pt] (3.636,-3.04) -- (3.636,-2.8);
  \begin{scope}\clip[rounded corners=2.4pt] (6.35,-3.05) rectangle (7.3,-2.79); \fill[white] (6.35,-3.05) rectangle (7.3,-2.79); \shade[left color=cOurs!10, right color=cOurs!30] (6.35,-3.05) rectangle (7.015,-2.79); \end{scope}
  \node[inner sep=0pt, font=\fontsize{8}{10}\selectfont\bfseries, text=cOursDeep] at (6.825,-2.92) {70{\fontsize{6.5}{8}\selectfont\%}};
  \node[anchor=east, inner sep=1pt, text=black!75] at (2,-2.92) {Two-Room$^{\ast}$};
  \fill[white] (3.182,-3.48) rectangle (3.571,-3.24); \draw[black!45, line width=0.5pt, dash pattern=on 1.6pt off 1.1pt] (3.182,-3.24) -- (3.571,-3.24) -- (3.571,-3.48) -- (3.182,-3.48);
  \begin{scope}\clip (2.08,-3.48) -- (2.08,-3.24) [rounded corners=1.3pt] -- (3.182,-3.24) -- (3.182,-3.48) [sharp corners] -- cycle; \shade[left color=cOursLight, right color=cOurs] (2.404,-3.48) rectangle (3.182,-3.24); \fill[cGrey!70] (2.08,-3.48) rectangle (2.404,-3.24); \end{scope}
  \draw[white, line width=0.8pt] (2.404,-3.48) -- (2.404,-3.24);
  \begin{scope}\clip[rounded corners=2.4pt] (6.35,-3.49) rectangle (7.3,-3.23); \fill[white] (6.35,-3.49) rectangle (7.3,-3.23); \shade[left color=cOurs!10, right color=cOurs!30] (6.35,-3.49) rectangle (6.983,-3.23); \end{scope}
  \node[inner sep=0pt, font=\fontsize{8}{10}\selectfont\bfseries, text=cOursDeep] at (6.825,-3.36) {67{\fontsize{6.5}{8}\selectfont\%}};
  \node[anchor=east, inner sep=1pt, text=black!75] at (2,-3.36) {DINO-WM};
  \fill[white] (2.988,-3.82) rectangle (3.571,-3.58); \draw[black!45, line width=0.5pt, dash pattern=on 1.6pt off 1.1pt] (2.988,-3.58) -- (3.571,-3.58) -- (3.571,-3.82) -- (2.988,-3.82);
  \begin{scope}\clip (2.08,-3.82) -- (2.08,-3.58) [rounded corners=1.3pt] -- (2.988,-3.58) -- (2.988,-3.82) [sharp corners] -- cycle; \shade[left color=cOursLight, right color=cOurs] (2.664,-3.82) rectangle (2.988,-3.58); \fill[cGrey!70] (2.08,-3.82) rectangle (2.664,-3.58); \end{scope}
  \draw[white, line width=0.8pt] (2.664,-3.82) -- (2.664,-3.58);
  \begin{scope}\clip[rounded corners=2.4pt] (6.35,-3.83) rectangle (7.3,-3.57); \fill[white] (6.35,-3.83) rectangle (7.3,-3.57); \shade[left color=cOurs!10, right color=cOurs!30] (6.35,-3.83) rectangle (6.689,-3.57); \end{scope}
  \node[inner sep=0pt, font=\fontsize{8}{10}\selectfont\bfseries, text=cOursDeep] at (6.825,-3.7) {36{\fontsize{6.5}{8}\selectfont\%}};
  \node[anchor=east, inner sep=1pt, text=black!75] at (2,-3.7) {JEPA-WM};
  \fill[white] (3.571,-4.16) rectangle (3.571,-3.92); \draw[black!45, line width=0.5pt, dash pattern=on 1.6pt off 1.1pt] (3.571,-3.92) -- (3.571,-3.92) -- (3.571,-4.16) -- (3.571,-4.16);
  \begin{scope}\clip (2.08,-4.16) -- (2.08,-3.92) [rounded corners=1.3pt] -- (3.571,-3.92) -- (3.571,-4.16) [sharp corners] -- cycle; \shade[left color=cOursLight, right color=cOurs] (2.988,-4.16) rectangle (3.571,-3.92); \fill[cGrey!70] (2.08,-4.16) rectangle (2.988,-3.92); \end{scope}
  \draw[white, line width=0.8pt] (2.988,-4.16) -- (2.988,-3.92);
  \begin{scope}\clip[rounded corners=2.4pt] (6.35,-4.17) rectangle (7.3,-3.91); \fill[white] (6.35,-4.17) rectangle (7.3,-3.91); \shade[left color=cOurs!10, right color=cOurs!30] (6.35,-4.17) rectangle (7.3,-3.91); \end{scope}
  \node[inner sep=0pt, font=\fontsize{8}{10}\selectfont\bfseries, text=cOursDeep] at (6.825,-4.04) {100{\fontsize{6.5}{8}\selectfont\%}};
  \node[anchor=east, inner sep=1pt, text=black!75] at (2,-4.04) {V-JEPA 2-AC};
  \fill[cOurs!10, rounded corners=3pt] (-0.72,-4.77) rectangle (7.38,-4.31);
  \foreach \lohgx in {3.118, 4.155, 5.193, 6.23} \draw[white, line width=0.5pt] (\lohgx,-4.77) -- (\lohgx,-4.31);
  \node[anchor=west, inner sep=1pt, font=\plotfont\bfseries] at (-0.62,-4.54) {Mean};
  \node[anchor=east, inner sep=1pt, text=black!75] at (2,-4.54) {all 12};
  \fill[white] (4.568,-4.69) rectangle (5.125,-4.39); \draw[black!45, line width=0.5pt, dash pattern=on 1.6pt off 1.1pt] (4.568,-4.39) -- (5.125,-4.39) -- (5.125,-4.69) -- (4.568,-4.69);
  \begin{scope}\clip (2.08,-4.69) -- (2.08,-4.39) [rounded corners=1.5pt] -- (4.568,-4.39) -- (4.568,-4.69) [sharp corners] -- cycle; \shade[left color=cOursLight, right color=cOurs] (3.050,-4.69) rectangle (4.568,-4.39); \fill[cGrey!70] (2.08,-4.69) rectangle (3.050,-4.39); \end{scope}
  \draw[white, line width=0.8pt] (3.050,-4.69) -- (3.050,-4.39);
  \node[inner sep=0pt, font=\fontsize{7}{8}\selectfont\bfseries, text=white] at (2.565,-4.54) {23.4\%};
  \node[inner sep=0pt, font=\fontsize{7}{8}\selectfont\bfseries, text=white, anchor=east] at (4.508,-4.54) {59.9\%};
  \node[inner sep=0pt, font=\fontsize{7}{8}\selectfont\bfseries, text=black!75, anchor=west] at (5.185,-4.54) {73.4\%};
  \draw[black!60, line width=0.5pt] (2.08,-4.77) -- (2.08,0.17);
  \node[inner sep=1.5pt, font=\plotfont\bfseries, text=black!80] at (6.825,0.42) {Gap closed};
  \begin{scope}[shift={(-0.62,0.42)}]
    \fill[cGrey!70, rounded corners=0.6pt] (0,-0.08) rectangle (0.22,0.08);
    \node[anchor=west, inner sep=1.5pt] (teaserKeyA) at (0.25,0) {Frozen planner};
    \shade[left color=cOursLight, right color=cOurs] ([shift={(0.13,-0.08)}]teaserKeyA.east) rectangle ++(0.22,0.16);
    \node[anchor=west, inner sep=1.5pt] (teaserKeyB) at ([xshift=0.38cm]teaserKeyA.east) {\method{} (ours)};
    \draw[black!45, line width=0.5pt, dash pattern=on 1.6pt off 1.1pt, fill=white] ([shift={(0.13,-0.08)}]teaserKeyB.east) rectangle ++(0.22,0.16);
    \node[anchor=west, inner sep=1.5pt] at ([xshift=0.38cm]teaserKeyB.east) {Gap to true hand};
  \end{scope}
\end{tikzpicture}%
  \caption{\textbf{One method closes most of the gap, in simulation and on real-data models.} Success (\%) in the first two episodes after a known hand returns; bottom row: mean over the 12 cells. Right: the share of the gap to a planner told the true hand that \method{} closes. LeWM: S1 and S2, on the 12 seeds on which all three planners ran; DINO-WM and JEPA-WM: S1, 6 seeds; RoboCasa: S1, 8 seeds (\cref{tab:robocasa-main}). $^{\ast}$Trained by us on the data of LeWM.}
  \label{fig:teaser}
\end{figure}

\textbf{This happens in real deployments.} One model, trained once, is used with many cameras, operators and controllers. \mbox{V-JEPA~2-AC}, the action-conditioned version of the \mbox{V-JEPA~2} video model (JEPA: joint-embedding predictive architecture), reads its action axes from the camera view, so its authors positioned the camera manually \citep[Sec.~4.3]{vjepa2}. The JEPA-WM models \citep{jepawms} needed a tuned camera and rescaled actions to move from DROID, a large data set of real robot manipulation \citep{droid}, to RoboCasa, a simulated kitchen \citep{robocasa}. When we move the camera of \mbox{V-JEPA~2-AC} in RoboCasa, the frame in which it reads the commands turns with the camera, as an offline analysis of 261 views of real robot data suggested (\cref{sec:robocasa}). We call the map from commands to applied actions the \emph{hand}: not a gripper, but the whole interface between the planner and the motors. In control terms, it is an unknown input map. A new operator, controller gain or delay changes the hand, not the world. A moved camera also changes the images; we treat it as a hand when the model still predicts well from the new view, so that the main mismatch is the frame of the actions (\cref{sec:setting}).

\textbf{Current fixes change the model, need training, or forget} (\cref{tab:paradigms}). Test-time training changes the weights during deployment \citep{jepattt,adajepa}, so a model adapted to one hand is miscalibrated after the next change. Methods trained to recognise the set-up infer it from recent observations \citep{syncworld} or assign it to one of a few fixed classes \citep{shiftwm}, but only within their training distribution. A calibration phase \citep[App.~B.4]{vjepa2}, or a single retained correction, must restart at every change. None of them exploits a simple property of real laboratories: a laboratory has only a few set-ups, and these recur.

\begin{table}[t]
  \centering
  \caption{\textbf{Only \method{} has all six properties.} (1) keeps the weights frozen, (2) needs no training, (3) estimates a continuous map (any linear map, not a few fixed classes), (4) recalls returning hands, (5) weighs each reading by its reliability, (6) works with the released model weights (checkpoints); \pmark{}: partly.}
  \label{tab:paradigms}
  \footnotesize
  \setlength{\tabcolsep}{3pt}
  \newcommand{\lohex}[1]{\,{\scriptsize\color{black!50}#1}}
  \begin{tabular}{@{}l*{6}{c}@{}}
    \toprule
    Approach & (1) & (2) & (3) & (4) & (5) & (6) \\
    \midrule
    Test-time training\lohex{JEPA-TTT, AdaJEPA} & \xmark & \pmark & \pmark & \xmark & \xmark & \cmark \\
    Trained in-context reader\lohex{SyncWorld} & \cmark & \xmark & \pmark & \xmark & \xmark & \xmark \\
    Calibration per episode\lohex{V-JEPA 2} & \cmark & \cmark & \cmark & \xmark & \xmark & \cmark \\
    Coarse shift classifier\lohex{ShiftWM-Bench} & \cmark & \xmark & \xmark & \xmark & \xmark & \cmark \\
    One kept correction\lohex{one memory} & \cmark & \cmark & \cmark & \pmark & \xmark & \cmark \\
    \oursrow \method{} (ours) & \cmark & \cmark & \cmark & \cmark & \cmark & \cmark \\
    \bottomrule
  \end{tabular}
\end{table}

\textbf{Our idea: fix the hands, not the world model.} \method{} (Library of Hands) is a small module that needs no training and keeps the world model frozen (\cref{fig:overview}). It works in two steps. \emph{Step 1: read the hand through the frozen model.} A model that is right about the world can tell which action moved the world. So we ask the frozen predictor which action best explains the change between two successive camera images, and we compare this action, a \emph{reading} of the applied action, with the command. The world model itself is the sensor: \method{} uses only the inputs of the model (the camera image, and the joint state for the models that take it), and no extra sensor. The predictor also tells us how far to trust each reading. Bayesian model comparison then chooses the map by its exact evidence (the probability of the readings under each candidate map), with no complexity penalty to tune. \emph{Step 2: remember the hand.} A library stores every hand that \method{} identifies. When a set-up comes back, \method{} only has to recognise its hand, which takes fewer readings than to identify it again.

We test \method{} in closed loop (the robot executes its own plans), with hands that change and come back across episodes. Our contributions are:
\begin{itemize}[leftmargin=1.1em]
  \item[$\star$] \textbf{Library of Hands (\method{}).} A method that reads the applied action through the frozen predictor, weighs each reading by its reliability, chooses a map by exact Bayesian evidence, and stores every identified hand, so that a returning hand is recalled rather than identified again (\cref{sec:method}).
  \item[$\star$] \textbf{Evidence across world models.} When hands change and come back, \method{} more than doubles the success of a frozen planner and costs little when nothing changes. It works unchanged with three world models \citep{lewm,dinowm,jepawms} on three simulated tasks, and with two of three models pretrained on real robot data in a simulated kitchen, where a moved camera turns the action frame as a hand does. It also leads test-time training on our schedules (\cref{sec:experiments}).
  \item[$\star$] \textbf{When and why it works.} The quality of the readings, which we can measure before deployment, sets the size of the gain. The gain comes from the readings and the library, not from the rule that picks a hand, as our analysis predicts. The library pays most when a hand takes several readings to identify (\cref{sec:models,sec:sensorq,sec:factorial}).
\end{itemize}
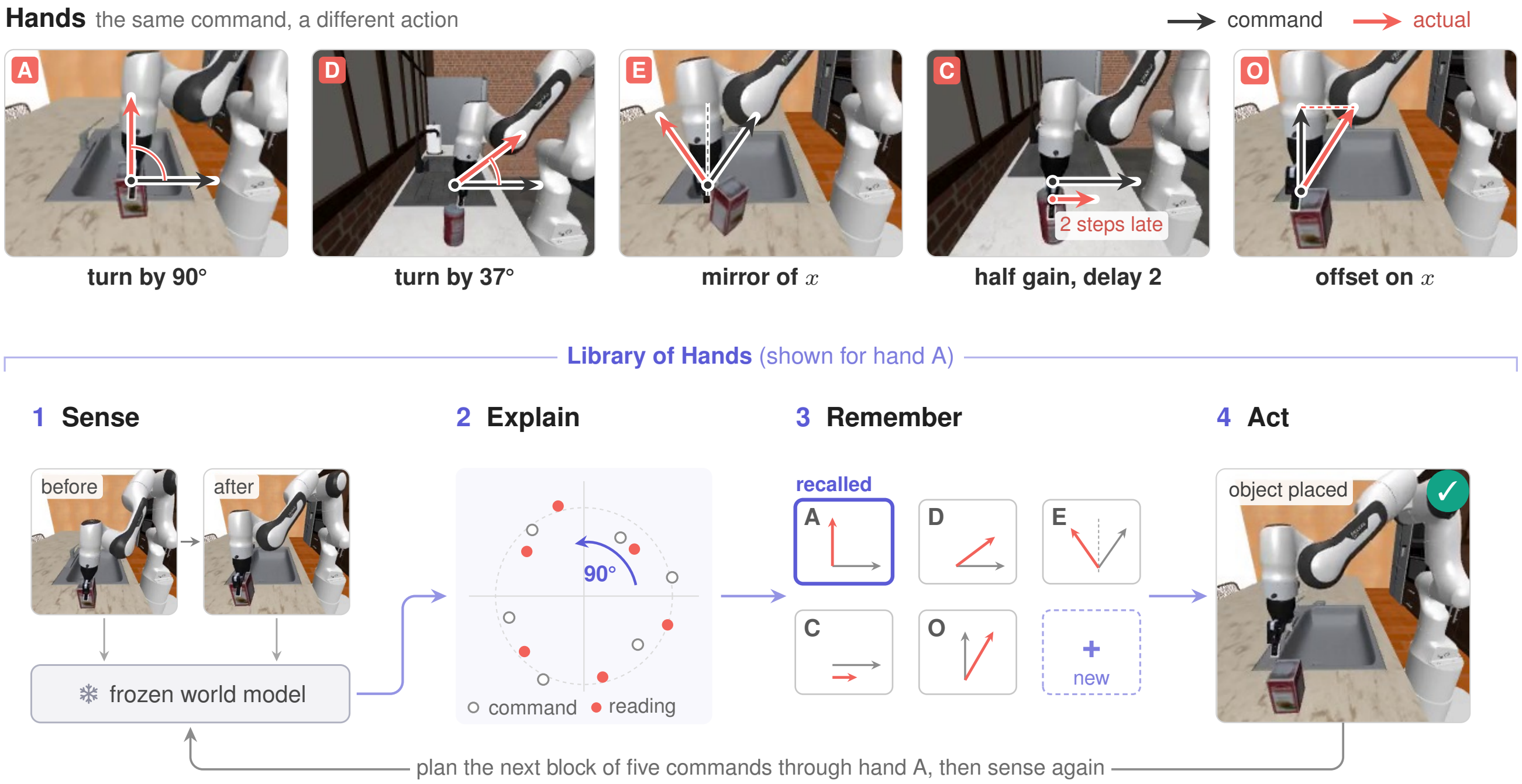
\begin{figure*}[t]
  \centering
  \tikzset{
    lohsnow/.pic={
      \foreach \lohang in {90,150,...,390} {
        \draw[#1, line width=0.55pt, line cap=round] (0,0) -- (\lohang:1);
        \draw[#1, line width=0.55pt, line cap=round] (\lohang:0.55) -- ++(\lohang+45:0.36);
        \draw[#1, line width=0.55pt, line cap=round] (\lohang:0.55) -- ++(\lohang-45:0.36);
      }
    },
  }
  \begin{tikzpicture}[
      x=1cm, y=1cm,
      font=\sffamily\fontsize{7}{8.4}\selectfont,
      ttl/.style={font=\sffamily\bfseries\fontsize{8.5}{10}\selectfont, text=black!88, inner sep=0pt, anchor=base west},
      desc/.style={font=\sffamily\fontsize{7}{8.4}\selectfont, text=black!58, inner sep=0pt, anchor=base west},
      tag/.style={font=\sffamily\fontsize{6.5}{7}\selectfont, fill=white, fill opacity=0.9, text opacity=1, rounded corners=1.5pt, inner sep=1.6pt},
      lettag/.style={font=\sffamily\bfseries\fontsize{7}{7}\selectfont, text=white, fill=cHand, fill opacity=0.95, text opacity=1, rounded corners=1.5pt, inner sep=1.8pt, anchor=north west},
      link/.style={-{Stealth[length=1.7mm, width=1.5mm]}, line width=0.7pt, black!45, rounded corners=4pt},
      flow/.style={-{Stealth[length=1.8mm, width=1.6mm]}, line width=0.8pt, cOurs!65, rounded corners=4pt},
      into/.style={-{Stealth[length=1.2mm, width=1.1mm]}, line width=0.55pt, black!35},
      img/.style={draw=black!18, line width=0.4pt, rounded corners=3pt},
      cmd/.style={black!78, line width=1.1pt, -{Stealth[length=2.2mm, width=2mm]}},
      act/.style={cHand, line width=1.3pt, -{Stealth[length=2.2mm, width=2mm]}},
      halo/.style={white, line width=3pt, line cap=round},
      gcmd/.style={black!45, line width=0.65pt, -{Stealth[length=1.1mm, width=1mm]}},
      gact/.style={cHand, line width=0.9pt, -{Stealth[length=1.1mm, width=1mm]}},
    ]
    \newcommand{\lohimg}[6]{%
      \begin{scope}
        \clip[rounded corners=3pt] (#1,#2) rectangle ++(#3,#4);
        \node[anchor=south west, inner sep=0] at (#1,#2) {\includegraphics[width=#3cm, trim=#6, clip]{figures/robocasa/#5.jpg}};
      \end{scope}
      \draw[img] (#1,#2) rectangle ++(#3,#4);
    }
    \newcommand{\lohdot}[1]{\fill[white] (#1) circle (0.075); \fill[black!80] (#1) circle (0.042);}

    \node[ttl] at (0,8.72) {Hands};
    \node[desc] at (1.02,8.72) {the same command, a different action};
    \draw[halo] (13.15,8.78) -- ++(0.55,0);
    \draw[cmd] (13.15,8.78) -- ++(0.55,0);
    \node[desc, text=black!75] at (13.82,8.72) {command};
    \draw[halo] (15.25,8.78) -- ++(0.55,0);
    \draw[act] (15.25,8.78) -- ++(0.55,0);
    \node[desc, text=cHand!85!black] at (15.92,8.72) {actual};

    \lohimg{0}{6.12}{3.2}{2.343}{s1_ep10_loh}{0 2 912 58}
    \coordinate (oA) at (1.43,6.98);
    \draw[halo] (oA) -- ++(0.95,0); \draw[cmd] (oA) -- ++(0.95,0);
    \draw[halo] (oA) -- ++(90:0.95); \draw[act] (oA) -- ++(90:0.95);
    \draw[white, line width=1.6pt] ([shift={(0:0.38)}]oA) arc[start angle=0, end angle=90, radius=0.38];
    \draw[cHand, line width=0.6pt] ([shift={(0:0.38)}]oA) arc[start angle=0, end angle=90, radius=0.38];
    \lohdot{oA}
    \node[lettag] at (0.07,8.39) {A};

    \lohimg{3.475}{6.12}{3.2}{2.343}{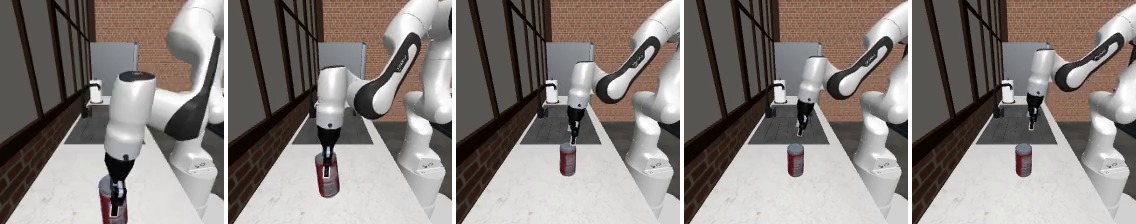}{456 40 456 20}
    \coordinate (oD) at (5.09,6.93);
    \draw[halo] (oD) -- ++(0.95,0); \draw[cmd] (oD) -- ++(0.95,0);
    \draw[halo] (oD) -- ++(37:0.95); \draw[act] (oD) -- ++(37:0.95);
    \draw[white, line width=1.6pt] ([shift={(0:0.5)}]oD) arc[start angle=0, end angle=37, radius=0.5];
    \draw[cHand, line width=0.6pt] ([shift={(0:0.5)}]oD) arc[start angle=0, end angle=37, radius=0.5];
    \lohdot{oD}
    \node[lettag] at (3.545,8.39) {D};

    \lohimg{6.95}{6.12}{3.2}{2.343}{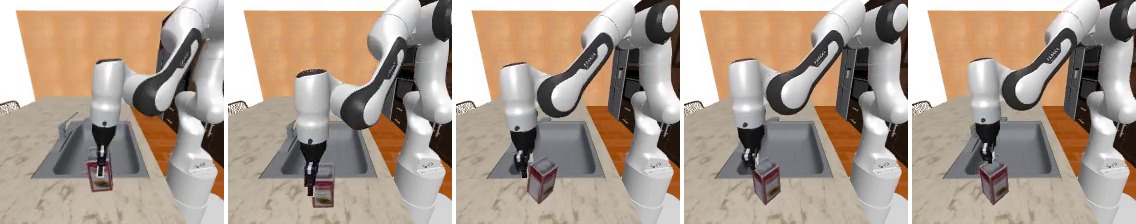}{456 4 456 56}
    \coordinate (oE) at (7.95,6.93);
    \draw[white, line width=1.4pt] ($(oE)+(0,-0.12)$) -- ++(0,1.05);
    \draw[black!55, line width=0.5pt, dash pattern=on 1.5pt off 1.2pt] ($(oE)+(0,-0.12)$) -- ++(0,1.05);
    \draw[halo] (oE) -- ++(55:0.95); \draw[cmd] (oE) -- ++(55:0.95);
    \draw[halo] (oE) -- ++(125:0.95); \draw[act] (oE) -- ++(125:0.95);
    \lohdot{oE}
    \node[lettag] at (7.02,8.39) {E};

    \lohimg{10.425}{6.12}{3.2}{2.343}{s8_ep31_loh}{228 17 684 43}
    \coordinate (oC) at (11.85,6.97);
    \draw[halo] (oC) -- ++(0.95,0); \draw[cmd] (oC) -- ++(0.95,0);
    \draw[halo] ($(oC)+(0,-0.2)$) -- ++(0.48,0); \draw[act] ($(oC)+(0,-0.2)$) -- ++(0.48,0);
    \lohdot{oC}
    \fill[white] ($(oC)+(0,-0.2)$) circle (0.06); \fill[cHand] ($(oC)+(0,-0.2)$) circle (0.035);
    \node[tag, text=cHand!85!black, anchor=north west] at ($(oC)+(0.02,-0.33)$) {2 steps late};
    \node[lettag] at (10.495,8.39) {C};

    \lohimg{13.9}{6.12}{3.2}{2.343}{s1_ep10_loh}{684 2 228 58}
    \coordinate (oO) at (14.66,6.86);
    \draw[halo] (oO) -- ++(0,0.95); \draw[cmd] (oO) -- ++(0,0.95);
    \draw[halo] (oO) -- ++(0.6,0.95); \draw[act] (oO) -- ++(0.6,0.95);
    \draw[white, line width=1.4pt] ($(oO)+(0,0.95)$) -- ++(0.6,0);
    \draw[cHand, line width=0.6pt, dash pattern=on 1.5pt off 1.2pt] ($(oO)+(0,0.95)$) -- ++(0.6,0);
    \lohdot{oO}
    \node[lettag] at (13.97,8.39) {O};

    \foreach \lohx/\lohwhat in {
        1.6/turn by 90\kern-0.12em\textdegree,
        5.075/turn by 37\kern-0.12em\textdegree,
        8.55/mirror of $x$,
        12.025/{half gain, delay 2},
        15.5/offset on $x$} {
      \node[font=\sffamily\bfseries\fontsize{7.5}{9}\selectfont, text=black!82, anchor=base, inner sep=0pt] at (\lohx,5.8) {\lohwhat};
    }

    \draw[cOurs!45, line width=0.6pt] (0,4.82) -- (0,4.98) -- (17.1,4.98) -- (17.1,4.82);
    \node[fill=white, inner sep=3pt, font=\sffamily\bfseries\fontsize{7.5}{9}\selectfont, text=cOurs] at (8.55,4.98) {Library of Hands {\mdseries\color{cOurs!80}(shown for hand A)}};
    \foreach \lohx/\lohn/\lohtitle in {
        0.3/1/Sense,
        5.1/2/Explain,
        8.94/3/Remember,
        13.7/4/Act} {
      \node[ttl, text=cOurs] (n) at (\lohx,4.2) {\lohn};
      \node[ttl] at ([xshift=1.6mm]n.base east) {\lohtitle};
    }
    \draw[flow] (3.98,1.175) -- (4.5,1.175) -- (4.5,2.28) -- (5.0,2.28);
    \draw[flow] (8.1,2.28) -- (8.84,2.28);
    \draw[flow] (12.94,2.28) -- (13.6,2.28);

    \lohimg{0.3}{2.07}{1.65}{1.65}{s1_ep10_loh}{228 0 684 0}
    \lohimg{2.25}{2.07}{1.65}{1.65}{s1_ep10_loh}{456 0 456 0}
    \node[tag, anchor=north west, text=black!70] at (0.35,3.67) {before};
    \node[tag, anchor=north west, text=black!70] at (2.3,3.67) {after};
    \draw[black!50, line width=0.6pt, -{Stealth[length=1.1mm, width=1.1mm]}] (1.99,2.895) -- (2.21,2.895);
    \draw[into] (1.125,2.03) -- (1.125,1.54);
    \draw[into] (3.075,2.03) -- (3.075,1.54);
    \draw[draw=cGrey!80, fill=cGrey!12, line width=0.55pt, rounded corners=3pt] (0.3,0.85) rectangle (3.9,1.5);
    \pic[scale=0.105] at (0.95,1.175) {lohsnow=cGrey!90!black};
    \node[font=\sffamily\fontsize{7.5}{9}\selectfont, text=black!72, anchor=west, inner sep=0pt] at (1.18,1.175) {frozen world model};

    \begin{scope}[shift={(6.55,2.28)}]
      \fill[cOurs!4, rounded corners=3pt] (-1.45,-1.45) rectangle (1.45,1.45);
      \draw[black!14, line width=0.4pt] (-1.3,0) -- (1.3,0);
      \draw[black!14, line width=0.4pt] (0,-1.08) -- (0,1.3);
      \draw[black!14, line width=0.4pt, dash pattern=on 1.2pt off 1.2pt] (0,0) circle (1.0);
      \foreach \lohang/\lohr/\lohe/\lohs in {12/1.02/4/1.06, 58/0.78/-6/0.82, 128/0.95/5/0.92, 196/0.88/-3/0.94, 244/1.05/7/1.0, 318/0.82/-5/0.78} {
        \draw[black!45, line width=0.6pt, fill=white] (\lohang:\lohr) circle (0.06);
        \fill[cHand] (\lohang+90+\lohe:\lohs) circle (0.064);
      }
      \draw[cOurs, line width=0.85pt, -{Stealth[length=1.5mm, width=1.4mm]}] (12:0.6) arc[start angle=12, end angle=100, radius=0.6];
      \node[font=\sffamily\bfseries\fontsize{7}{7}\selectfont, text=cOurs, inner sep=0.5pt] at (56:0.31) {90\kern-0.12em\textdegree};
      \draw[black!45, line width=0.6pt, fill=white] (-1.25,-1.26) circle (0.055);
      \node[font=\sffamily\fontsize{6.5}{7}\selectfont, text=black!60, anchor=west, inner sep=1pt] at (-1.12,-1.26) {command};
      \fill[cHand] (0.14,-1.26) circle (0.058);
      \node[font=\sffamily\fontsize{6.5}{7}\selectfont, text=black!60, anchor=west, inner sep=1pt] at (0.24,-1.26) {reading};
    \end{scope}

    \newcommand{\lohcard}[4]{%
      \draw[#3, fill=white, rounded corners=2.5pt] (#1,#2) rectangle ++(1.1,0.95);
      \node[font=\sffamily\bfseries\fontsize{7.5}{8}\selectfont, text=black!65, anchor=north west, inner sep=0pt] at (#1+0.09,#2+0.86) {#4};
    }
    \lohcard{8.94}{2.42}{cOurs, line width=1.2pt}{A}
    \lohcard{10.34}{2.42}{black!22, line width=0.5pt}{D}
    \lohcard{11.74}{2.42}{black!22, line width=0.5pt}{E}
    \lohcard{8.94}{1.17}{black!22, line width=0.5pt}{C}
    \lohcard{10.34}{1.17}{black!22, line width=0.5pt}{O}
    \draw[cOurs!55, fill=none, rounded corners=2.5pt, line width=0.6pt, dash pattern=on 1.6pt off 1.1pt] (11.74,1.17) rectangle ++(1.1,0.95);
    \node[font=\sffamily\bfseries\fontsize{11}{11}\selectfont, text=cOurs!80] at (12.29,1.69) {+};
    \node[font=\sffamily\fontsize{6.5}{7}\selectfont, text=cOurs!80, anchor=base] at (12.29,1.28) {new};
    \draw[gcmd] (9.36,2.62) -- ++(0.55,0);  \draw[gact] (9.36,2.62) -- ++(90:0.55);
    \draw[gcmd] (10.76,2.62) -- ++(0.55,0); \draw[gact] (10.76,2.62) -- ++(37:0.55);
    \draw[black!35, line width=0.4pt, dash pattern=on 1pt off 0.8pt] (12.37,2.55) -- ++(0,0.62);
    \draw[gcmd] (12.37,2.6) -- ++(55:0.55); \draw[gact] (12.37,2.6) -- ++(125:0.55);
    \draw[gcmd] (9.36,1.5) -- ++(0.55,0);   \draw[gact] (9.36,1.36) -- ++(0.28,0);
    \draw[gcmd] (10.86,1.33) -- ++(0,0.55); \draw[gact] (10.86,1.33) -- ++(0.32,0.55);
    \node[font=\sffamily\bfseries\fontsize{6.5}{7}\selectfont, text=cOurs, anchor=south west, inner sep=0pt] at (8.94,3.46) {recalled};

    \lohimg{13.7}{0.85}{2.87}{2.87}{s1_ep10_loh}{912 0 0 0}
    \node[circle, fill=cThree, inner sep=0pt, minimum size=4.8mm, text=white, font=\bfseries\fontsize{8.5}{8.5}\selectfont] at (16.32,3.47) {\ding{51}};
    \node[tag, anchor=north west, text=black!75] at (13.78,3.64) {object placed};

    \draw[link] (15.135,0.85) -- (15.135,0.32) -- (2.1,0.32) -- (2.1,0.8);
    \node[fill=white, inner sep=2pt, font=\sffamily\fontsize{7}{8.4}\selectfont, text=black!58] at (8.55,0.32) {plan the next block of five commands through hand A, then sense again};
  \end{tikzpicture}
  \caption{\textbf{Overview of \method{}.} \emph{Top:} a hand changes how each command moves the robot; the world does not change. The five main hands of \cref{tab:hands-main}, drawn on RoboCasa frames (the arrows are schematic). \emph{Bottom:} \method{} under hand A (DINO-WM in RoboCasa, the episode of \cref{fig:robocasa}). After each block of five steps, \method{} (1) asks the frozen model which action explains the change between two images (\cref{sec:sensor}); (2) fits a map and a delay to these readings (\cref{sec:mapfit}; the plot is schematic); (3) recalls the matching hand from its library, or stores a new one (\cref{sec:library}); and (4) plans the next commands through this hand (\cref{sec:plan}).}
  \label{fig:overview}
\end{figure*}

\section{Problem Setting}
\label{sec:setting}

\paragraph{Planning with a frozen world model.} An encoder (a neural network) turns an image $o_t$ into a latent state $z_t$, a vector that summarises the scene. A frozen predictor $f$ takes $z_t$ and an action block $a_t \in \R^{L \times m}$ and predicts the next state, $\hat z_{t+1} = f(z_t, a_t)$. A \emph{block} is a sequence of $L = 5$ consecutive control steps, each with an $m$-dimensional action; the predictor treats the whole block as one transition. The index $t$ therefore counts blocks, and the index $k$ below counts control steps. A planner, here the cross-entropy method (CEM), a sampling-based optimiser \citep{cem}, searches for the command blocks whose predicted states reach a goal state $z_g$, the latent state of a goal image.

\paragraph{Hands.} Between the planner and the robot, a \emph{hand} changes each command before it reaches the motors. In episode $e$, step $k$ applies
\begin{equation}
  u_k = M_e\, c_{k-d_e} + b_e,
  \label{eq:interface}
\end{equation}
where $c_k$ is the command. The matrix $M_e$ turns and scales the command (it can also mirror it, or be any linear map), $d_e \in \{0, \dots, 5\}$ delays it by $d_e$ control steps, and $b_e$ adds an offset. The robot then clips $u_k$ to its limits. In training, the hand is the identity, so the frozen predictor expects the applied action $u$ as its input. The robot does not know which hand is active. A hand belongs to the deployment, not to the world: the physics that the predictor learned stays correct, and only the meaning of a command changes. This is our main assumption: after a change, the frozen predictor stays accurate on what the robot observes, and the main mismatch can be written as a transformation of the actions. A new controller gain or delay meets it by construction. A moved camera also changes the images, so it meets the assumption only if the model still predicts well from the new view (\cref{sec:robocasa} tests this). Hands with a dead band (small commands have no effect) or a saturation (large commands are capped) do not fit \cref{eq:interface}; they test \method{} outside its model.

\paragraph{Task and measures.} The robot plays a sequence of episodes, each with a goal and one hand. Between episodes the hand can change, and hands seen before can come back. The robot must reach each goal with the same frozen model and with no label of the active hand. We measure success in three ways: over all episodes; over the first two episodes after a hand comes back (\emph{after returns}); and over runs in which the hand never changes (\emph{no change}). Success is the percentage of episodes that reach the goal, and a difference in success is given in percentage points (pp).
\section{Method}
\label{sec:method}

\paragraph{Our idea.} \method{} uses the frozen world model as a sensor of its own hand. If the predictor $f$ is right about the world, and an action has a visible effect, then what the robot sees tells us which action reached the robot. We find this action by running $f$ backwards. Its comparison with the command reveals the hand. \method{} trains nothing, and the weights of the world model never change. \Cref{fig:overview} shows the four stages, and \cref{alg:hands} gives the loop with every constant.

\subsection{Sense: reading the applied action}
\label{sec:sensor}

After each block of $L = 5$ control steps (\cref{sec:setting}), the robot knows the latent state $z_t$ before the block and the observed state $z_{t+1}$ after it. The applied action is the action that best explains this change under the frozen predictor:
\begin{equation}
  u^*_t = \argmin_{a} \; \lVert z_{t+1} - f(z_t, a) \rVert_2^2 .
  \label{eq:inversion}
\end{equation}
We solve \cref{eq:inversion} with gradient descent from nine starting points and keep the best one. We use a block only if this action explains it well, that is, if its error is small against the surprise threshold $\theta$ of the task, the 95th percentile of the prediction error of the frozen model on data with the unchanged hand (\cref{app:protocol}). An image shows only the net effect of a block, so the reading is the block sum $s_t = g^{-1} B u^*_t$. Here $B$ adds up the five steps of each action axis, and the gain $g$ (0.94--1.05 on our simulated tasks) corrects the scale of the inversion. We fit $g$ once, on 1,200 blocks with the unchanged hand and known actions; no episode with a changed hand and no label is used.

\paragraph{How far to trust a reading.} A reading is only as good as the visible effect of the action. When an action has little visible effect on the next observation, for example because the motion of the robot is blocked or the action saturates, a change of the action changes the predicted state only slightly, and the reading is weak (\cref{sec:setup} gives the cases in our tasks). The Jacobian $J_t$ of $f(z_t, \cdot)$ at $u^*_t$, the matrix of derivatives of the predicted state with respect to the action, measures this effect. With a unit Gaussian prior on the normalised action and a latent noise variance $s_z = \theta / D$ per dimension ($D$ is the latent dimension), a Laplace approximation (a Gaussian fitted at the optimum) of \cref{eq:inversion} gives the uncertainty of the reading,
\begin{equation}
  \Sigma_t = s_z \left(J_t^\top J_t + s_z I\right)^{-1},
  \label{eq:laplace}
\end{equation}
and its covariance $C_t = g^{-2} B \Sigma_t B^\top$ \citep{mackay1992practical}. A flat Jacobian gives a wide covariance, so the reading counts for little. The map fit uses the shape of $C_t$ as the noise of each reading.
\subsection{Explain: choosing a map}
\label{sec:mapfit}

Next, \method{} explains the readings with a map. For a delay $d$, let $\bar c_t^{(d)} = \sum_{i=0}^{4} c_{5t+i-d}$ be the sum of the commands that block $t$ applies under this delay. A map predicts the reading as $s_t = M \bar c_t^{(d)} + 5b + \varepsilon_t$ (the offset acts at each of the five steps), with the noise $\varepsilon_t \sim \mathcal{N}(0, \sigma^2 \tilde R_t)$ and $\tilde R_t$ the shape of $C_t$ (that is, $C_t$ up to a scale). We compare five classes of map (the number of parameters in parentheses): identity (0); turn and scale $\gamma R(\phi)$, a rotation by the angle $\phi$ with the gain $\gamma$ (2); mirror (2); linear (4); and affine, a linear map with an offset (6). A class is a form of map, not a hand: turn and scale alone holds every turn and gain. Each class has six delays $d \in \{0, \dots, 5\}$, so there are 30 models in all. The identity with no delay has the prior probability $0.5$. A conjugate prior, centred on the identity, gives each model a closed-form evidence, the probability of the readings under the model \citep{mackay1992bayesian,kass1995bayes}. The evidence is exact, so there is no complexity penalty to tune.

\paragraph{Actuator limits.} A physical actuator has a limited range, so many robots clip each applied action to a fixed interval, for example $[-1, 1]$. A clipped action moves the robot less than the linear map predicts, so a map fitted to the raw commands underestimates the gain. A gain above 1 cannot be observed when the actions saturate, so we limit the current map to a largest singular value (its largest gain in any direction) of 1, $T = \hat M / \max(1, \sigma_{\max}(\hat M))$. The effective commands $\tilde c = T^{-1}(\mathrm{clip}(Tc + b) - b)$ then make the linear model exact, and a fixed-point iteration refits the map on them.

\subsection{Remember: a library of hands}
\label{sec:library}

A stored hand is easier to recognise than a new hand is to identify. The library has no fixed list of hands: it keeps the map of every hand seen so far. In each episode, several sources compete to explain the readings $\mathcal{D}$ of the current segment (the blocks since the start of the episode or the last detected switch): each stored hand $k$, and a ``new'' source that fits a map from $\mathcal{D}$ alone. A stored hand scores $\mathcal{D}$ with its own posterior over the 30 models as the prior. So a stored hand needs only enough readings to be recognised, while a new hand must be identified.

\paragraph{Which hand to expect.} Hands usually stay, so the prior over the sources favours the hand of the previous episode (a \emph{sticky} prior). That hand gets $1 - h$, the new source gets $hq$, and the other stored hands share $h(1-q)$ in proportion to their use. The switch rate $h$ and the new-hand rate $q$ are learned from the last 20 events. The planner uses the most probable model of the source with the most posterior mass.

\paragraph{Keeping the library clean.} Inside an episode, the last four blocks are also scored alone, so a switch in the middle of an episode can start a new segment. At the end of an episode, a segment creates a new hand only if the new source explains it best and the segment has at least four blocks. Otherwise, only the hands that won the segment learn from its blocks, so the data of one hand do not overwrite another. Two hands merge when the evidence favours one hand over two, and unused hands fade (\cref{app:protocol}). A map with a smallest singular value below 0.2 (the weakest hand has gain 0.5) says that the commands have no effect: an obstruction or a lost reading, not a hand. The planner never uses it, and no segment that it explains best is stored (\cref{app:floor}).

\subsection{Act: planning through the map}
\label{sec:plan}

CEM samples command blocks, keeps the best ones (the \emph{elites}) and refits its sampling distribution to them (300 samples, 30 iterations, 30 elites, a horizon of five blocks). Before the frozen model predicts the outcome of a candidate, \method{} passes it through the chosen map, delay and clip. The frozen model therefore sees the actions that will really reach the robot. The cost is the squared distance from the final state to the goal. The robot replans at once when the one-block prediction error exceeds the surprise threshold $\theta$ (\cref{sec:sensor}), and otherwise after five blocks.
\subsection{What the theory predicts}
\label{sec:theory}

We analyse \method{} as Bayesian regression of the readings on the commands (Appendix~\ref{app:theory}). We write $\E[s_t] = X_t \beta$, where $\beta$ holds the parameters of a map and $X_t$ is linear in $\bar c_t^{(d)}$.

\begin{proposition}[Identifiability (Informal)]
\label{prop:ident}
For a fixed class and delay, $n$ readings identify a map if and only if the reliability-weighted Fisher information $\mathcal I_n = \sigma^{-2} \sum_{t \le n} X_t^\top \tilde R_t^{-1} X_t$ is nonsingular (invertible). One block with $\bar c_t^{(d)} \neq 0$ identifies a turn with a gain, or a mirror. 
\end{proposition}

\begin{remarkbox}
\textbf{Remark.} \emph{For a fixed transformation class and delay, identifiability is determined by the rank of the accumulated information matrix rather than the number of observations alone. Rank deficiency leaves some parameter directions unidentifiable, whereas a single block with $\bar c_t^{(d)} \neq 0$ suffices for the rotation-with-gain and reflection classes. Identifiability should be distinguished from estimation precision: increasing the noise variance reduces Fisher information without changing its rank, and therefore preserves identifiability while weakening the precision attainable from the observations.}
\end{remarkbox}

A tracking bound (Proposition~\ref{prop:th-hard}, Corollary~\ref{cor:th-tie}) adds that the library pays only a few nats (units of log-evidence) at each switch, so selection rules that differ in this prior, such as Fixed Share, an algorithm that tracks the best of several predictors over time \citep{herbster1998fixedshare}, and the cumulative-sum (CUSUM) change detector \citep{page1954cusum}, choose almost the same hand (\cref{sec:factorial}), and every rule plans the first block after a switch with the previous hand.

Lemma~\ref{lem:th-noise} explains the false alarms when nothing changes, and Proposition~\ref{prop:th-delay} shows that identifying a delay requires, in general, one more block of readings than identifying the map alone. Of 24 testable predictions, 16 hold on the logs of the final method (Table~\ref{tab:th-checks}).
\section{Experiments}
\label{sec:experiments}

Our experiments address four questions. (i) Does \method{} recover the success lost to a changed hand, without a loss when the hand does not change (\cref{sec:main})? (ii) Does it transfer to other world models (\cref{sec:models})? (iii) Why does it work: does the gain follow the quality of the sensor (\cref{sec:sensorq}), and which component of \method{} produces it (\cref{sec:factorial})? (iv) Does it outperform learned and calibration methods, and a planner with a larger search budget (\cref{sec:competitors})?

\subsection{Experimental Setup}
\label{sec:setup}

\paragraph{World models and tasks.} Our main world model is LeWM (LeWorldModel; \citealp{lewm}), a joint-embedding predictive architecture (JEPA) trained from pixels, with its released checkpoints (trained weights) for three simulated control tasks. In \emph{PushT} \citep{ibc,diffusionpolicy}, a circular pusher pushes a T-shaped block to a target position and angle. In \emph{Reacher} \citep{dmc}, a planar arm with two joints moves to a target joint configuration. In \emph{Two-Room} \citep{pldm}, a point agent moves to a target in a plane that a wall with a door divides into two rooms. In all three tasks, the action is 2-D and the world model observes the scene through images (\cref{tab:tasks}). To test whether \method{} depends on LeWM, we also run two other world models on the same three tasks. DINO-WM \citep{dinowm} adds action features to the frozen features of DINOv2, a pretrained image encoder \citep{oquab2024dinov2}. JEPA-WM \citep{jepawms} feeds the action through adaptive layer normalisation. On PushT we use their released checkpoints and data. Reacher and Two-Room have no released checkpoints, so we train both models on the data of LeWM with their authors' recipes (one run each). A small adapter gives each model the interface of LeWM. The method code does not change. In these tasks, readings are weak when the agent pushes against a wall (Two-Room) and when an action is clipped at the actuator limits (Reacher and Two-Room clip each applied action to $[-1, 1]$). Without the clip model of \cref{sec:mapfit}, the stored map of the unchanged hand on Two-Room had a gain of 0.70 instead of 1. Cube, in which a simulated 3-D robot arm with 5-D actions moves a cube, is a further test (\cref{sec:discussion}).

\paragraph{Choice of hands.} The hands are test conditions; \method{} is not given a list of them. Each hand changes one term of \cref{eq:interface}, as a real change of set-up does (\cref{tab:hands-main}), so that each result can be attributed to that term. Each value also tests a different part of \method{}. Under the turn of 90\degr{}, the frozen planner almost always fails (7.8\%; \cref{tab:byhand}), so hand A shows the full loss. The turn of 37\degr{} is not a multiple of 45\degr{}, so a choice among a few fixed turns cannot express it. No turn can express the mirror, so it needs its own class of map. Hand C combines a gain with a delay, so the map fit must also find the delay. The random turns T are new hands that no list contains. They are drawn from the whole range $[-90\degr, 90\degr]$, so the gain does not rest on the value 37\degr{}: on them, \method{} raises success from 28.6\% to 57.0\% (\cref{tab:byhand}). The offset $b$ translates the applied action: at every step, the robot moves by an additional constant displacement, as with an actuator bias or a miscalibrated zero. We evaluate the offset on a separate schedule (\cref{sec:more}). Hands outside \cref{eq:interface}, with dead bands and saturation, evaluate \method{} beyond its model.

\begin{table}[t]
  \centering
  \caption{\textbf{The main hands.} Each changes one part of \cref{eq:interface}. Delays are in control steps. \Cref{tab:hands} gives the others.}
  \label{tab:hands-main}
  \footnotesize
  \setlength{\tabcolsep}{2.5pt}
  \begin{tabular}{@{}llll@{}}
    \toprule
    Hand & Change & Part & Real cause \\
    \midrule
    N & none & none & the training set-up \\
    A, D & turn 90\degr{}, 37\degr{} & direction & a moved camera \\
    T & random turn & direction & a manually set camera \\
    E & mirror of $x$ & handedness & an inverted axis \\
    C & half gain, delay 2 & scale, timing & a slower controller \\
    O & offset on $x$ & translation & an actuator bias \\
    \bottomrule
  \end{tabular}
\end{table}

\paragraph{Schedules.} A schedule sets the hand of every episode, and a \emph{return} is a switch back to a hand that was already active earlier in the run. Each schedule has 140 episodes and tests one situation. In S1 (\emph{fixed returns}), hands return in a fixed, predetermined order: the schedule cycles through A, C, A, C, D, A, D, and each hand stays for five consecutive episodes. S1 thus mixes the first visits of three hands with their returns at known times, so it shows recall visit by visit (\cref{fig:memory}). In S2 (\emph{random returns}), the last hand stays with probability 0.6. Otherwise, the schedule draws a new turn in $[-90\degr, 90\degr]$ with probability 0.1, and else one of A, C, D, E, N with equal probability. Hands thus return at random times and new hands appear, so neither the order nor the set of hands is known in advance. S3 (no returns) applies a new turn in every episode: there is nothing to recall, so S3 measures the identification of new hands alone. S4 (60 episodes) never changes the hand, and it measures the cost of \method{} when nothing changes. The \emph{after returns} measure is the success in the first two episodes after each return (the switch episode and the next one), pooled over S1 and S2.

\paragraph{Baselines.} We compare \method{} with nine baselines in three groups. Simple: (1) the frozen planner; (2) identification in every episode, with no memory; (3) one memory, which keeps only the last map. Hand trackers: (4) Fixed Share over the same library \citep{herbster1998fixedshare}; (5) a change detector in the style of MBCD \citep{mbcd}, which switches between stored models with a CUSUM test; (6) a method in the style of ShiftWM-Bench \citep{shiftwm}, which assigns the change to one of a few fixed classes, with no memory. Adaptation and search: (7) test-time training of the action encoder of the world model, and AdaJEPA, an adaptive latent world model \citep{adajepa}; (8) a calibration at the start of every episode \citep[App.~B.4]{vjepa2}; (9) the frozen planner with two and four times more CEM iterations. Baselines (5)--(8) are our re-implementations. Two oracles, which receive privileged information, mark the ceiling: one plans with the true hand, and one is told which hand is active but fits its map with the sensor of \method{}.

\paragraph{Protocol.} We evaluate all methods with fixed protocols and report paired bootstrap confidence intervals across runs. A run is one task, seed and schedule, and all methods play the same episodes. We give each paired difference as its mean over runs, with its 95\% bootstrap confidence interval (CI) in square brackets. The main comparison, the \emph{locked test}, uses 20 new seeds and sign-flip permutation tests with Holm's correction over its eight comparisons (\cref{app:protocol}). Each test has a pass threshold (a \emph{bar}) that we fixed before its runs; \cref{app:prereg} gives every verdict and the development history of the method. All results of \method{} use the final method of \cref{sec:method}, with the floor of \cref{sec:library}; \method{} without the floor is an ablation (\cref{sec:factorial}).

\begin{table}[t]
  \centering
  \caption{\textbf{Locked test on 20 new seeds.} Success (\%) on S1--S3, after returns, and with no change (S4). \textbf{Bold}: best; \underline{underlined}: second (oracle excluded). $^\dagger$Earlier sensor (\cref{sec:factorial}). $^\ddagger$Run on 12 of the 20 seeds; on these 12 seeds, the frozen planner and \method{} reach 28.4 and 62.4 over all episodes, and 26.1 and 60.3 after returns.}
  \label{tab:main}
  \footnotesize
  \setlength{\tabcolsep}{3.2pt}
  \begin{tabular}{@{}lrrrrrr@{}}
    \toprule
    Method & PushT & Reach. & 2-Room & All & Returns & S4 \\
    \midrule
    Frozen planner & 20.6 & 28.6 & 36.6 & 28.6 & 26.6 & \best{78.1} \\
    ShiftWM-style & 36.2 & 41.0 & 42.0 & 39.7 & 35.8 & 77.3 \\
    No memory$^\dagger$ & 48.4 & 44.3 & 47.0 & 46.6 & 43.8 & 76.9 \\
    One memory$^\dagger$ & 54.4 & 48.7 & \second{48.6} & 50.6 & 45.0 & 76.3 \\
    MBCD-style$^\dagger$ & 61.5 & 53.7 & 48.1 & 54.4 & 52.0 & 74.2 \\
    Fixed Share$^\dagger$ & \second{63.4} & \second{58.7} & 47.1 & \second{56.4} & \second{54.9} & 75.0 \\
    \oursrow \method{} (ours) & \best{64.7} & \best{65.4} & \best{56.9} & \best{62.3} & \best{60.4} & \second{77.5} \\
    \midrule
    \textcolor{black!55}{True hand$^\ddagger$} & \textcolor{black!55}{83.0} & \textcolor{black!55}{81.7} & \textcolor{black!55}{67.8} & \textcolor{black!55}{77.5} & \textcolor{black!55}{77.0} & \textcolor{black!55}{78.1} \\
    \bottomrule
  \end{tabular}
\end{table}

\subsection{\method{} Recalls Returning Hands}
\label{sec:main}

\Cref{tab:main} gives the locked test. \obs{1} \textbf{\method{} more than doubles the success of the frozen planner.} Over all episodes, success rises from 28.6\% to 62.3\%. On the 12 seeds on which we also ran the true hand, \method{} closes 69.2\% of the gap to it (frozen planner 28.4\%, \method{} 62.4\%, true hand 77.5\%). After returns, \method{} is on average 33.9 pp above the frozen planner (95\% CI [31.1, 36.6]). \obs{2} \textbf{\method{} beats every baseline.} It is best on every task over all episodes, and after returns it leads Fixed Share by 5.5 pp and the MBCD-style detector by 8.4. It leads one memory by 15.5 and the ShiftWM-style method by 24.6 (all Holm-corrected $p < 0.001$; \cref{app:results}). \obs{3} \textbf{It costs little when nothing changes.} With no change, \method{} loses 0.6 pp [$-0.9$, $-0.2$]. This loss is statistically significant (Holm-corrected $p = 0.003$) but small. In 95.4\% of the S4 episodes, every plan uses the exact identity; false alarms in the other episodes cause almost all of the loss.

\begin{figure}[t]
  \centering
\begin{tikzpicture}
  \begin{axis}[lohplot, name=ma, set layers,
      width=6.55cm, height=2.3cm,
      xmin=0.55, xmax=8.45, ymin=0, ymax=100,
      xtick={1,2,3,4,5,6,7,8}, ytick={0,25,50,75,100}, ymajorgrids,
      xlabel={Visit of the hand (5 episodes each)},
      ylabel={Success (\%)},
    ]
    \begin{pgfonlayer}{axis background}
      \fill[lohShadeA!80] (axis cs:0.55,0) rectangle (axis cs:1.5,100);
    \end{pgfonlayer}
    \node[anchor=north, inner sep=1pt, font=\scriptsize\itshape, text=black!55] at (axis cs:1.025,99) {new};
    \node[anchor=north, inner sep=1pt, font=\scriptsize\itshape, text=black!55] at (axis cs:4.97,99) {returns of a known hand};
    \addplot[fill=black, fill opacity=0.05, draw=none, forget plot] coordinates {(1,70.0) (2,70.8) (3,72.5) (4,69.7) (5,75.8) (6,68.6) (7,71.1) (8,69.4) (8,81.1) (7,83.1) (6,79.2) (5,85.6) (4,81.7) (3,82.8) (2,81.1) (1,82.5)} -- cycle;
    \addplot[fill=cOurs, fill opacity=0.1, draw=none, forget plot] coordinates {(1,53.9) (2,63.6) (3,64.7) (4,65.8) (5,65.3) (6,65.3) (7,64.2) (8,65.3) (8,77.8) (7,75.8) (6,76.1) (5,74.7) (4,76.7) (3,75.3) (2,72.5) (1,65.0)} -- cycle;
    \addplot[fill=cThree, fill opacity=0.1, draw=none, forget plot] coordinates {(1,56.7) (2,54.2) (3,57.5) (4,59.4) (5,58.9) (6,55.8) (7,60.5) (8,54.4) (8,63.9) (7,70.0) (6,66.4) (5,70.0) (4,69.4) (3,67.2) (2,63.9) (1,66.9)} -- cycle;
    \addplot[fill=cBar, fill opacity=0.1, draw=none, forget plot] coordinates {(1,36.9) (2,40.8) (3,36.1) (4,38.6) (5,40.3) (6,36.9) (7,36.1) (8,34.7) (8,47.2) (7,47.2) (6,48.6) (5,51.9) (4,49.2) (3,48.1) (2,50.6) (1,47.8)} -- cycle;
    \addplot[fill=cGrey, fill opacity=0.1, draw=none, forget plot] coordinates {(1,15.8) (2,21.4) (3,20.0) (4,18.9) (5,19.2) (6,17.5) (7,15.8) (8,17.8) (8,28.9) (7,26.4) (6,30.0) (5,30.3) (4,29.4) (3,29.5) (2,33.1) (1,26.1)} -- cycle;
    \addplot[black!72, line width=0.9pt, densely dashed] coordinates {(1,76.4) (2,76.1) (3,77.8) (4,75.8) (5,80.8) (6,73.9) (7,77.2) (8,75.6)};
    \addplot[cOurs, line width=3.6pt, opacity=0.16, forget plot] coordinates {(1,59.2) (2,68.1) (3,70.0) (4,71.4) (5,70.0) (6,70.8) (7,70.3) (8,71.7)};
    \addplot[cOurs, line width=1.4pt, mark=*, mark size=1.7pt, mark options={fill=cOurs, draw=white, line width=0.35pt}] coordinates {(1,59.2) (2,68.1) (3,70.0) (4,71.4) (5,70.0) (6,70.8) (7,70.3) (8,71.7)};
    \addplot[cThree, line width=1.0pt, mark=*, mark size=1.4pt, mark options={fill=cThree, draw=white, line width=0.35pt}] coordinates {(1,61.7) (2,58.9) (3,62.5) (4,64.7) (5,64.4) (6,61.4) (7,65.3) (8,59.2)};
    \addplot[cBar, line width=1.0pt, mark=*, mark size=1.4pt, mark options={fill=cBar, draw=white, line width=0.35pt}] coordinates {(1,42.5) (2,45.8) (3,41.9) (4,43.9) (5,45.8) (6,42.5) (7,41.9) (8,40.8)};
    \addplot[cGrey, line width=1.0pt, mark=*, mark size=1.4pt, mark options={fill=cGrey, draw=white, line width=0.35pt}] coordinates {(1,21.1) (2,27.2) (3,24.4) (4,24.2) (5,24.7) (6,23.6) (7,21.1) (8,23.3)};
    \node[anchor=south west, inner sep=0pt, font=\fontsize{8}{9.6}\selectfont] at ([xshift=-0.95cm, yshift=0.22cm]rel axis cs:0,1) {\textbf{(a)}\enspace Success by visit of a hand};
  \end{axis}
  \begin{axis}[lohplot, name=mb, set layers, at={($(ma.south west)+(0,-1.62cm)$)}, anchor=north west,
      width=6.55cm, height=2.3cm,
      xmin=0.7, xmax=5.3, ymin=0, ymax=100,
      xtick={1,2,3,4,5}, ytick={0,25,50,75,100}, ymajorgrids,
      xlabel={Episode after a hand returns},
      ylabel={Success (\%)},
      legend style={at={(0.42,-0.47)}, anchor=north, legend columns=3, column sep=7pt, row sep=1pt},
    ]
    \begin{pgfonlayer}{axis background}
      \fill[lohShadeA!80] (axis cs:0.7,0) rectangle (axis cs:1.5,100);
    \end{pgfonlayer}
    \node[anchor=north, inner sep=1pt, font=\scriptsize\itshape, text=black!55, align=center] at (axis cs:1.1,99) {blind\\[-2pt]start};
    \addplot[fill=black, fill opacity=0.05, draw=none, forget plot] coordinates {(1,71.7) (2,74.7) (3,72.0) (4,70.5) (5,73.0) (5,81.2) (4,79.8) (3,80.0) (2,82.3) (1,82.2)} -- cycle;
    \addplot[fill=cOurs, fill opacity=0.1, draw=none, forget plot] coordinates {(1,49.5) (2,72.0) (3,68.8) (4,71.8) (5,70.7) (5,79.5) (4,79.2) (3,76.0) (2,78.5) (1,58.2)} -- cycle;
    \addplot[fill=cThree, fill opacity=0.1, draw=none, forget plot] coordinates {(1,37.5) (2,57.3) (3,63.3) (4,67.3) (5,69.2) (5,78.3) (4,75.5) (3,71.8) (2,64.8) (1,44.5)} -- cycle;
    \addplot[fill=cBar, fill opacity=0.1, draw=none, forget plot] coordinates {(1,39.8) (2,36.7) (3,37.7) (4,39.5) (5,36.7) (5,46.2) (4,46.8) (3,47.0) (2,44.3) (1,49.5)} -- cycle;
    \addplot[fill=cGrey, fill opacity=0.1, draw=none, forget plot] coordinates {(1,19.5) (2,15.2) (3,17.7) (4,16.3) (5,17.2) (5,26.0) (4,23.7) (3,25.7) (2,26.2) (1,28.5)} -- cycle;
    \addplot[black!72, line width=0.9pt, densely dashed] coordinates {(1,77.2) (2,78.5) (3,76.0) (4,75.3) (5,77.3)};
    \addlegendentry{True hand}
    \addplot[cOurs, line width=3.6pt, opacity=0.16, forget plot] coordinates {(1,54.0) (2,75.3) (3,72.3) (4,75.7) (5,75.3)};
    \addplot[cOurs, line width=1.4pt, mark=*, mark size=1.7pt, mark options={fill=cOurs, draw=white, line width=0.35pt}] coordinates {(1,54.0) (2,75.3) (3,72.3) (4,75.7) (5,75.3)};
    \addlegendentry{\method{} (ours)}
    \addplot[cThree, line width=1.0pt, mark=*, mark size=1.4pt, mark options={fill=cThree, draw=white, line width=0.35pt}] coordinates {(1,41.0) (2,61.0) (3,67.5) (4,71.5) (5,73.8)};
    \addlegendentry{One memory}
    \addplot[cBar, line width=1.0pt, mark=*, mark size=1.4pt, mark options={fill=cBar, draw=white, line width=0.35pt}] coordinates {(1,44.7) (2,40.3) (3,42.2) (4,43.2) (5,41.5)};
    \addlegendentry{No memory}
    \addplot[cGrey, line width=1.0pt, mark=*, mark size=1.4pt, mark options={fill=cGrey, draw=white, line width=0.35pt}] coordinates {(1,24.0) (2,20.5) (3,21.5) (4,20.0) (5,21.5)};
    \addlegendentry{Frozen planner}
    \node[anchor=south west, inner sep=0pt, font=\fontsize{8}{9.6}\selectfont] at ([xshift=-0.95cm, yshift=0.22cm]rel axis cs:0,1) {\textbf{(b)}\enspace Success after a hand returns};
  \end{axis}
  \path (mb.south) ++(0,-1.84);
\end{tikzpicture}%
  \caption{\textbf{A hand that returns is recalled, not learned again.} Success on S1 over the three LeWM tasks (8 seeds, 24 runs per method). (a)~By visit of a hand (blocks of five episodes). (b)~By episode after a known hand returns. Bands: 95\% CIs over runs.}
  \label{fig:memory}
\end{figure}
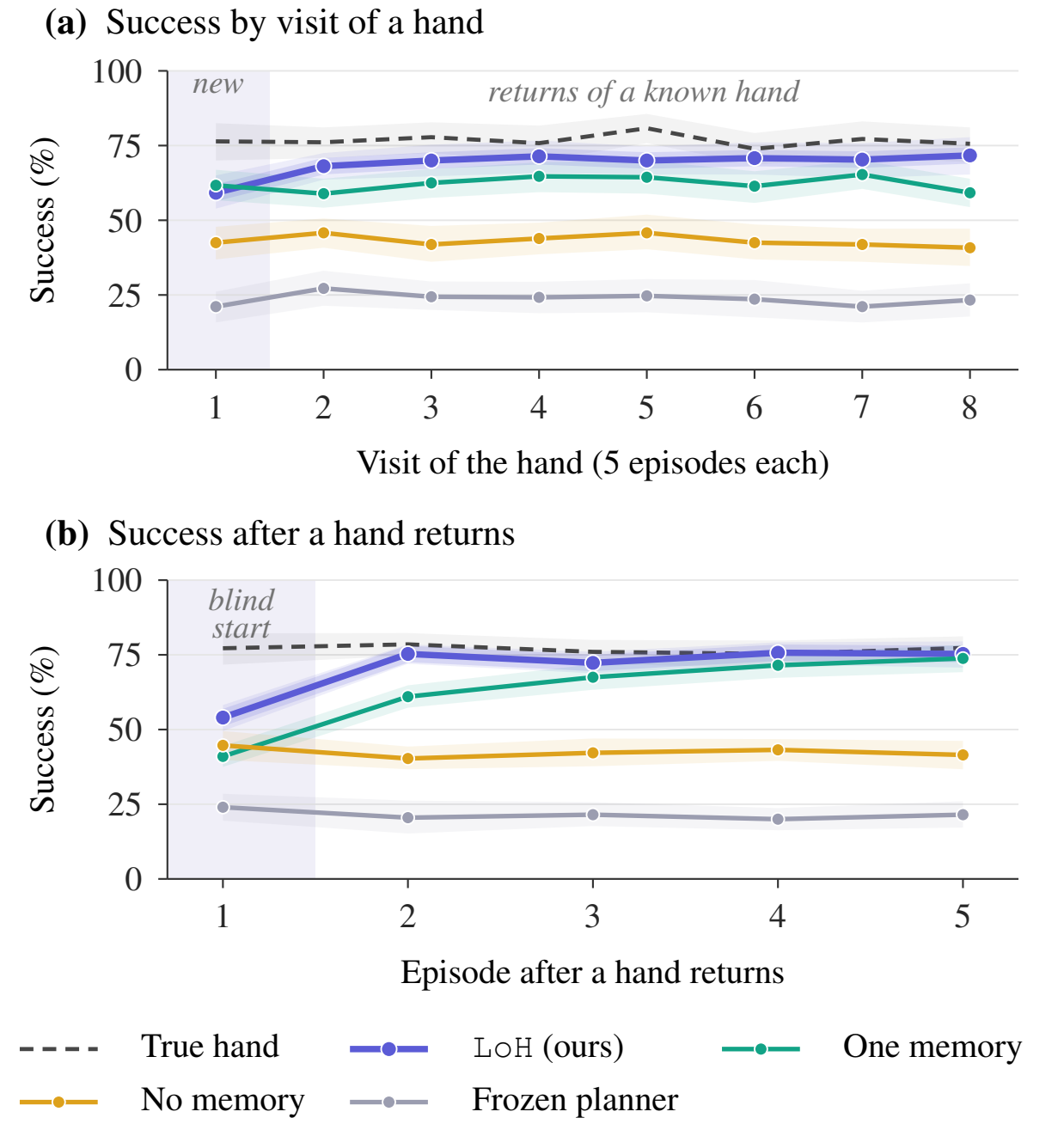

\Cref{fig:memory} shows how the library makes this gain on S1. At the first visit of a hand, \method{} must identify it (59.2\%, against 76.4\% for the true hand). When the hand returns, the library recalls it (70.5\% against 76.9\% over the 25 returns). In the second episode after a return, \method{} matches the true hand (75.3\% against 78.5\%) and leads one memory by 14.3 pp. One memory with the same sensor must identify the hand again after each switch and needs about four episodes (63.0\% over the returns; no memory: 42.4\%).

\begin{table}[t]
  \centering
  \caption{\textbf{One method, three world models.} Values in pp: \method{} minus the frozen planner (S1, S2: after returns; S4: no change) or minus one memory (1 mem.). LeWM: 20 seeds; others: 6 seeds (DINO-WM PushT S2: 12) and the first 35, 70 and 20 episodes of S1, S2 and S4. $R^2$: quality of the sensor, the share of the variance of the applied block action that the readings explain. $^\ast$Trained by us on LeWM's data.}
  \label{tab:models}
  \footnotesize
  \setlength{\tabcolsep}{2.6pt}
  \resizebox{\columnwidth}{!}{%
  \begin{tabular}{@{}llrrrrrr@{}}
    \toprule
    Model & Task & $R^2$ & S1 & 1 mem. & S4 & S2 & 1 mem. \\
    \midrule
    LeWM & PushT & 0.98 & $\mathbf{+52.2}$ & $+11.4$ & $-0.8$ & $\mathbf{+36.9}$ & $+9.0$ \\
     & Reacher & 0.70 & $\mathbf{+48.9}$ & $+17.3$ & $-0.8$ & $\mathbf{+28.7}$ & $+11.5$ \\
     & 2-Room & 0.90 & $\mathbf{+23.5}$ & $+10.0$ & $-0.1$ & $\mathbf{+13.0}$ & $+6.0$ \\
    \midrule
    DINO-WM & PushT & 0.998 & $\mathbf{+58.3}$ & $-8.3$ & $-0.8$ & $\mathbf{+46.0}$ & $+5.5$ \\
     & Reacher$^\ast$ & 0.981 & $\mathbf{+56.2}$ & $+4.2$ & $-2.5$ & $\mathbf{+57.9}$ & $+5.5$ \\
     & 2-Room$^\ast$ & 0.950 & $\mathbf{+31.3}$ & $+6.2$ & $-0.8$ & $\mathbf{+26.3}$ & $+11.5$ \\
    \midrule
    JEPA-WM & PushT & 0.997 & $\mathbf{+41.7}$ & $-2.1$ & $0.0$ & $\mathbf{+39.7}$ & $+2.7$ \\
     & Reacher$^\ast$ & 0.981 & $\mathbf{+64.6}$ & $-4.2$ & $+0.8$ & $\mathbf{+63.1}$ & $+10.0$ \\
     & 2-Room$^\ast$ & 0.929 & $\mathbf{+43.8}$ & $+12.5$ & $0.0$ & $\mathbf{+30.3}$ & $+10.4$ \\
    \bottomrule
  \end{tabular}}
\end{table}

\subsection{One Method, Three World Models}
\label{sec:models}

\Cref{tab:models} runs the same method on two other world models, which feed the action into the predictor in different ways, on the same three tasks; \cref{fig:teaser} gives the share of the gap closed in each cell (one world model on one task). \obs{1} \textbf{Both read their own actions well} (sensor $R^2$ 0.93--0.998; without the joint-state (proprioceptive) input, 0.87--0.88 on Reacher; \cref{app:proprio}). \obs{2} \textbf{\method{} gains in every cell.} After returns, it is 31.3 to 64.6 pp above the frozen planner on S1 and 26.3 to 63.1 on S2, where hands come back at random. With no change, it stays within 2 pp of the frozen planner in eight of the nine cells; on DINO-WM Reacher it loses 2.5 pp [$-7.5$, $+2.5$]. \obs{3} \textbf{The library pays most when a hand is slow to identify.} Recall saves the readings that a fresh fit needs, so its value depends on how many readings identify a hand. On S1, the library leads one memory with the same sensor by 10.0 to 17.3 pp on LeWM, whose readings are less accurate ($R^2$ 0.70--0.98). On DINO-WM and JEPA-WM ($R^2$ 0.93--0.998), one memory is close in five of six cells ($-8.3$ to $+6.2$ pp): one accurate block identifies a hand again, so recall saves little. Over the nine cells, the lead of the library falls as the $R^2$ of the sensor rises (Spearman $\rho = -0.89$ on S1 and $-0.72$ on S2; post hoc). In RoboCasa, a fresh fit is confident by the third of 12 plans, so recall also saves little there (\cref{sec:robocasa}). Very noisy readings, in turn, spoil the stored maps (\cref{sec:sensorq}). The library is thus most useful when a hand takes several readings to identify, and each reading is still accurate enough to store a correct map. On S2, where new hands also appear, it leads in four of six cells (5.5 to 11.5 pp, 95\% interval above 0).

\subsection{World Models Pretrained on Real Robot Data}
\label{sec:robocasa}

We test \method{} on the three world models that \citet{jepawms} pretrained on real robot data \citep{droid}: DINO-WM, JEPA-WM and \mbox{V-JEPA~2-AC}. Each plans in RoboCasa \citep{robocasa}, a simulated 3-D kitchen, with its released planner unchanged, and the hand acts on the commands of the end effector (the gripper). An admission test (a turn of 90\degr{} must cost the frozen planner 15 pp) admits the \emph{place} task, in which the robot puts an object at a target, for all three models; we do not run the \emph{reach} task, where the frozen planner rarely succeeds even with the true hand. Schedule S1 runs 35 episodes in blocks of five, and the success criteria are those of the simulated tasks (\cref{app:robocasa}).

\begin{table}[t]
  \centering
  \caption{\textbf{World models pretrained on real robot data.} RoboCasa, place task (S1: 8 seeds; no change: 4 seeds): success after returns on S1 (\%), share of the gap to the true hand closed, and \method{} minus frozen with no change (S4).}
  \label{tab:robocasa-main}
  \resizebox{\columnwidth}{!}{%
  \begin{tabular}{@{}lrrrrrrr@{}}
    \toprule
    Model & $R^2$ & Frozen & True hand & \method{} & Gap & 1 mem. & No change \\
    \midrule
    DINO-WM & 0.80 & 7.8 & 35.9 & $\mathbf{26.6}$ & 67\% & 29.7 & $+7.5$ \\
    JEPA-WM & 0.70 & 14.1 & 35.9 & 21.9 & 36\% & 25.0 & $-6.2$ \\
    \mbox{V-JEPA~2-AC} & 0.68 & 21.9 & 35.9 & $\mathbf{35.9}$ & 100\% & 31.2 & $+8.8$ \\
    \bottomrule
  \end{tabular}}
\end{table}

\begin{figure}[t]
  \centering
  \setlength{\tabcolsep}{0pt}
  \begin{tabular}{@{}r@{\hspace{3pt}}c@{}}
    \rotatebox{90}{\scriptsize\hspace{6pt}Frozen} & \includegraphics[width=0.95\columnwidth]{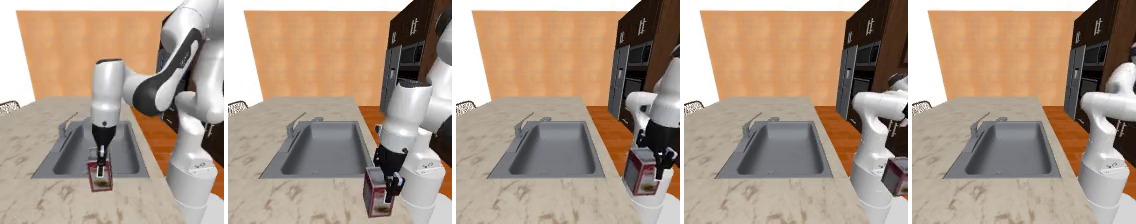} \\[1pt]
    \rotatebox{90}{\scriptsize\hspace{8pt}\method{}} & \includegraphics[width=0.95\columnwidth]{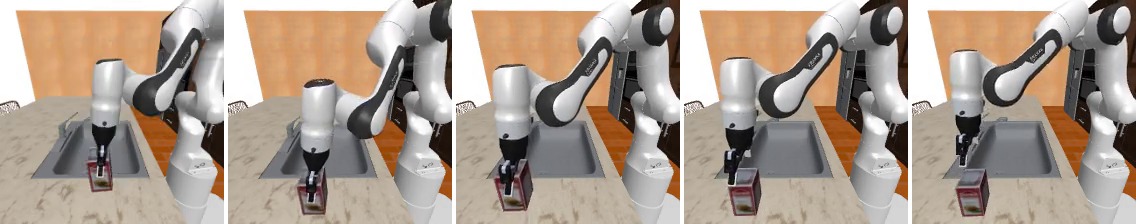} \\
  \end{tabular}
  \caption{\textbf{The same returning hand, two planners.} DINO-WM in RoboCasa (episode 10 of one run): hand A, a turn of 90\degr{}, returns after five episodes of the mirror E; robot camera at planning steps 0 to 60. The frozen arm swings away and fails; \method{} recalls hand A and places the object.}
  \label{fig:robocasa}
\end{figure}

On DINO-WM, \method{} lifts success after returns from 7.8\% to 26.6\% (\cref{fig:robocasa}), does not lose when nothing changes, and over all 35 episodes equals the planner told the hand (30.7\% and 30.4\%). On S2 (random returns), it gains 11.4 pp [5.3, 17.1]. On \mbox{V-JEPA~2-AC}, it gains 14.1 pp after returns, with a wide interval ([$-3.1$, $+29.7$]), and equals the planner told the hand (35.9\%). On JEPA-WM, it gains only 7.8 pp [3.1, 10.9] and costs 6.2 pp with no change. \method{} thus works on two of the three models (\cref{app:robocasa} gives the criteria). Each model reads the RoboCasa commands in a frame turned by 22 to 39\degr{}, which follows the camera: a camera move by $\theta$ turns it by $-0.68\,\theta$ on DINO-WM and $-0.85\,\theta$ on \mbox{V-JEPA~2-AC} over all camera yaw angles, near the slope of $-0.67$ that it showed offline on DROID images (\cref{fig:rc-camera}). A camera move thus turns the frame as a hand does (the test fails on JEPA-WM; \cref{app:robocasa}). \method{} removes this turn with the injected hand, so it can equal the planner told the hand. \method{} recalls a returning hand at its first plan in 27 to 38\% of the returns; its lead over one memory ($-3.1$ to $+4.7$ pp) is not significant: a fresh fit is confident by plan 3 of 12, so recall saves little (post hoc).

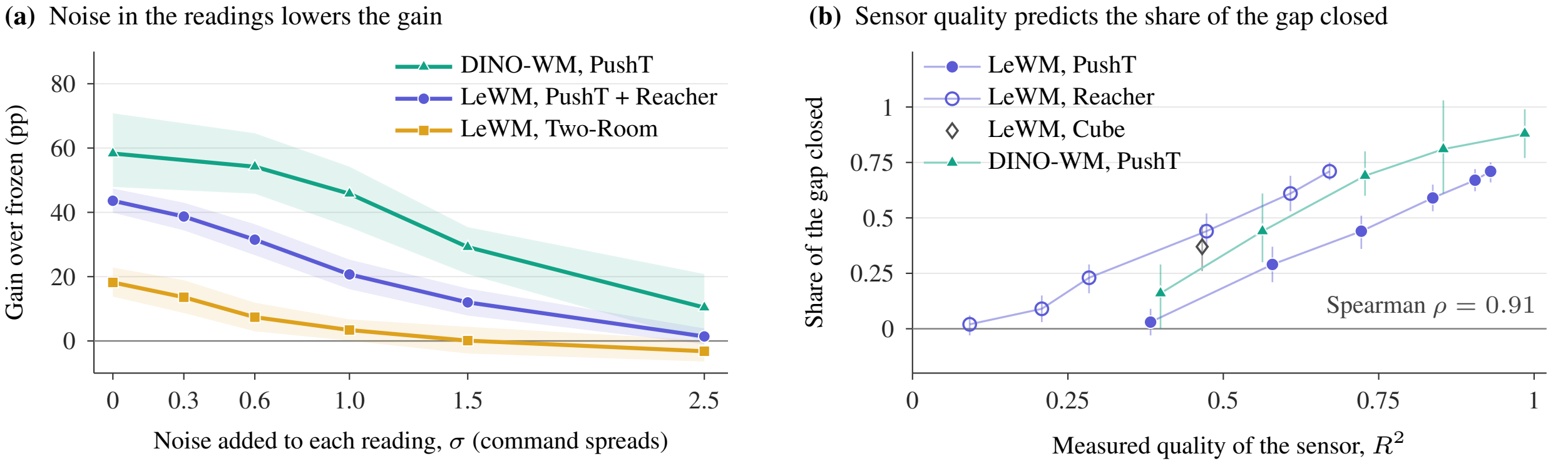
\begin{figure*}[t]
  \centering
\begin{tikzpicture}
  \begin{axis}[lohplot, name=pa,
      width=6.7cm, height=3.4cm,
      xmin=-0.08, xmax=2.6, ymin=-10, ymax=90,
      xtick={0,0.3,0.6,1,1.5,2.5}, xticklabels={0,0.3,0.6,1.0,1.5,2.5},
      ytick={0,20,40,60,80}, ymajorgrids,
      xlabel={Noise added to each reading, $\sigma$ (command spreads)},
      ylabel={Gain over frozen (pp)},
      legend style={at={(1,1.02)}, anchor=north east},
    ]
    \draw[black!50, line width=0.45pt] (axis cs:-0.08,0) -- (axis cs:2.6,0);
    \addplot[fill=cThree, fill opacity=0.12, draw=none, forget plot] coordinates {(0,47.9) (0.6,45.8) (1,35.4) (1.5,20.8) (2.5,0) (2.5,20.8) (1.5,35.4) (1,54.2) (0.6,64.6) (0,70.8)} -- cycle;
    \addplot[fill=cOurs, fill opacity=0.12, draw=none, forget plot] coordinates {(0,39.9) (0.3,34.4) (0.6,26.7) (1,16.1) (1.5,7.9) (2.5,-1) (2.5,3.9) (1.5,16.3) (1,25.3) (0.6,36.3) (0.3,43) (0,47.4)} -- cycle;
    \addplot[fill=cBar, fill opacity=0.12, draw=none, forget plot] coordinates {(0,13.8) (0.3,8.7) (0.6,3) (1,0.1) (1.5,-3.9) (2.5,-6.4) (2.5,0.2) (1.5,4.4) (1,6.7) (0.6,11.9) (0.3,18.9) (0,22.8)} -- cycle;
    \addplot[cThree, line width=1.2pt, mark=triangle*, mark size=2.1pt, mark options={fill=cThree, draw=white, line width=0.35pt}]
      coordinates {(0,58.3) (0.6,54.2) (1,45.8) (1.5,29.2) (2.5,10.4)};
    \addlegendentry{DINO-WM, PushT}
    \addplot[cOurs, line width=1.2pt, mark=*, mark size=1.75pt, mark options={fill=cOurs, draw=white, line width=0.35pt}]
      coordinates {(0,43.6) (0.3,38.7) (0.6,31.5) (1,20.7) (1.5,12) (2.5,1.4)};
    \addlegendentry{LeWM, PushT + Reacher}
    \addplot[cBar, line width=1.2pt, mark=square*, mark size=1.6pt, mark options={fill=cBar, draw=white, line width=0.35pt}]
      coordinates {(0,18.2) (0.3,13.6) (0.6,7.4) (1,3.4) (1.5,0.1) (2.5,-3.2)};
    \addlegendentry{LeWM, Two-Room}
    \node[anchor=south west, inner sep=0pt, font=\fontsize{8}{9.6}\selectfont] at ([xshift=-0.95cm, yshift=0.22cm]rel axis cs:0,1) {\textbf{(a)}\enspace Noise in the readings lowers the gain};
  \end{axis}
  \begin{axis}[lohplot, at={(pa.south east)}, anchor=south west, xshift=1.95cm,
      width=6.7cm, height=3.4cm,
      xmin=0, xmax=1.02, ymin=-0.2, ymax=1.25,
      xtick={0,0.25,0.5,0.75,1}, xticklabels={0,0.25,0.5,0.75,1},
      ytick={0,0.25,0.5,0.75,1}, yticklabels={0,0.25,0.5,0.75,1}, ymajorgrids,
      xlabel={Measured quality of the sensor, $R^2$},
      ylabel={Share of the gap closed},
      legend style={at={(0.01,1.02)}, anchor=north west},
    ]
    \draw[black!50, line width=0.45pt] (axis cs:0,0) -- (axis cs:1.02,0);
    \addplot[cOurs!50, line width=0.6pt, mark=*, mark size=1.9pt, mark options={solid, fill=cOurs, draw=white, line width=0.35pt},
        error bars/.cd, y dir=both, y explicit, error mark=none,
        error bar style={cOurs, line width=0.5pt, opacity=0.45}]
      coordinates {(0.383,0.03) += (0,0.06) -= (0,0.06) (0.579,0.29) += (0,0.08) -= (0,0.08) (0.722,0.44) += (0,0.07) -= (0,0.08) (0.837,0.59) += (0,0.06) -= (0,0.06) (0.905,0.67) += (0,0.05) -= (0,0.05) (0.930,0.71) += (0,0.04) -= (0,0.05)};
    \addlegendentry{LeWM, PushT}
    \addplot[cOurs!50, line width=0.6pt, mark=o, mark size=1.9pt, mark options={solid, fill=white, draw=cOurs, line width=0.65pt},
        error bars/.cd, y dir=both, y explicit, error mark=none,
        error bar style={cOurs, line width=0.5pt, opacity=0.45}]
      coordinates {(0.092,0.02) += (0,0.04) -= (0,0.05) (0.208,0.09) += (0,0.06) -= (0,0.06) (0.284,0.23) += (0,0.06) -= (0,0.07) (0.473,0.44) += (0,0.08) -= (0,0.08) (0.608,0.61) += (0,0.08) -= (0,0.08) (0.671,0.71) += (0,0.04) -= (0,0.04)};
    \addlegendentry{LeWM, Reacher}
    \addplot[black!70, only marks, mark=diamond, mark size=2.2pt, mark options={solid, fill=white, draw=black!70, line width=0.65pt},
        error bars/.cd, y dir=both, y explicit, error mark=none,
        error bar style={black!70, line width=0.5pt, opacity=0.45}]
      coordinates {(0.466,0.37) += (0,0.10) -= (0,0.11)};
    \addlegendentry{LeWM, Cube}
    \addplot[cThree!50, line width=0.6pt, mark=triangle*, mark size=2.2pt, mark options={solid, fill=cThree, draw=white, line width=0.35pt},
        error bars/.cd, y dir=both, y explicit, error mark=none,
        error bar style={cThree, line width=0.5pt, opacity=0.45}]
      coordinates {(0.399,0.16) += (0,0.13) -= (0,0.16) (0.563,0.44) += (0,0.17) -= (0,0.14) (0.728,0.69) += (0,0.11) -= (0,0.09) (0.854,0.81) += (0,0.22) -= (0,0.20) (0.985,0.88) += (0,0.11) -= (0,0.11)};
    \addlegendentry{DINO-WM, PushT}
    \node[anchor=south east, inner sep=1pt, font=\plotfont, text=black!75] at (axis cs:1.01,0.03) {Spearman $\rho = 0.91$};
    \node[anchor=south west, inner sep=0pt, font=\fontsize{8}{9.6}\selectfont] at ([xshift=-1.1cm, yshift=0.22cm]rel axis cs:0,1) {\textbf{(b)}\enspace Sensor quality predicts the share of the gap closed};
  \end{axis}
\end{tikzpicture}%
  \caption{\textbf{The quality of the sensor sets the size of the gain.} (a)~Gain over the frozen planner after returns, with noise of standard deviation $\sigma$ (in units of the standard deviation of the commands) added to each reading (LeWM: 8 seeds; DINO-WM: 6 seeds). (b)~Share of the gap closed against the $R^2$ of the readings (18 data points; Two-Room has no unsaturated blocks). Bands and bars: 95\% bootstrap intervals.}
  \label{fig:sensor}
\end{figure*}

\subsection{The Quality of the Sensor Sets the Gain}
\label{sec:sensorq}

\method{} rests on one claim: the frozen model is a sensor of its own hand. If this is true, a worse sensor must give a smaller gain. We add Gaussian noise to each reading, with a standard deviation of $\sigma$ times the standard deviation of the commands, add its variance to the reliability of the reading, and change nothing else. The gain over the frozen planner falls at every step: on LeWM from 43.6 to 1.4 pp on PushT and Reacher and from 18.2 to $-3.2$ on Two-Room, and on DINO-WM from 58.3 to 10.4 (\cref{fig:sensor}a). All three tests of this fall pass. Across models and noise levels, the share of the possible gain that \method{} captures rises with the measured quality of the sensor (\cref{fig:sensor}b; Spearman rank correlation $\rho = 0.91$, 18 data points). A fourth test fails: the advantage of the library over no memory does not grow with the noise ($+0.2$ [$-3.9$, $+4.4$]), because noisy readings spoil stored maps. The value of the library thus peaks at an intermediate quality of the readings: on PushT, the task with the most accurate readings, it is largest at $\sigma = 0.6$; on Two-Room, at $\sigma = 0.3$; on Reacher, the least accurate, it only falls (\cref{tab:sweep}; point estimates, post hoc).

\subsection{Where the Gain Comes From}
\label{sec:factorial}

The gain should come from the sensor, the choice of map and the library. The baselines marked $\dagger$ in \cref{tab:main} use an earlier sensor: a manually tuned motion threshold, no reliability weights and no clip model. Their gap thus mixes the sensor and the library. \Cref{fig:factorial} therefore removes one stage of \cref{fig:overview} at a time (per task: \cref{tab:factorial-detail}).

\begin{figure}[t]
  \centering
\begin{tikzpicture}[x=1cm, y=0.056cm, font=\plotfont, every node/.append style={inner sep=0pt, text=black!80}]
  \def\w{0.31}
  \foreach \v in {20,40} \draw[black!9, line width=0.35pt] (-0.5,\v) -- (6.9,\v);
  \foreach \x/\v in {1.05/55.2, 2.13/52.0, 3.21/44.5, 4.29/26.0, 6.55/58.4}
    \fill[cOurs!15] (\x-\w,\v) rectangle (\x+\w,61.1);
  \fill[cOurs] (-\w,0) rectangle (\w,61.1);
  \foreach \x/\v in {1.05/55.2, 2.13/52.0, 3.21/44.5, 4.29/26.0, 5.5/62.0, 6.55/58.4}
    \fill[cGrey] (\x-\w,0) rectangle (\x+\w,\v);
  \draw[cOursDeep, line width=0.6pt, dash pattern=on 2.4pt off 1.6pt] (-\w,61.1) -- (6.9,61.1);
  \foreach \x/\lo/\hi in {1.05/53.3/57.2, 2.13/49.7/54.2, 3.21/41.4/47.6, 4.29/21.5/30.5, 5.5/60.2/63.8, 6.55/56.6/60.2} {
    \draw[black!80, line width=0.55pt] (\x,\lo) -- (\x,\hi);
    \draw[black!80, line width=0.55pt] (\x-0.05,\lo) -- (\x+0.05,\lo) (\x-0.05,\hi) -- (\x+0.05,\hi);
  }
  \draw[black!60, line width=0.45pt] (-0.5,0) -- (6.9,0);
  \foreach \v in {0,20,40,60} {
    \draw[black!60, line width=0.4pt] (-0.5,\v) -- (-0.56,\v);
    \node[anchor=east] at (-0.6,\v) {\v};
  }
  \node[rotate=90, anchor=south] at (-0.92,30) {Success after returns (\%)};
  \node[anchor=base, font=\plotfont\bfseries] at (0,65.5) {61.1};
  \foreach \x/\t in {1.05/$-$5.9, 2.13/$-$9.1, 3.21/$-$16.6, 4.29/$-$35.1, 5.5/+0.9, 6.55/$-$2.7}
    \node[anchor=base] at (\x,65.5) {\t};
  \draw[black!45, line width=0.4pt] (1.05-\w,73) -- (4.29+\w,73) (5.5-\w,73) -- (6.55+\w,73);
  \node[anchor=base] at (2.67,75) {One stage removed};
  \node[anchor=base] at (6.025,75) {Other rule};
  \node[anchor=base, font=\plotfont\bfseries] at (0,-8) {\method{}};
  \foreach \x/\t in {1.05/Sense, 2.13/Explain, 3.21/Remember, 4.29/Act, 5.5/Fixed, 6.55/CUSUM}
    \node[anchor=base] at (\x,-8) {\t};
  \node[anchor=base] at (5.5,-15) {Share};
\end{tikzpicture}%
  \caption{\textbf{Ablation study of framework components and the selection rule.} Results are aggregated over 48 runs across multiple random seeds. Values denote changes in success rate relative to \method{}, with error bars indicating 95\% intervals. Component ablations are defined as follows: \emph{Sense}, replacing the sensor with its earlier variant, with the floor disabled in both configurations; \emph{Explain}, using a single fixed map; \emph{Remember}, removing memory; and \emph{Act}, using a frozen planner.}

  \label{fig:factorial}
\end{figure}
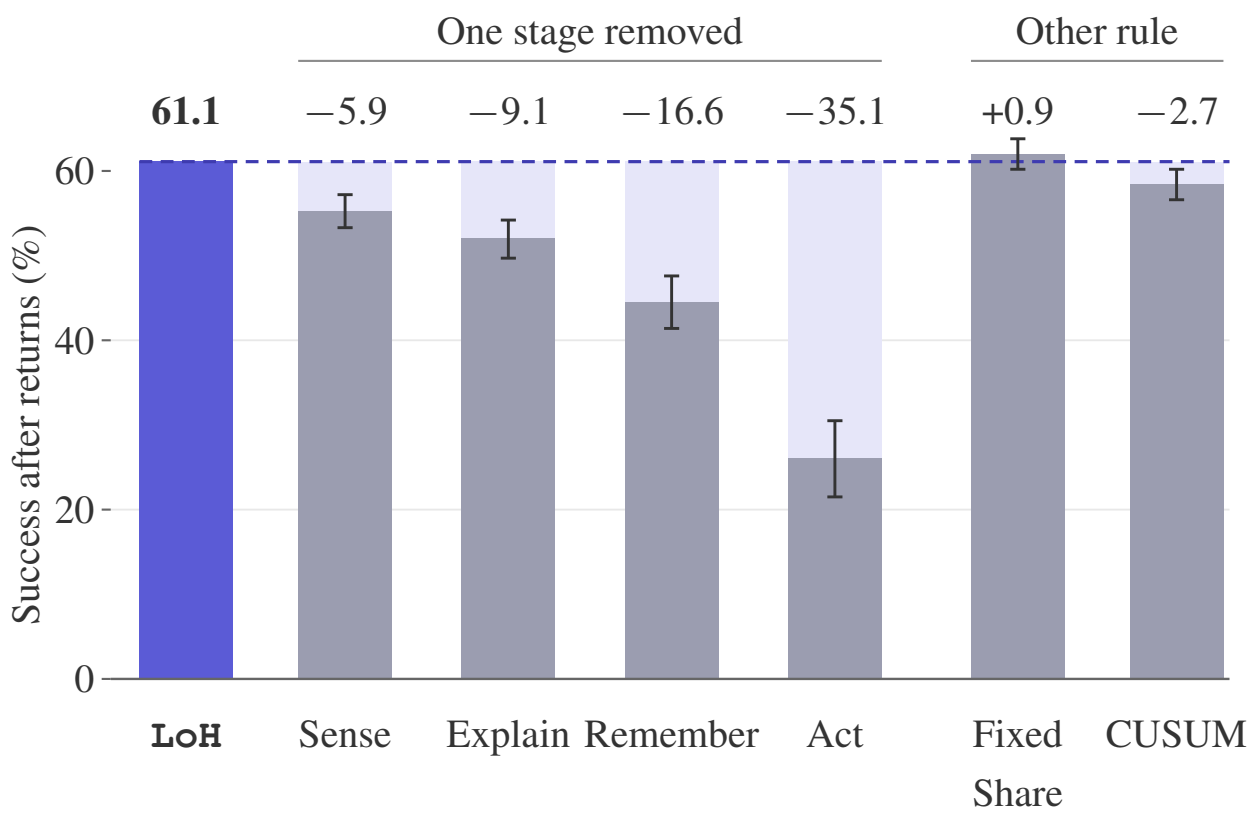
\begin{itemize}
  \item \textbf{Sense.} The full sensor adds 5.9 pp [3.9, 7.8] with the sticky prior and 6.2 pp [3.9, 8.6] with Fixed Share. Noise on the readings removes almost the whole gain (\cref{sec:sensorq}).
  \item \textbf{Explain.} One fixed linear map costs 9.1 pp [6.9, 11.4]. No delay search costs 6.1 pp [3.6, 8.5], and 25.1 [20.6, 29.6] on the delayed hand C.
  \item \textbf{Remember.} With the same sensor, \method{} is 16.6 pp [13.5, 19.7] above no memory and 12.3 pp [10.1, 14.5] above one memory.
  \item \textbf{The floor.} On the locked test, \method{} without the floor is within 0.4 pp of \method{} (S1--S3: $-0.4$ [$-0.9$, $+0.1$]). The floor matters in long runs: in the 500-episode streams of \cref{sec:competitors} on PushT with no change, the library without it stores maps under which the commands have no effect, and its AUC falls from 82.5 to 52.9 (frozen planner: 83.3; \cref{app:floor}).
  \item \textbf{Act.} Without a map in the plan, only the frozen planner remains: 35.1 pp [30.6, 39.6] below \method{}, and 33.9 [31.1, 36.6] on all 20 seeds of the locked test.
  \item \textbf{The rule that picks a hand matters little.} With the same sensor and library, the sticky prior ties with Fixed Share ($-0.9$ [$-2.7$, $+0.9$]), as \cref{sec:theory} predicts, and is 2.7 pp [0.9, 4.5] above CUSUM. The $+5.5$ over Fixed Share in \cref{tab:main} comes from the sensor, so we do not claim the rule.
  \item \textbf{Recognition is the remaining gap.} An oracle that is told the hand, with the same sensor, is 15.3 pp [13.2, 17.3] higher after returns. The gap is the first block after a switch, planned with the previous hand. A cue from the first image of an episode closes about 40\% of this gap ($+6.0$ [4.0, 8.2]; \cref{tab:further}).
\end{itemize}

\subsection{Comparison with Alternatives and Compute}
\label{sec:competitors}

\paragraph{Test-time training.} On the protocol and the eight dynamics shifts of JEPA-TTT \citep{jepattt}, a test-time training method for latent world models (streams of 500 episodes; \cref{app:jepattt}), \method{} leads JEPA-TTT by 14.3 pp of best success on the action rotation and by 2.8 pp of the area under the success curve (AUC) over six shifts. It loses on Two-Room, where the turn depends on the position. With no change, it stays within 1 pp of the frozen planner, while JEPA-TTT loses up to 11. In our schedules, continual test-time training, AdaJEPA \citep{adajepa}, JEPA-TTT and a calibration in every episode \citep[App.~B.4]{vjepa2} are 34.1, 33.7, 33.8 and 18.4 pp below \method{} after returns.

\paragraph{The gain is not extra compute.} A correct map causes fewer surprises, so \method{} replans about half as often and uses 0.78--0.82 times the GPU time per step of the frozen planner. More search does not help: with four times the CEM iterations, the frozen planner costs 3.6--4.2 times as much per step, loses 2.6 pp [0.6, 4.9] after returns, and stays 36.1 pp [31.0, 41.2] below \method{} (\cref{app:jepattt}).

\paragraph{Further results (\cref{tab:further}).}\label{sec:more} With 2, 4 and 8 recurring hands, \method{} is 4.4, 4.8 and 6.9 pp above one memory (same sensor); with 16 and 24, 8.1 and 4.2, and it ties Fixed Share; fresh seeds and Two-Room give the same verdicts. Over 400 episodes, success does not decay ($-2.8$ [$-8.5$, $+3.1$]). On hands with dead bands and saturations, which a linear map cannot express, \method{} is still 15.9 pp [14.0, 17.7] above the frozen planner. An offset $b$ costs the frozen planner 41.8 pp, and \method{} wins back 23.4 [14.0, 32.6]; on Two-Room, where it costs 3.1, \method{} is 6.2 below it, mostly in the first episode after a switch (post hoc). A partial run on DINO-WM and JEPA-WM points the same way, with no verdict. With one hand for 140 episodes, \method{} equals the true hand ($-0.3$ pp).
\section{Related Work}
\label{sec:related}

\paragraph{Latent world models and the meaning of an action.} Latent world models plan from pixels by searching through a learned predictor \citep{lewm,dinowm,vjepa2,pldm,robojepa}. Latent actions are relative to the camera \citep{garrido}, and calibrated methods draw the robot axes into the image with known camera poses \citep{axisguide}. We instead treat the hand as a property of the deployment and identify it through the released model.

\paragraph{Test-time adaptation and trained identification.} Test-time training changes the weights during deployment \citep{sun2020ttt,jepattt,adajepa}. Sandwich-Residuals fits a residual on the action embedding in each episode \citep{sandwich}, and RWM, which uses a world model as a reference trajectory, trains a residual controller online \citep{rwm}. Methods trained to recognise the set-up infer it from recent observations: with context encoders \citep{rma}, in-context world models \citep{syncworld,icwm}, or a meta-learned belief through a frozen dynamics model \citep{scout}. The closest work, ShiftWM-Bench \citep{shiftwm}, uses the same frozen LeWM and tasks. It trains an ensemble that sorts residuals into three or four fixed classes, with no memory across episodes. Our 90\degr{} and mirror hands lie outside all of these classes. \method{} instead estimates a continuous map through the released model and keeps it across episodes.

\paragraph{Inversion, action grounding and adaptive control.} V-JEPA~2 inverts its frozen predictor on a random trajectory to fit a $2 \times 2$ map offline, as a calibration phase that its experiments do not use \citep[App.~B.4]{vjepa2}. AIME (action inference by maximising evidence) infers actions under a frozen world model for imitation \citep{aime}. Grounded action transformation \citep{gat}, disturbance observers and L1 adaptive control \citep{l1book} correct the input of a fixed model. Multiple-model adaptive control and SpikeWorld choose among fixed models \citep{athans1977mmac,spikeworld}. ECA (efference-copy adaptation) identifies and inverts an actuator map (ordering, direction and gain) from the error of a frozen forward model, for policies on states, and withdraws when the identity or a delayed identity explains the data better \citep{eca}. We identify the map online through a frozen \emph{pixel} world model.

\paragraph{Libraries of models and tracking.} In motor control, MOSAIC, a set of paired forward and inverse models \citep{mosaic}, and COIN, a model of contextual inference \citep{coin}, infer the context over a set of motor memories, also from a cue before the movement. In model-based reinforcement learning, growing sets of dynamics models with a sticky or Chinese-restaurant prior appear in \citet{xu2020gpmixture}, in meta-learned online adaptation (MOLe; \citealp{mole}), in lifelong incremental reinforcement learning (LLIRL; \citealp{llirl}) and in change-point detection for non-stationary reinforcement learning (MBCD; \citealp{mbcd}). \Citet{fssl} analyse first visits and revisits of recurring dynamics. Choosing a stored hand is tracking the best expert. Fixed Share pays the identification cost in every segment \citep{herbster1998fixedshare}, and Mixing Past Posteriors pays it once for each distinct expert \citep{bousquet2002mpp,koolen2012sleep}. Change-point methods detect switches \citep{page1954cusum,adams2007bocpd}.
\raggedbottom
\section{Limitations and Discussion}
\label{sec:discussion}

\paragraph{Switches inside an episode.} When the hand changes in the middle of an episode, \method{} has a mean gain of 6.6 pp (95\% CI [4.2, 9.4]), below its bar of 10 pp. On new seeds, the mean gain is 9.9 pp (95\% CI [8.0, 11.8]). It replans only after a surprise, so it reads the new hand late.

\paragraph{The blind first block.} Every rule plans the first block of an episode with the previous hand. When the hand goes back to ``no change'', this costs 9.4 pp [7.2, 11.4] against the frozen planner (S2, post hoc). After 120 episodes of one hand, nearly all the loss is this episode. It is the recognition gap of \cref{sec:factorial}, which a cue from the first image of an episode reduces.

\paragraph{Changes of the world.} When the world changes and the hand does not (the friction or the centre of gravity of the PushT block), \method{} loses 5.0 pp [3.2, 7.1], so the bar of $-3$ failed. The change adds scatter to the readings, and model selection sometimes accepts a false map.

\paragraph{When the planner is the limit.} On Cube, the true hand is only 12 pp above the frozen planner, and \method{} closes half of this gap; two of the three bars fail (\cref{app:results}).

\paragraph{The frame of a real-data model.} The models pretrained on real robot data read the RoboCasa commands in a frame that follows the camera on two of three models (\cref{sec:robocasa}). Our sensor removes only the gain, so \method{} learns the turn as a permanent hand; a sensor that removes the full calibrated map is the next step. Behind the robot, outside the DROID cameras, the sensor reads little: there, the change of view breaks our assumption that the model stays accurate (\cref{sec:setting}). Our other hands are injected, and a comparison with models trained on randomised interfaces \citep{syncworld,icwm} remains open.

\paragraph{Toward robots.} The library has no fixed list of hands: it adds an entry for each new hand, and we test it with up to 24. Its five classes of map are fixed, however, so it cannot express a kind of hand outside them. A real robot needs more. The first step is a module that discovers new kinds of hand online, such as hands that change with the state (the Two-Room loss of \cref{app:jepattt}). The others are harder and longer tasks \citep{libero}, other robot bodies, and a test on a real robot.

\section{Conclusion}
\label{sec:conclusion}

\method{} fixes the hands, not the world model. In closed-loop tests, it more than doubles the success of a frozen planner when hands change and come back, costs 0.6 pp when they do not, and works unchanged on three world models and on two of three models pretrained on real robot data. The quality of the sensor sets the size of the gain: the frozen model keeps the physics, and a small map read through it keeps the interface.

\label{endofmain}

\section*{Acknowledgements}

We thank Nathan Duran (University of the West of England) for feedback on this work.

\clearpage
\appendix
\onecolumn

\section{Protocol}
\label{app:protocol}

\Cref{alg:hands} gives the loop of \method{} in one episode, and \cref{tab:constants} gives its constants. The constants of the library are the same in every cell. Each task calibrates only the surprise threshold $\theta$ and the gain $g$, on unshifted data.

\begin{algorithm}[ht]
  \caption{\method{} in one episode}
  \label{alg:hands}
  \begin{algorithmic}[1]
    \STATE {\bfseries Input:} frozen predictor $f$, library $\mathcal{L}$, switch rates $h, q$
    \STATE $(\hat M, \hat b, \hat d) \leftarrow$ map of the most probable source under the prior
    \REPEAT
      \STATE \textbf{Act:} plan with CEM through $(\hat M, \hat b, \hat d)$; execute blocks until a surprise or five blocks
      \STATE \textbf{Sense:} for each new block, read $s_t$ by \cref{eq:inversion} and its covariance by \cref{eq:laplace}
      \STATE \textbf{Explain:} score every source and model on the segment; cut the segment if the last four blocks favour another source
      \STATE $(\hat M, \hat b, \hat d) \leftarrow$ most probable model of the winning source
    \UNTIL{the goal is reached or the budget is spent}
    \STATE \textbf{Remember:} for each segment, create an entry or update the entries with responsibility $\geq 0.5$; update $h, q$
  \end{algorithmic}
\end{algorithm}

\begin{table}[ht]
  \centering
  \caption{Constants of \method{}.}
  \label{tab:constants}
  \small
  \begin{tabular}{@{}l>{\raggedright\arraybackslash}p{0.86\linewidth}@{}}
    \toprule
    Stage & Constants \\
    \midrule
    Sense & Adam, 100 steps, learning rate 0.1, nine starts: the command turned by $0\degr$, $\pm 45\degr$, $\pm 90\degr$ and $180\degr$, the mirrored command, the mirrored command turned by $180\degr$, and zero. A block counts if $\mathrm{err}_a < \theta$, or if $\mathrm{err}_a < 3\theta$ and $\mathrm{err}_a < 0.25\,\mathrm{err}_c$ ($\mathrm{err}_c$: the error of the command). Jacobian: central differences, step 0.01. $\tilde R_t = C_t / \bar r$, eigenvalues clipped to $[0.1, 1000]$; $\bar r$ is the median of $\operatorname{tr}(C)/2$ over the last 400 blocks. $g$: least squares on 1,200 nominal blocks. \\
    Explain & Class weights 1.0, 1.0, 0.3, 0.5 and 0.25 (identity, similarity, reflection, linear, affine); delay weight 1 for $d = 0$ and 0.1 for $d > 0$; prior 0.5 on the identity with $d = 0$. Noise prior $\sigma^2 \sim \mathrm{IG}(2, \sigma_0)$ with $\sigma_0 = 0.01\, s_2$, where $s_2$ is the expected squared norm of a command block sum. Map prior $\beta - \beta_0 \mid \sigma^2 \sim \mathcal{N}(0, \sigma^2 V_0)$ with $V_0 = \tau^2/\sigma_0$ and $\tau = 1$ (offset: $0.1^2 s_2/\sigma_0$). Clip model: at most 12 refits, tolerance 0.002. \\
    Remember & $h$ and $q$: Beta(1,1) posterior means over the last 20 episodes and the last 20 switches. Change point: the last 4 blocks, posterior above 0.6. New entry: $r_{\mathrm{new}} > 0.5$ and at least 4 blocks. Learning: entries with $r_k \geq 0.5$. Merge: log Bayes factor above 3. Use counts decay by 0.98 per episode. Removal: $n_k < 0.2$ and no win for 50 episodes, or no win within 20 episodes of creation. At most 64 blocks per entry and 16 entries. Floor: smallest singular value at least 0.2. \\
    Act & CEM: 300 samples, 30 iterations, 30 elites, initial standard deviation 1.0 (normalised), horizon 5 blocks, warm start. $\theta$: the 95th percentile of the frozen one-block error on 20 unshifted episodes (9.02 on PushT, 39.7 on Reacher, 59.0 on Two-Room, 272 on Cube, 0.100 on DINO-WM PushT). \\
    \bottomrule
  \end{tabular}
\end{table}

\paragraph{World models and tasks.} LeWM \citep{lewm} uses a ViT-tiny encoder, a 192-dimensional latent, a predictor with 6 layers and a history of 3 frames. One model step is a block of 5 environment steps. Each task starts from a random state of a random expert episode, and the goal is the state 25 environment steps later in that episode. Episode $i$ of seed $s$ has the same start, goal, random seeds and hand in every arm. Each episode has 50 environment steps (Two-Room: 25, where the planner has no headroom at 50). \Cref{tab:tasks} gives the tasks.

\begin{table}[ht]
  \centering
  \caption{Tasks, with the sensor gain $g$ and the sensor $R^2$ on nominal LeWM data.}
  \label{tab:tasks}
  \small
  \begin{tabular}{@{}llllrr@{}}
    \toprule
    Task & Raw action & Clip & Success & $g$ & $R^2$ \\
    \midrule
    PushT & 2-D pusher action & none & agent and block within 20 px, block angle within 20\degr & 0.940 & 0.98 \\
    Reacher & 2-D joint torques & $[-1, 1]$ & every joint within 0.05 rad of the target & 0.957 & 0.70 \\
    Two-Room & 2-D velocity & $[-1, 1]$ & agent within 16 px of the target & 0.967 & 0.90 \\
    Cube & 5-D end effector & none & cube within 4 cm of the target & 1.054 & 0.80 \\
    \bottomrule
  \end{tabular}
\end{table}

\paragraph{DINO-WM and JEPA-WM.} An adapter gives each model the interface of LeWM, and the method code does not change. The state is the vector of visual and proprioceptive features, scaled so that squared distance equals the planning cost of the model. Reacher and Two-Room have no public checkpoints, so we trained each model once on the data of LeWM, with the recipe of its authors and a 90/10 split by episode.
\label{app:proprio}
These models take the proprioceptive state as an input, so their high sensor $R^2$ could come from the joint state. We repeated each calibration with the inversion energy restricted to the visual part of the state. The reading was fixed in advance: $R^2 \geq 0.90$ keeps the claim that the model reads its actions from images. On PushT and Two-Room, the masked $R^2$ is 0.93--0.99, so the claim holds. On Reacher, it is 0.87--0.88, so part of the reading there comes from the joint state.

\paragraph{Further hands and schedules.} \Cref{tab:hands-main} gives the main hands, and \cref{tab:hands} gives the others. The parts of a hand act in a fixed order: latency, mirror, rotation, gain, dead band, saturation, offset, then the clip of the environment. \Cref{tab:schedules} gives the schedules. A \emph{return} is a switch to a hand that occurred earlier in the run. \emph{After returns} counts the switch episode and the next episode if the hand stays, and pools S1 and S2.

\begin{table}[ht]
  \centering
  \caption{Further hands. F and G change the world, not the interface.}
  \label{tab:hands}
  \small
  \begin{tabular}{@{}l>{\raggedright\arraybackslash}p{0.42\linewidth}>{\raggedright\arraybackslash}p{0.38\linewidth}@{}}
    \toprule
    Code & Definition & Real change with the same effect \\
    \midrule
    B, H & turn by $-90\degr$; turn by $180\degr$ & a moved camera; a camera on the other side \\
    M & mirror of $x$, then a turn by $90\degr$ & swapped and turned axes \\
    OA & turn by $90\degr$, then the offset of O & a moved camera and a bias \\
    Z, X & dead band; saturation (non-linear) & a worn or limited actuator \\
    ZA, XD & turn by $90\degr$, then dead band; turn by $37\degr$, then saturation & both of the above \\
    F, G & PushT friction; centre of gravity moved by 20 px & a new surface or object \\
    \bottomrule
  \end{tabular}
\end{table}

\begin{table}[ht]
  \centering
  \caption{Schedules. S1 to S4 are the main schedules.}
  \label{tab:schedules}
  \small
  \begin{tabular}{@{}l>{\raggedright\arraybackslash}p{0.74\linewidth}r@{}}
    \toprule
    Schedule & Content & Episodes \\
    \midrule
    S1 & A, C, A, C, D, A, D in blocks of 5 episodes, 4 times & 140 \\
    S2 & keep the last hand with $p = 0.6$; else a new random turn with $p = 0.1$, else uniform over A, C, D, E, N & 140 \\
    S3 & a new random turn in every episode & 140 \\
    S4 & N in every episode & 60 \\
    S5, S6 & as S1 and S2 with the non-linear hands & 140 \\
    S7, S8 & PushT: hands and world changes in blocks of 5 & 140, 160 \\
    S9 & as S1 with the offset hands: A, O, A, O, OA, A, OA & 140 \\
    K2, K4, K8 & keep with $p = 0.6$, else uniform over 2, 4 or 8 hands & 140 \\
    K16, K24 & keep with $p = 0.6$, else uniform over 12 turns and 4 or 12 mirrored turns & 400 \\
    M1 & with $p = 0.5$, the hand switches after 20 steps of an episode & 140 \\
    L1, L2, L3 & A throughout; B for 120 episodes, then N for 20; N for 70, then B for 70 & 140 \\
    G50--G200 & A for 5 episodes, N for $G$, then A for 10 & 65--215 \\
    R01--R45 & A and a turn of $90\degr{} + \delta$ in blocks of 5, 3 times, $\delta$ from $1\degr{}$ to $45\degr{}$ & 30 \\
    \bottomrule
  \end{tabular}
\end{table}

\paragraph{Arms.} The frozen planner runs CEM on the frozen model with the identity map. The true-hand arm plans through the full chain of the true hand. The earlier sensor ($^\dagger$) has no Laplace reliability, no clip model and a gain of 1. No memory$^\dagger$ identifies the hand in each episode, and one memory$^\dagger$ keeps one memory of 64 blocks. Fixed Share$^\dagger$ uses the same library with a motion gate and a prior of 0.8 on the last hand \citep{herbster1998fixedshare}. MBCD-style$^\dagger$ uses a CUSUM selector that switches at 3 nats \citep{mbcd,page1954cusum}. ShiftWM-style scores seven fixed hypotheses by exact evidence, with no memory \citep{shiftwm}. Same-sensor arms give these comparisons the full sensor of \method{}. Continual test-time training takes one Adam step (learning rate $10^{-4}$) on a copy of the action encoder after each transition \citep{jepattt}. The calibration arm adds a random excitation block at the start of every episode. More search uses 60 or 120 CEM iterations.

\paragraph{Statistics.} The unit is the run. For each run, we take the mean success of one arm minus another over the episodes of the measure. The bootstrap resamples the run-level differences 20,000 times, and the sign-flip test uses 20,000 sign vectors. Holm's correction covers the eight comparisons of the locked rules. All other rules are bars on the estimate or on the interval, with no correction for multiplicity. Development used 69 seeds. The locked test uses 20 further seeds, and the true-hand arm ran on 12 of them; paired comparisons with it use only these 12 seeds. Later tests use the locked-test seeds or fresh seeds; some of these sets reuse task lists on other schedules or arms. Reacher runs do not repeat bit for bit across GPU nodes because of the MuJoCo rendering, so all comparisons are paired within one set of runs. GPU ratios come from shared GPUs and are approximate.

\FloatBarrier
\section{Theory}
\label{app:theory}

We analyse \method{} as Bayesian regression of the readings on the commands. This appendix states the main results with proof sketches. All logarithms are natural.

\paragraph{Model.} Block $t$ covers $L = 5$ steps, and $c_k = 0$ for $k < 0$. A hand applies $u_k = \operatorname{clip}(M c_{k-d} + b)$. The commanded window sum is $\bar c_t^{(d)} = \sum_{i=0}^{L-1} c_{Lt+i-d}$. Given a model $h = (\text{class}, d)$, the reading is $s_t = X_t \beta + \varepsilon_t$ with $\varepsilon_t \sim \mathcal{N}(0, \sigma^2 \tilde R_t)$. With $Q$ the turn by $90\degr$ and $F = \mathrm{diag}(-1, 1)$, the designs are $[\bar c_t, Q \bar c_t]$ (similarity), $[F \bar c_t, QF \bar c_t]$ (reflection), $\bar c_t^\top \otimes I_2$ (linear) and $[\bar c_t^\top \otimes I_2, I_2]$ (affine). The prior is normal-inverse-gamma, so each model has a closed-form evidence $Z(\mathcal{D} \mid h)$ \citep{mackay1992bayesian}. A stored entry $\ell$ uses its own map posterior as the prior of the map. The prior over sources for episode $e$ is
\begin{equation}
  \pi_e(\text{last}) = 1 - \rho_e, \qquad \pi_e(\text{new}) = \rho_e \xi_e, \qquad \pi_e(\ell) = \rho_e (1 - \xi_e)\, \frac{U_\ell}{\sum_{j \neq \text{last}} U_j},
  \label{eq:th-prior}
\end{equation}
where $\rho_e$ and $\xi_e$ are the rates $h$ and $q$ of the main text, and the use counts follow $U_\ell \leftarrow \gamma (U_\ell + r_\ell)$ with $\gamma = 0.98$.

\paragraph{Assumptions.} \emph{A1:} no applied step reaches the clip, or the clip model is at its fixed point with $T = M$ and $\sigma_{\max}(M) \leq 1$. \emph{A2:} given the past and $\bar c_t$, $\E[s_t] = X_t \beta$ and $\operatorname{Cov}(s_t) = \sigma^2 \tilde R_t$. \emph{A3:} the errors are independent across blocks given the past, and commands may depend on past readings. \emph{A4:} in the Fisher statements, $\sigma^2$ is known. For tracking: \emph{T1:} the hand switches only between episodes. \emph{T2:} the comparator is a label oracle with the fits of the method, with loss $L'_{\text{label}}$. \emph{T3:} each earlier hand has exactly one entry, which holds all its past blocks and no other block. \emph{T4:} the prior over sources is fixed before the episode. On the logs, A2 holds for the typical block and fails in the tails (the kurtosis of the whitened residuals is 152 on PushT), and T3 holds only in part.

\begin{proposition}[Identifiability of the map]
\label{prop:th-map}
Fix the class and the delay, and let $\mathcal{I}_n = \sigma^{-2} \sum_{t \leq n} X_t^\top \tilde R_t^{-1} X_t$. Under A1--A4:
\begin{enumerate}[label=(\alph*)]
  \item $\beta$ is identifiable from $s_{1:n}$ if and only if $\mathcal{I}_n$ is nonsingular.
  \item One block with $\bar c_t \neq 0$ identifies a similarity or a reflection. A linear map needs two window sums that are not parallel, and an affine map three that are not on one line.
  \item The posterior and the evidence keep their form when $\bar c_t$ depends on $s_{1:t-1}$.
\end{enumerate}
\end{proposition}
\begin{proof}[Sketch]
The log-likelihood is quadratic in $\beta$ with Hessian $-\mathcal{I}_n$. Two values give the same distribution if and only if $X_t(\beta - \beta') = 0$ for all $t$, which is $(\beta - \beta')^\top \mathcal{I}_n (\beta - \beta') = 0$, because the eigenvalues of $\tilde R_t$ are bounded; the model is linear in $\beta$ \citep{rothenberg1971}. For (b), $[\bar c, Q\bar c]$ is $\lVert \bar c \rVert$ times an orthogonal matrix, and $(\bar c_t^\top, 1)$ must span $\R^3$ in the affine class. For (c), the joint density factorises into the block likelihoods and factors free of $\beta$ and $\sigma^2$ \citep{laiwei1982}.
\end{proof}

\begin{proposition}[Clipped actions]
\label{prop:th-clip}
Let $\tilde c_k = T^{-1}(\operatorname{clip}(T c_k + b) - b)$ be the effective commands of a transform $T$. (a) If $M$ is invertible, $\sigma_{\max}(M) \leq 1$ and $T = M$, the mean of a reading is exactly $M \tilde c_t^{(d)} + Lb$; if $\sigma_{\max}(M) > 1$, this fails on clipped steps. (b) An axis that is clipped at every step of a block gives no information about its row of $M$, and a gain above 1 is not identified from saturated commands. (c) $\tilde R_t$ depends only on the state and on $u_t^*$, so the reliability cannot down-weight a block because it was clipped.
\end{proposition}
\begin{proof}[Sketch]
(a) By definition, $T \tilde c_k + b = \operatorname{clip}(T c_k + b)$; sum over the block. (b) For a fixed clip pattern, the Jacobian of the mean in $\operatorname{vec}(M)$ is $\sum_k c_{k-d}^\top \otimes \Gamma_k$, where $\Gamma_k$ is the diagonal 0/1 matrix of the unclipped axes, so its row for a fully clipped axis is zero. (c) A clipped and an unclipped command with the same applied action give the same $J_t$.
\end{proof}

\begin{proposition}[Identification of the delay]
\label{prop:th-delay}
Under A1--A3: (a) for the same invertible map and offset, delays $d$ and $d'$ are indistinguishable on a set of blocks if and only if $\bar c_t^{(d)} = \bar c_t^{(d')}$ on each block, and in later blocks constant or $L$-periodic commands hide the delay; (b) a free map matches any readings on one block (similarity, reflection), two (linear) or three (affine) with window sums in general position, so these classes need at least two, three and four blocks to identify the delay; (c) with known $\sigma^2$ and fixed designs, the expected log-likelihood gap of a wrong delay is at most the reliability-weighted energy of the lag-$L$ command changes $\frac{1}{2\sigma^2} \sum_t \lVert M(\bar c_t^{(d)} - \bar c_t^{(d')}) \rVert^2_{\tilde R_t^{-1}}$.
\end{proposition}
\begin{proof}[Sketch]
A telescoping sum gives $\bar c_t^{(d)} - \bar c_t^{(d')} = \sum_{j=d}^{d'-1} (c_{Lt+L-1-j} - c_{Lt-1-j})$, which gives (a). For (b), the designs are invertible at a nonzero window sum (Proposition~\ref{prop:th-map}), so a map of the class reaches any target. For (c), take the projection onto the design of $d'$ in whitened coordinates, and bound the minimum over maps by the choice $M' = M$.
\end{proof}

\begin{proposition}[Regret of the library against a label oracle]
\label{prop:th-hard}
Under T1--T4, for every sequence of readings, the cumulative log loss $L_A$ of the predictive of the library satisfies
\begin{equation}
  L_A - L'_{\text{label}} \leq \sum_{e=2}^{E} \ln \frac{1}{\pi_e(z_e)},
  \label{eq:th-regret}
\end{equation}
where $\pi_e(z_e)$ is the prior mass of the source that holds the true hand. For a return after $G$ episodes to an entry with posterior mass $\bar r$ at its last visit, the use-count term $\ln(\sum_{j \neq \text{last}} U_j / U_{z_e})$ of \cref{eq:th-prior} is at most $\ln\frac{1}{1 - \gamma} + (G - 1) \ln\frac{1}{\gamma} + \ln\frac{1}{\bar r}$ nats, with $\ln\frac{1}{1 - \gamma} = 3.91$ and $\ln\frac{1}{\gamma} = 0.020$.
\end{proposition}
\begin{proof}[Sketch]
Under T1 and T4, the score is the exact Bayes mixture over sources, so its loss in episode $e$ is $-\ln \sum_\ell \pi_e(\ell) Z_\ell(\mathcal{D}_e) \leq -\ln \pi_e(z_e) - \ln Z_{z_e}(\mathcal{D}_e)$. Under T2 and T3, the last term is the predictive of the oracle. For the return share, $U_{z_e} \geq \gamma^G \bar r$, and each episode adds a mass of at most 1, so $\sum_j U_j \leq \gamma/(1 - \gamma)$.
\end{proof}

This applies the analysis of Mixing Past Posteriors \citep{bousquet2002mpp} to a hard-assignment library: the comparator pays the identification cost of each hand once, while a method that restarts from the prior pays it at every return. Fixed Share over the entries \citep{herbster1998fixedshare} has the same form, with the cost $\ln(K_e/\alpha)$ at each switch. Proposition~\ref{prop:th-hard} holds exactly for a simplified library, with no capacity limit, merges, pruning or window test. It bounds the log loss of the readings, not the success of the planner.

\begin{corollary}[When the selection rule cannot matter]
\label{cor:th-tie}
Let two libraries share the sensor, the fits, the planner and their entries at the start of an episode, and differ only in the prior over sources. (a) Before the first reading, each plans with the mode of its prior, which is the last source when $\rho_e < 1/2$ (Fixed Share: when $\alpha < K_e/(K_e + 1)$). Both plan the first block of a switch with the previous hand. (b) Let $\ell^*$ have the largest evidence, $\mathcal{E}$ its log evidence margin over the other sources, and $\Lambda_\pi = \max_{\ell \neq \ell^*} \ln(\pi(\ell)/\pi(\ell^*))$ its switch cost. If $\mathcal{E}$ exceeds both switch costs, both libraries select $\ell^*$. (c) On the same randomness, the two libraries give the planner the same map until a block violates (b). A CUSUM rule with threshold $c$ satisfies the same statements, with the switch cost $c$ on its first informed block.
\end{corollary}
\begin{proof}[Sketch]
A library selects $\arg\max_\ell [\ln \pi(\ell) + \ln Z_\ell]$. For $\ell \neq \ell^*$, the difference is at least $\mathcal{E} - \Lambda_\pi > 0$. Part (c) follows by induction over the blocks, because the map depends only on the entry and the readings.
\end{proof}

On our logs, the switch costs are a few nats (median 2.31 for the sticky prior, 2.5 for Fixed Share with three other sources, 3 for CUSUM), while the first reading of a return separates the hands by much more (tertile cuts 11 and 75 nats in the Gaussian scale). The corollary thus predicts a tie. With the same sensor, the sticky prior ties with Fixed Share ($-0.9$ [$-2.7$, $+0.9$]) and is 2.7 pp [0.9, 4.5] above CUSUM, so the prediction holds for Fixed Share but not fully for CUSUM.

\begin{lemma}[False alarms on the first block]
\label{lem:th-noise}
Let the identity be true, with Gaussian errors. Compare it with the similarity class on one block, with $\tilde R_1 = r I$ and map prior precision $\lambda_0 I$, and let $\psi = \lVert s_1 - \bar c_1 \rVert^2 / r$ and $\kappa = \lVert \bar c_1 \rVert^2 / (r \lambda_0)$. Then
\begin{equation}
  \ln \mathrm{BF} = (a_0 + 1) \ln \frac{b_0 + \frac{1}{2}\psi}{b_0 + \frac{1}{2}\psi / (1 + \kappa)} - \ln(1 + \kappa).
  \label{eq:th-first}
\end{equation}
For prior odds $O$ and $A = ((1 + \kappa) O)^{1/(a_0 + 1)} < 1 + \kappa$, the similarity model wins with probability $\exp(-x^* b_0 / \sigma^2)$, where $x^* = (A - 1) / (1 - A/(1 + \kappa))$. This grows with $\sigma^2 / b_0$ and tends to 1, so a noise prior far below the real noise causes false alarms.
\end{lemma}
\begin{proof}[Sketch]
The whitened design is $(\lVert \bar c_1 \rVert^2 / r)^{1/2}$ times an orthogonal matrix, so the posterior and prior precisions satisfy $\det P_1 / \det P_0 = (1 + \kappa)^2$. The Woodbury identity gives $\mathrm{RSS}_2 = \psi / (1 + \kappa)$, and $\mathrm{RSS}_1 = \psi$. Under A2, $\psi / 2\sigma^2$ is a standard exponential variable, which gives the probability.
\end{proof}

\paragraph{Checks on the logs.} A script fixed the criteria of \cref{tab:th-checks} before its first run. The checks use runs outside the locked test. Of 24 criteria, 16 hold on the final method (14 without the floor). The failures give three corrections. The noise prior sets the false-alarm rate of the first blocks (P5a, which led to Lemma~\ref{lem:th-noise}; this explanation is post hoc). The sensor errors are heavy-tailed, so the Fisher statements hold as orderings, not as calibrated numbers (P2a, P2b). The reliability also acts through contact, not only through the clip (P1d). We do not claim new regret bounds, the convergence of the fixed-point iteration of the clip model, or any bound on the success of the planner.

\begin{table}[ht]
  \centering
  \caption{Measurable predictions and their checks (PushT / Reacher / Two-Room where per task).}
  \label{tab:th-checks}
  \small
  \begin{tabular}{@{}l>{\raggedright\arraybackslash}p{0.47\linewidth}>{\raggedright\arraybackslash}p{0.3\linewidth}cc@{}}
    \toprule
    ID & Prediction & Result & Held & Failed \\
    \midrule
    P1a & The size of $\tilde R_t$ predicts the residual: slope in $[0.5, 1.5]$ & 1.19 / 0.92 / 0.97 & 3 & 0 \\
    P1b & The direction of $\tilde R_t$ is informative & Reacher: ratio 1.93 & 1 & 0 \\
    P1c & The calibrated gain is $1 \pm 0.03$ through the origin & 0.740 / 1.018 / 1.042 & 1 & 2 \\
    P1d & $\tilde R_t$ does not flag clipped blocks & trace ratio 1.01 (Reacher), 10.0 (Two-Room) & 0 & 2 \\
    P2a & Angle error falls as $\mathcal{I}_\theta^{-1/2}$: slope in $[-0.75, -0.25]$ & $-0.20$ / $-0.42$ / $-0.58$ & 2 & 1 \\
    P2b & Posterior scale calibrated: median $|z|$ in $[0.45, 1.0]$ & 0.96 / 1.32 / 0.65 & 2 & 1 \\
    P3a & One block does not identify the delay: at most 0.2 right & 0.17 (370 plans) & 1 & 0 \\
    P3b & The right delay rises with lag-5 energy: at least +0.2 & 0.23 / 0.45 / 0.66 by tertile & 1 & 0 \\
    P4a & Returns are right sooner than first visits: at least +0.10 & 0.71 against 0.36 & 1 & 0 \\
    P4b & Stored maps are more precise than fresh maps & error 0.029 against 0.094 & 1 & 0 \\
    P4c & Recognition fails where the first block separates weakly & 0.36 / 0.38 / 0.13 by tertile & 1 & 0 \\
    P4d & The selection rule ties: difference at most 0.05 & 0.49 against 0.49 & 1 & 0 \\
    P5a & False alarms of a fresh fit do not depend on the noise scale & 0.147 against 0.325 at twice the noise & 0 & 1 \\
    P5b & No change: false alarms at most the Gaussian model + 0.03 & 0.119 against 0.147 & 1 & 0 \\
    P5c & The world-change excess comes from dependence in episodes & real 0.152, shuffled 0.151 & 0 & 1 \\
    \midrule
    Total & & & 16 & 8 \\
    \bottomrule
  \end{tabular}
\end{table}

\FloatBarrier
\section{Additional Results}
\label{app:results}

All values are means over runs, and differences are in pp. Square brackets give the 95\% percentile bootstrap CI of the mean over runs, not the range of the runs.

\subsection{The locked test and the ablation per task}

\begin{table}[ht]
  \centering
  \caption{Locked test: \method{} minus each arm, per task (20 seeds; true hand: the 12 seeds on which it ran, paired). Runs per pooled cell: 120 after returns, 60 on S4 (oracle: 72 and 36). Pooled: mean with 95\% CI. SD: standard deviation of the pooled per-run differences. $^\dagger$Earlier sensor.}
  \label{tab:lock-detail}
  \small
  \begin{tabular}{@{}llrrrrr@{}}
    \toprule
    \method{} minus & Measure & PushT & Reacher & Two-Room & Pooled & SD \\
    \midrule
    Frozen planner & after returns & $+44.5$ & $+38.8$ & $+18.2$ & $+33.9$ [31.1, 36.6] & 15.5 \\
    Frozen planner & S4 & $-0.8$ & $-0.8$ & $-0.1$ & $-0.6$ [$-0.9$, $-0.2$] & 1.4 \\
    ShiftWM-style & after returns & $+29.6$ & $+29.1$ & $+15.1$ & $+24.6$ [22.6, 26.6] & 11.3 \\
    No memory$^\dagger$ & after returns & $+15.2$ & $+25.9$ & $+8.9$ & $+16.7$ [14.6, 18.7] & 11.6 \\
    One memory$^\dagger$ & after returns & $+14.0$ & $+23.3$ & $+9.1$ & $+15.5$ [13.6, 17.4] & 10.7 \\
    MBCD-style$^\dagger$ & after returns & $+2.4$ & $+14.4$ & $+8.4$ & $+8.4$ [6.7, 10.0] & 9.2 \\
    Fixed Share$^\dagger$ & after returns & $-0.8$ & $+8.7$ & $+8.7$ & $+5.5$ [4.1, 7.0] & 8.1 \\
    True hand & after returns & $-19.0$ & $-16.2$ & $-14.9$ & $-16.7$ [$-18.2$, $-15.2$] & 6.5 \\
    \bottomrule
  \end{tabular}
\end{table}

\begin{table}[ht]
  \centering
  \caption{Success for each hand (S1 and S2; 20 seeds; true hand: 12 of them). $^\dagger$Earlier sensor.}
  \label{tab:byhand}
  \small
  \begin{tabular}{@{}lrrrrrr@{}}
    \toprule
    Arm & A ($90\degr$) & C (gain, delay) & D ($37\degr$) & E (mirror) & N (no change) & T (new turn) \\
    \midrule
    Frozen planner & 7.8 & 34.5 & 31.4 & 9.1 & 77.5 & 28.6 \\
    One memory$^\dagger$ & 41.4 & 51.6 & 54.2 & 36.6 & 74.0 & 49.4 \\
    Fixed Share$^\dagger$ & 60.5 & 56.6 & 62.9 & 54.6 & 63.1 & 51.8 \\
    \textbf{\method{}} & \textbf{66.1} & \textbf{64.6} & \textbf{69.1} & \textbf{59.7} & \textbf{68.2} & \textbf{57.0} \\
    True hand & 77.9 & 72.6 & 80.0 & 78.3 & 76.7 & 76.1 \\
    \bottomrule
  \end{tabular}
\end{table}

\begin{table}[ht]
  \centering
  \caption{The ablation per task, after returns (8 seeds; 16 runs per cell). The first two rows compare two sensors without the floor; the other rows use the floor on both sides. On 12 further seeds (72 runs each), \method{} minus one model is $+8.3$ [6.5, 10.1], and \method{} minus no delay on hand C is $+21.4$ [18.1, 24.8].}
  \label{tab:factorial-detail}
  \small
  \begin{tabular}{@{}llrrr@{}}
    \toprule
    Stage & Comparison & PushT & Reacher & Two-Room \\
    \midrule
    Sense & Full minus earlier sensor (sticky prior) & $+2.5$ [$-0.8$, 5.4] & $+8.6$ [5.3, 11.9] & $+6.6$ [3.5, 9.5] \\
    Sense & Full minus earlier sensor (Fixed Share) & $+1.4$ [$-1.3$, 4.1] & $+7.9$ [3.5, 12.5] & $+9.3$ [6.0, 13.1] \\
    Explain & \method{} minus one model (linear, no delay) & $+12.7$ [8.8, 16.4] & $+11.2$ [7.7, 14.3] & $+3.5$ [0.5, 6.6] \\
    Explain & \method{} minus five classes with no delay & $+11.4$ [7.8, 15.0] & $+5.0$ [0.3, 9.7] & $+1.8$ [$-0.7$, 4.5] \\
    Remember & \method{} minus no memory & $+19.3$ [15.3, 23.6] & $+22.9$ [18.7, 27.2] & $+7.6$ [3.3, 12.1] \\
    Remember & \method{} minus one memory & $+13.1$ [8.9, 17.1] & $+15.6$ [12.2, 18.9] & $+8.3$ [5.8, 10.8] \\
    Rule & \method{} minus Fixed Share & $-0.1$ [$-3.2$, 2.6] & $-3.2$ [$-6.9$, 0.5] & $+0.8$ [$-1.2$, 2.7] \\
    Rule & \method{} minus CUSUM & $+2.6$ [$-0.4$, 5.6] & $+4.6$ [0.6, 8.2] & $+0.8$ [$-1.2$, 3.0] \\
    Gap & Recognition oracle minus \method{} & $+17.7$ [14.5, 21.1] & $+14.1$ [10.2, 18.3] & $+14.0$ [11.0, 17.0] \\
    \bottomrule
  \end{tabular}
\end{table}

\FloatBarrier
\subsection{Further tests}

\Cref{tab:further} gives the tests that \cref{sec:more} and \cref{sec:discussion} summarise. Unless a row says otherwise, differences are \method{} minus the named arm, and \cref{app:prereg} gives the number of runs of each test.

\begin{table}[!ht]
\centering
\caption{Further tests on LeWM, unless a row names another model.}
\label{tab:further}
\small
\setlength{\tabcolsep}{4pt}
\begin{tabular}{@{}>{\raggedright\arraybackslash}p{0.2\linewidth}>{\raggedright\arraybackslash}p{0.77\linewidth}@{}}
  \toprule
  Test & Result \\
  \midrule
  Offset hand (S9) & The offset costs the frozen planner 41.8 pp [30.1, 52.7] on hand O. \method{} minus frozen on O: $+23.4$ [14.0, 32.6] (56\% of the gap); after returns: $+37.1$ [32.0, 42.4]. Fresh seeds: $+42.2$, $+21.8$, $+34.0$. On Two-Room, the offset costs only 3.1 pp, and \method{} is 6.2 [3.1, 9.4] below the frozen planner, mostly in the first episode after hand A (post hoc). \\
  Offset hand on DINO-WM and JEPA-WM & Partial: the grid reached its time limit with 15 and 13 of 18 complete \method{} runs, so we give no verdict. On these runs, after returns: $+60.8$ [52.5, 69.2] and $+58.7$ [49.0, 68.3]; on O: $+22.7$ [4.7, 40.0] and $+22.3$ [4.6, 38.5]. On Two-Room, \method{} again loses on O ($-20.0$ [$-34.0$, $-4.0$] and $-16.0$ [$-24.0$, $-10.0$]). \\
  One hand (L1) & \method{} minus the true hand, episodes 5--139: $-0.3$ [$-1.9$, $+1.2$]; Two-Room $-0.3$ [$-0.9$, $+0.4$]. The library ends with one entry. \\
  Long stay (L2) & First five episodes on N after 120 episodes of B: $-10.0$ [$-15.0$, $-3.8$]; Two-Room $-12.5$ [$-22.5$, $-2.5$]. On PushT and Reacher, the whole loss is the first episode on N (37.5\% against 81.2\%). \\
  One switch (L3) & After a switch from N to B: $+76.6$ [73.3, 80.0]; Two-Room $+55.4$ [50.7, 60.0]. Before the switch, the cost is 0.7 pp. \\
  2, 4, 8 hands & All episodes (PushT and Reacher, 32 runs per cell): \method{} 76.0, 71.3, 67.6; Fixed Share 75.2, 71.5, 67.2; one memory 71.7, 66.5, 60.7; frozen 45.3, 27.1, 20.4. The library ends with 2.4, 4.3 and 8.6 entries. \\
  16, 24 hands & After returns, 400 episodes: \method{} 57.5, 53.0; Fixed Share 58.1, 52.9; one memory 49.4, 48.7; frozen 15.8, 11.6; true hand 82.5, 82.8. \method{} minus one memory, both pools: $+6.1$ [4.5, 7.8]; fresh seeds $+6.2$ [4.3, 8.2]; Two-Room $+6.4$ [5.0, 8.0]. \method{} minus Fixed Share, 24 hands: $0.0$ [$-1.5$, $+1.6$]; fresh seeds $+1.4$ [$-1.0$, $+3.8$]; Two-Room $+0.6$ [$-1.6$, $+3.0$]. \\
  Long runs & S2 over 400 episodes (8 runs): 64.6, 73.1, 71.8 and 70.4\% per 100 episodes; frozen 29.4, 29.8, 26.6 and 28.9\%. Episodes 300--399 minus 100--199: $-2.8$ [$-8.5$, $+3.1$]. \\
  Forgetting & A returns after $G = 50$, 100 or 200 episodes of N. The first informed plan uses the entry of A in 11, 10 and 0 of 16 runs. Minus one memory, first 5 episodes back, $G \leq 100$: $+3.8$ [$-5.6$, $+13.1$]; Two-Room $+7.5$ [0.0, $+17.5$]. \\
  Near hands & A and a turn of $90\degr{} + \delta$. True hand minus \method{}, $\delta \leq 5\degr{}$: $+3.5$ [0.6, 6.2]; fresh seeds $+3.2$ [0.6, 6.0]; Two-Room $+2.5$ [0.6, 4.4]. The library treats two hands as one up to about $20\degr{}$. \\
  Non-linear hands & Dead bands and saturations (S5 and S6, 12 seeds). Minus frozen, pooled: $+15.9$ [14.0, 17.7]. \\
  Budgets & Budgets of 25, 50 and 100 steps (Two-Room: 15, 25, 50). Minus frozen: $+11.4$ [8.5, 14.4], $+35.1$ [30.7, 39.6], $+38.7$ [33.9, 43.3]. Minus one memory: $+4.6$ [3.0, 6.3], $+12.3$ [10.2, 14.5], $+5.3$ [3.1, 7.4]. \\
  First-frame cue & A context image before the first action enters the recall prior. After returns: $+6.3$ [4.5, 8.3] on 12 seeds, and $+6.0$ [4.0, 8.2] on the 8 seeds of the ablation; fresh seeds: $+8.3$ [6.0, 10.5]. On new hands only, it costs $-3.3$ [$-4.5$, $-2.0$], and $-2.4$ [$-4.2$, $-0.6$] on fresh seeds. It is not part of \method{}. \\
  Switch inside an episode & Minus frozen, switch episodes (M1): $+6.6$ [4.2, 9.4], SD 6.6 over 24 runs; fresh seeds $+9.9$ [8.0, 11.8]; pooled $+8.3$ [6.6, 10.0]. \\
  Return to no change & Hand N inside S2, minus frozen: $-9.4$ [$-11.4$, $-7.2$] (post hoc). \\
  World changes & Friction and centre of gravity. S7, minus frozen: $-5.0$ [$-7.1$, $-3.2$]. S8, world change after no change: $-0.8$ [$-2.9$, $+1.3$]. \\
  Cube & After returns, minus frozen: $+4.6$ [2.8, 6.4]. No change: $+1.1$ [$-0.3$, $+2.4$]. S1--S3: frozen 37.1, \method{} 43.1, true hand 49.1. \\
  \bottomrule
\end{tabular}
\end{table}

\begin{table}[!ht]
  \centering
  \caption{The sensor-noise sweep, after returns: gain over the frozen planner, and the value of the library $V$ (\method{} minus no memory, same noise). LeWM: 16 runs per cell (8 seeds). DINO-WM: 6 runs, S1.}
  \label{tab:sweep}
  \small
  \begin{tabular}{@{}lrrrrrrr@{}}
    \toprule
    & \multicolumn{2}{c}{PushT} & \multicolumn{2}{c}{Reacher} & \multicolumn{2}{c}{Two-Room} & DINO-WM \\
    \cmidrule(lr){2-3} \cmidrule(lr){4-5} \cmidrule(lr){6-7} \cmidrule(lr){8-8}
    $\sigma$ & gain & $V$ & gain & $V$ & gain & $V$ & gain \\
    \midrule
    0 & $+47.6$ & $+19.3$ & $+39.7$ & $+22.9$ & $+18.2$ & $+7.6$ & $+58.3$ \\
    0.3 & $+43.7$ & $+22.4$ & $+33.7$ & $+21.7$ & $+13.6$ & $+8.1$ & -- \\
    0.6 & $+38.8$ & $+27.0$ & $+24.2$ & $+15.7$ & $+7.4$ & $+5.2$ & $+54.2$ \\
    1.0 & $+28.8$ & $+23.8$ & $+12.6$ & $+8.5$ & $+3.4$ & $+2.2$ & $+45.8$ \\
    1.5 & $+18.8$ & $+17.0$ & $+5.1$ & $+1.9$ & $+0.1$ & $+2.2$ & $+29.2$ \\
    2.5 & $+1.9$ & $+3.1$ & $+0.9$ & $+1.5$ & $-3.2$ & $-2.2$ & $+10.4$ \\
    \bottomrule
  \end{tabular}
\end{table}

\FloatBarrier
\subsection{Ablation: the floor}
\label{app:floor}

The floor makes a map admissible only if the smallest singular value of its linear part is at least 0.2, a value that we chose before any run (the weakest test hand has gain 0.5). The planner uses only admissible maps, and a segment whose best explanation is not admissible is not stored. \Cref{tab:floor} compares \method{} with and without the floor. On the locked test, the floor changes success by $+0.4$ pp [$-0.1$, $+0.9$] over S1--S3, $+0.1$ [$-0.9$, $+1.1$] after returns and $+0.1$ [$-0.3$, $+0.4$] with no change. It matters in long runs. In the 500-episode streams of the JEPA-TTT protocol on PushT with no change, the AUC of \method{} without the floor falls to 52.9, against 83.3 for the frozen planner. In a failed episode, some blocks show commands with almost no visible effect, so the library stores a map with a gain near zero (all such maps were below 0.08). A planner that uses this map sees almost no effect of any command, so CEM inflates the commands and the pusher leaves the image. The sensor then reads a constant, which confirms the map. With the floor, the AUC is 82.5.

\begin{table}[ht]
  \centering
  \caption{\method{} with and without the floor. Locked test (20 seeds): success (\%) over S1--S3, after returns and with no change (S4). 500-episode streams of the JEPA-TTT protocol on PushT (three seeds): AUC of held-out success with no change and under the action rotation.}
  \label{tab:floor}
  \small
  \begin{tabular}{@{}lrrrrr@{}}
    \toprule
    & \multicolumn{3}{c}{Locked test} & \multicolumn{2}{c}{Streams (AUC)} \\
    \cmidrule(lr){2-4} \cmidrule(lr){5-6}
    Arm & S1--S3 & After returns & S4 & No change & Rotation \\
    \midrule
    \method{} & 62.3 & 60.4 & 77.5 & 82.5 & 78.1 \\
    \method{} without the floor & 61.9 & 60.3 & 77.4 & 52.9 & 64.5 \\
    Frozen planner & 28.6 & 26.6 & 78.1 & 83.3 & 2.0 \\
    \bottomrule
  \end{tabular}
\end{table}

Against a floor of 0.2, floors of 0.1, 0.3 and 0.4 change success after returns by $-1.7$ [$-3.2$, $-0.3$], $+0.7$ [$-1.0$, $+2.5$] and $+0.2$ pp [$-1.1$, $+1.6$] (post hoc).

\FloatBarrier
\subsection{Comparison with test-time training}
\label{app:jepattt}

We ran our re-implementations of JEPA-TTT \citep{jepattt} and AdaJEPA \citep{adajepa} on LeWM, on the protocol and the eight shifts of JEPA-TTT (\cref{tab:jepattt}). For each shift and seed, a stream of 500 episodes runs under one shift, and every 50 episodes, 100 held-out episodes start from a copy of each method. Best is the largest held-out success after at least 50 episodes, and AUC is the normalised area under the curve. The maps of \method{} act on $(x, y)$, so it does not run on the two Cube shifts, which also change $z$.

\begin{table}[ht]
  \centering
  \caption{The JEPA-TTT protocol (three seeds). Held-out success (\%), Best / AUC; one value where the two are equal. No change: AUC with no shift.}
  \label{tab:jepattt}
  \small
  \begin{tabular}{@{}llrrrr@{}}
    \toprule
    Task & Shift & Frozen & AdaJEPA & JEPA-TTT & \method{} \\
    \midrule
    PushT & action rotation & 2.0 & 4.3 & 68.0 / 50.9 & $\mathbf{82.3}$ / $\mathbf{78.1}$ \\
     & contact rotation & 19.3 & 19.3 & 19.3 / 14.6 & 19.3 / 18.3 \\
    Two-Room & wave & 42.3 & 40.3 & $\mathbf{63.3}$ / $\mathbf{54.4}$ & 49.7 / 45.9 \\
     & grid & 46.0 & 48.3 & $\mathbf{67.0}$ / $\mathbf{59.5}$ & 49.0 / 45.2 \\
    Reacher & phase & 23.3 & 18.0 & 22.0 / 17.4 & $\mathbf{25.7}$ / $\mathbf{23.2}$ \\
     & harmonic & 20.7 & 20.3 & 22.3 / 17.7 & $\mathbf{23.0}$ / $\mathbf{20.3}$ \\
    Cube & crosstalk & 31.0 & 32.0 & 37.0 / 32.8 & -- \\
     & delayed & 18.7 & 18.3 & 41.0 / 32.7 & -- \\
    \midrule
    \multicolumn{2}{@{}l}{No change: PushT} & 83.3 & 79.3 & 73.3 & 82.5 \\
    \multicolumn{2}{@{}l}{No change: Two-Room} & 74.7 & 74.7 & 73.7 & 74.6 \\
    \multicolumn{2}{@{}l}{No change: Reacher} & 82.0 & 77.3 & 70.7 & 81.5 \\
    \multicolumn{2}{@{}l}{No change: Cube} & 53.0 & 53.0 & 49.9 & 52.0 \\
    \bottomrule
  \end{tabular}
\end{table}

\method{} wins when the command itself changes: a few blocks identify the map, while gradient steps on the weights need many episodes. It leads by 60.7 pp on the action rotation after 50 episodes. It loses on the two Two-Room shifts ($-8.5$ and $-14.3$ pp of AUC). There, the command turns by up to 75\degr{}, with an angle that is a wave in the position of the agent, so the mean turn is close to zero. One map per hand can follow only this mean, while JEPA-TTT changes the model itself.

\Cref{tab:competitors-detail} gives the learned and calibration methods on our schedules. After returns, the three methods that adapt weights stay near the frozen planner (26.1 to 26.6\%, against 26.1\%).

\begin{table}[ht]
  \centering
  \caption{Learned and calibration methods on our schedules (12 seeds; 72 runs after returns, 36 on S4).}
  \label{tab:competitors-detail}
  \small
  \begin{tabular}{@{}lrrrr@{}}
    \toprule
    Arm & After returns & S4 & \method{} minus arm, after returns & Arm minus frozen, S4 \\
    \midrule
    Continual test-time training & 26.1 & 74.3 & $+34.1$ [31.3, 36.9] & $-3.8$ [$-5.6$, $-2.0$] \\
    AdaJEPA (reset each episode) & 26.6 & 76.5 & $+33.7$ [30.2, 37.2] & $-1.6$ [$-3.2$, 0.0] \\
    JEPA-TTT (persistent) & 26.4 & 71.4 & $+33.8$ [30.0, 37.6] & $-6.7$ [$-8.7$, $-4.9$] \\
    Calibration in every episode & 41.8 & 64.7 & $+18.4$ [15.7, 21.2] & $-13.4$ [$-16.7$, $-10.2$] \\
    \method{} & 60.3 & 77.7 & -- & $-0.4$ \\
    Frozen planner & 26.1 & 78.1 & $+34.2$ & -- \\
    \bottomrule
  \end{tabular}
\end{table}

\paragraph{Compute.} GPU time per environment step relative to the frozen planner, on the locked runs (PushT, Reacher, Two-Room), is 0.82, 0.78 and 0.82 for \method{}, which makes 2.57, 2.44 and 1.54 plans per episode against 5.45, 5.29 and 3.25. With two and four times the CEM iterations, the frozen planner costs 1.91--2.03 and 3.59--4.18 times as much per step. Per plan, \method{} costs 1.26 to 1.36 times a frozen plan (measured without the floor, which does not change the cost of a plan); it is cheaper per step because it replans about half as often.

\FloatBarrier
\subsection{RoboCasa: world models trained on real robot data}
\label{app:robocasa}

\paragraph{Set-up.} We use the three world models that \citet{jepawms} pretrained on DROID \citep{droid} and their RoboCasa \citep{robocasa} evaluation configurations, with the planner unchanged. The task is place in PnPCounterTop. Each episode has at most 60 planning steps, and one model action repeats for 5 environment steps. The hand acts on the $x$ and $y$ commands of the end-effector translation from the start of planning, so the set-up of the goal never sees it. The sensor inverts the planner's own rollout, and a calibration of 8 episodes with no hand gives its $R^2$, gain and surprise threshold.

\paragraph{Rules fixed in advance.} A gate admits a model and a task only if the frozen planner with no hand succeeds at least 15 pp more often than under a turn of 90\degr{}. It admits place for all three models. Reach fails the gate on DINO-WM and JEPA-WM, and the frozen \mbox{V-JEPA~2-AC} planner succeeds in none of 16 later episodes on it, so we do not run reach. Hand C does not lower the success of the frozen DINO-WM planner on place, so a rule fixed before the test replaces it with the first of B and E that costs 15 pp: E on DINO-WM and \mbox{V-JEPA~2-AC}, and B on JEPA-WM. The rules are P0, sensor $R^2$ at least 0.5; P1, at least $+10$ over the frozen planner after returns on S1; and P2, at most 2 pp below it with no change. After returns are episodes 10, 11, 15, 16, 25, 26, 30 and 31. On \mbox{V-JEPA~2-AC} and JEPA-WM, four seeds were added after the reading of the first four, under a rule fixed before they ran that the 8-seed verdict replaces the first. On \mbox{V-JEPA~2-AC}, the first four seeds gave $+28.1$ [9.4, 43.8]; on the four added seeds, the frozen planner, \method{} and the true hand all succeed in 25.0\%, so these scenes leave no headroom.

\begin{table}[ht]
  \centering\small
  \caption{Results in RoboCasa, place. Success after returns on S1 in \%. Differences with 95\% bootstrap intervals; seeds in parentheses.}
  \label{tab:rc-results}
  \resizebox{\textwidth}{!}{%
  \begin{tabular}{@{}lccc@{}}
    \toprule
    & DINO-WM & JEPA-WM & \mbox{V-JEPA~2-AC} \\
    \midrule
    Frozen encoder & DINOv2 ViT-S/14 & DINOv3 ViT-L & V-JEPA~2 ViT-g \\
    Calibration: $R^2$ / gain / turn of the map & 0.80 / 1.08 / 39.3\degr{} & 0.70 / 0.73 / 22.3\degr{} & 0.68 / 0.95 / 35.2\degr{} \\
    S1 schedule & A, E, A, E, D, A, D & A, B, A, B, D, A, D & A, E, A, E, D, A, D \\
    Frozen / true hand / \method{} / one memory & 7.8 / 35.9 / 26.6 / 29.7 & 14.1 / 35.9 / 21.9 / 25.0 & 21.9 / 35.9 / 35.9 / 31.2 \\
    P0: sensor $R^2$ (bar 0.5) & pass & pass & pass \\
    P1: \method{} $-$ frozen (bar $+10$) & $+18.8$ [$+4.7$, $+34.4$] (8), pass & $+7.8$ [$+3.1$, $+10.9$] (8), fail & $+14.1$ [$-3.1$, $+29.7$] (8), pass \\
    P2: S4, \method{} $-$ frozen (bar $-2$) & $+7.5$ [$-1.2$, $+16.2$] (4), pass & $-6.2$ [$-10.0$, $-2.5$] (4), fail & $+8.8$ [$+2.5$, $+13.8$] (4), pass \\
    Share of the gap closed & 0.67 & 0.36 & 1.00 \\
    \method{} $-$ one memory & $-3.1$ [$-9.4$, $+4.7$] (8) & $-3.1$ [$-10.9$, $+4.7$] (8) & $+4.7$ [$-7.8$, $+15.6$] (8) \\
    All 35 episodes: frozen / true hand / \method{} & 9.3 / 30.4 / 30.7 & 13.9 / 31.8 / 22.5 & 19.3 / 35.4 / 35.7 \\
    S2: \method{} $-$ frozen (bar $+10$) & $+11.4$ [$+5.3$, $+17.1$] (6), pass & not run & not run \\
    \bottomrule
  \end{tabular}}
\end{table}

\paragraph{The frame of the commands.} Each calibration map is turned by 22 to 39\degr{}: the models read the RoboCasa commands in a turned frame. The sensor removes only the gain, so with no hand the library learns this turn as a hand. On S4, \method{} plans with the identity in only 0--5\% of its informed plans. This helps a little on DINO-WM and \mbox{V-JEPA~2-AC} and hurts on JEPA-WM, which agrees with the failure of P2 there but does not settle its cause: the calibration of JEPA-WM varies between runs (gain 0.73 and 0.94). The planner told the true hand removes only the injected hand, while \method{} removes both the hand and this turn, so it can equal that planner (post hoc).

\paragraph{Recognition.} At the first informed plan of a return, \method{} chooses the entry of the returning hand in 24, 17 and 21 of 64 returns on DINO-WM, JEPA-WM and \mbox{V-JEPA~2-AC}. It mostly learns a returning hand again. A fresh fit is confident by the second or third of 12 plans here, so recall saves little, and the lead over one memory is not significant on any model (post hoc).

\begin{figure}[ht]
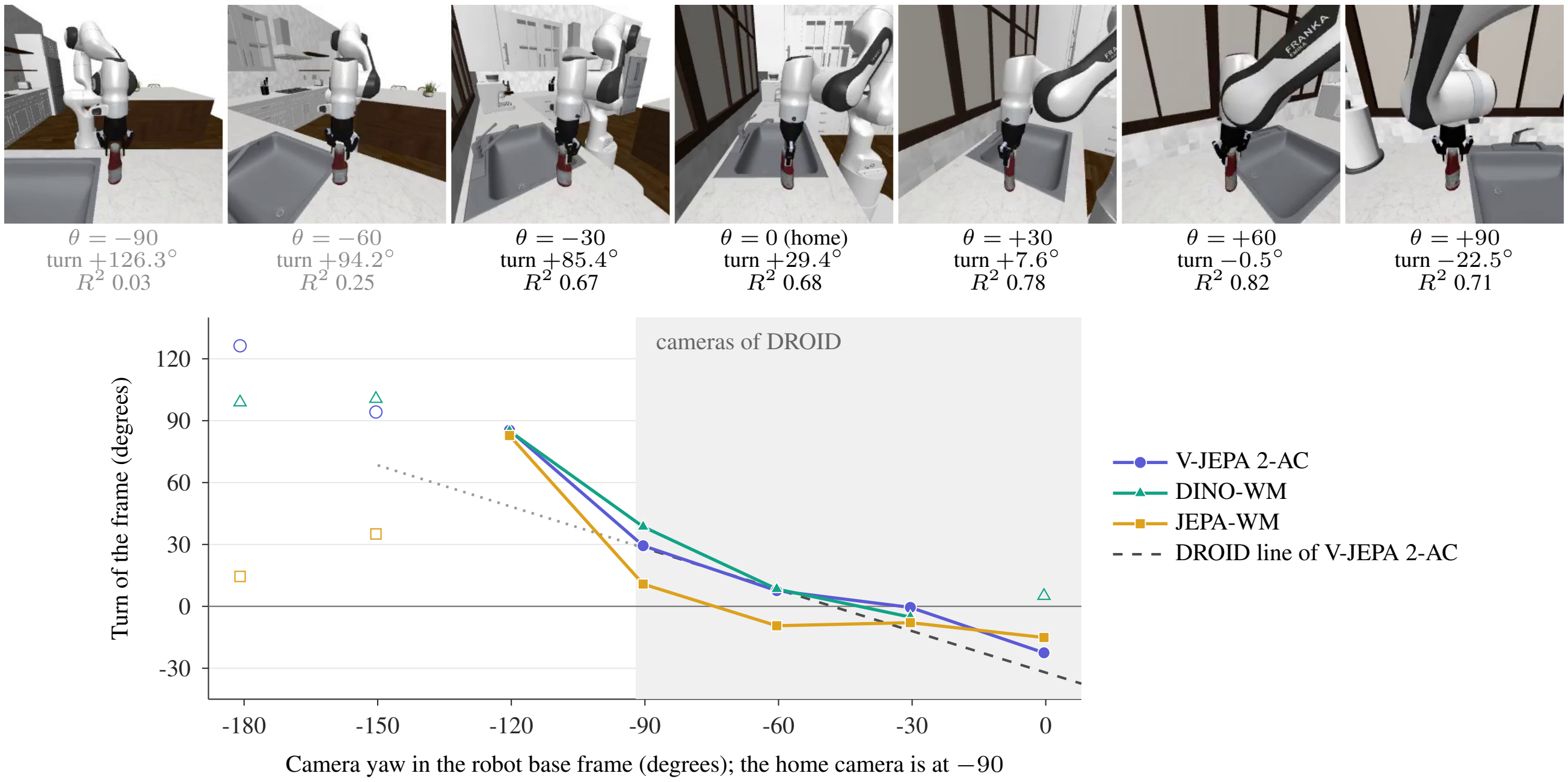

  \centering

  \caption{\textbf{The frame of each model turns with the camera.} (a)~The view of \mbox{V-JEPA~2-AC} at each camera yaw $\theta$ (degrees), with the turn of the frame in which the model reads the commands and the $R^2$ of the sensor (grey: below 0.5). (b)~RoboCasa place, 8 calibration episodes per yaw, no injected hand. Filled marks: $R^2 \geq 0.5$. Shaded: the camera yaws of the 261 DROID views. Dashed: the line that the frame of \mbox{V-JEPA~2-AC} follows on DROID, fixed before the runs.}
  \label{fig:rc-camera}
\end{figure}

\paragraph{A moved camera.} We moved the planner's camera by a yaw $\theta$ from $-90\degr{}$ to $+90\degr{}$ in steps of $30\degr{}$, with no injected hand and the goal rendered by the same camera. The rules were fixed before the runs. K0: the sensor $R^2$ is at least 0.5. K1: the slope of the turn on $\theta$ lies in $[-1.2, -0.3]$, with an interval that excludes 0. K1-DROID: this slope overlaps the slope of \mbox{V-JEPA~2-AC} on 261 DROID views, $-0.67$ [$-0.81$, $-0.56$]. K1 and K1-DROID pass on DINO-WM ($-0.68$ [$-0.87$, $-0.47$]) and \mbox{V-JEPA~2-AC} ($-0.85$ [$-1.02$, $-0.65$]). K1 fails on JEPA-WM ($-0.32$ [$-1.01$, $+0.17$]). Behind the robot, outside the DROID camera range, the sensor fails on all three models (\cref{fig:rc-camera}). A second stage measured the cost of the camera at the yaws where K0 passes. It is at least 15 pp (K2) only on \mbox{V-JEPA~2-AC}, and on the other two models a camera turned toward the DROID frame helps the frozen planner. We have not yet tested \method{} with cameras that change and come back.

\begin{figure}[htbp]
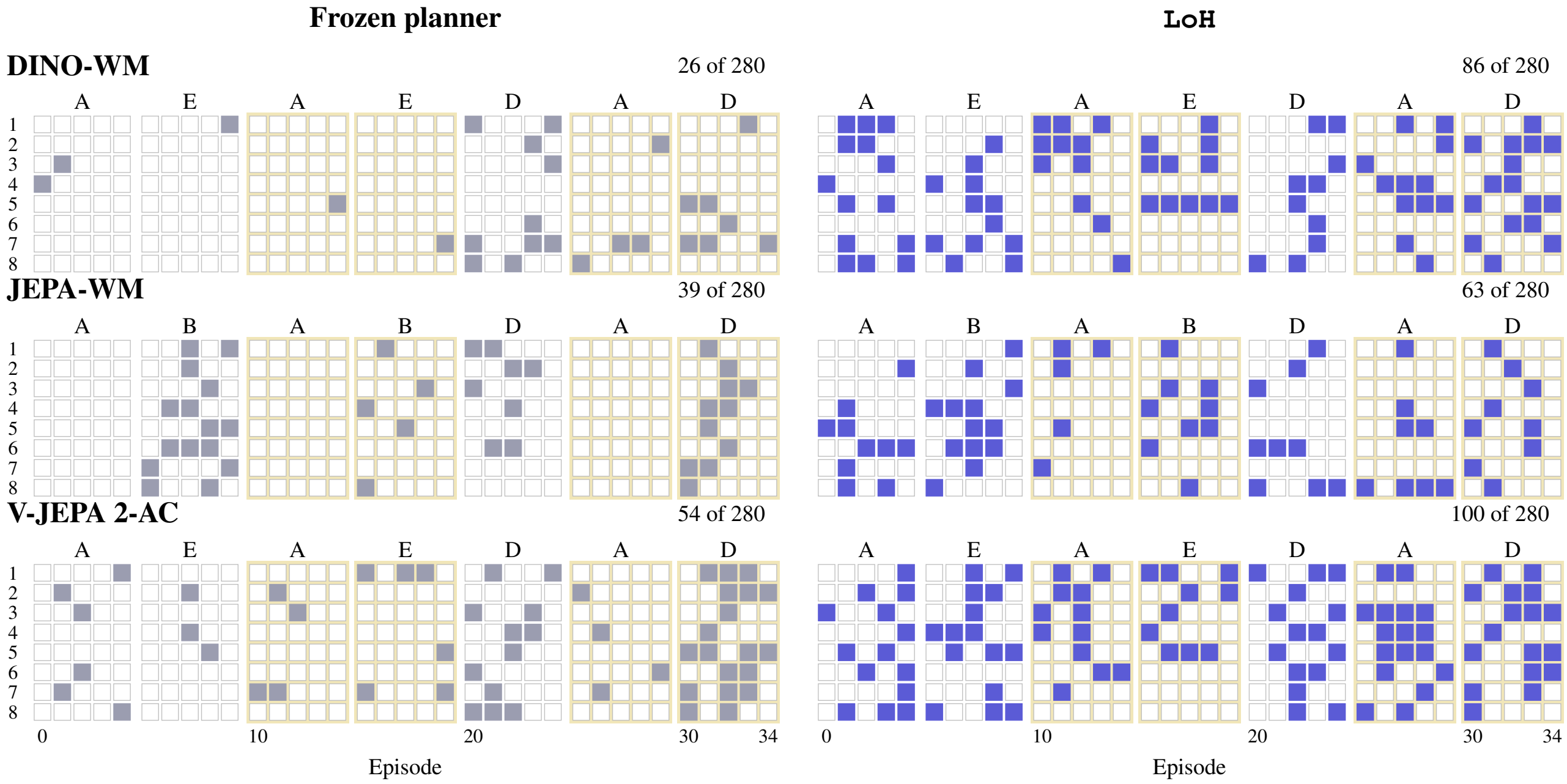

  \centering
%
  \caption{Every S1 episode of the three models on place (rows: the 8 runs; columns: episodes 0--34; filled: success; left: frozen planner, right: \method{}). The letters give the hand of each block of five episodes, and shaded blocks are returns of a known hand. In each row, both planners start every episode from the same scene and goal.}
  \label{fig:rc-grid}
\end{figure}

\clearpage
\section{Pre-registered Rules and Development History}
\label{app:prereg}

\paragraph{Development history.} The method reached its final form in two steps. First, we adopted \method{} without the floor, although its adoption rule failed on one of three criteria: the criterion of no loss under changes of the world failed on both seed sets. We adopted it because this loss has a separate cause (\cref{sec:discussion}), and froze it before the locked test. Second, after the locked test, the 500-episode streams of the JEPA-TTT protocol showed a failure of this version (\cref{app:floor}), and we added the floor. Its pilot used 3 of the 20 locked-test seeds, and its confirmation used fresh seeds. On the 17 locked-test seeds that the pilot did not use, the locked rules give the same verdicts. Every result of \method{} in the paper uses the final method, with the floor, unless it says otherwise; \cref{tab:floor} gives \method{} without the floor. Two other decisions did not follow a rule. A latent check to stop the world-change loss was designed, but a calibration showed that it could not separate false maps from new hands after one block, so we did not run its test. A calibrated noise prior was not adopted, because its gain did not replicate on fresh seeds (EBc1, EBc4).

\paragraph{Rules.} We wrote each rule below, with its bar, before the runs that it judges. \Cref{tab:prereg} lists every rule on the final method, including the rules that fail. A bar applies to the estimate unless it says ``CI''. ``Ours'' is \method{}. $n$ counts runs, or seeds where marked s, or episodes where marked ep. TR is Two-Room. The rules on \method{} without the floor give 47 passes, 10 failures and 4 rules that we did not run; the table repeats those whose verdict could change.

{\footnotesize
\setlength{\tabcolsep}{3pt}
\begin{longtable}{@{}l>{\raggedright\arraybackslash}p{0.27\linewidth}lrr@{\enspace}ll@{}}
  \caption{Pre-registered rules on the final method. $^\dagger$Earlier sensor.}
  \label{tab:prereg} \\
  \toprule
  Rule & Measure & Bar & Value & & $n$ & Verdict \\
  \midrule
  \endfirsthead
  \toprule
  Rule & Measure & Bar & Value & & $n$ & Verdict \\
  \midrule
  \endhead
  \bottomrule
  \endfoot
  \multicolumn{7}{@{}l}{\emph{Locked test, 20 seeds}} \\
  L1 & ours minus frozen, after returns & $\geq +10$ & $+33.9$ & [31.1, 36.6] & 120 & pass \\
  L2 & ours minus four baselines$^\dagger$, after returns & $> 0$, CI & $+5.5$ to $+24.6$ & lower $\geq +4.1$ & 120 & pass \\
  L3 & ours minus frozen, S4 & $\geq -1$ & $-0.6$ & [$-0.9$, $-0.2$] & 60 & pass \\
  L4 & ours minus ShiftWM-style, S1--S3 & $\geq +5$ & $+22.6$ & [20.6, 24.7] & 180 & pass \\
  L5 & ours minus no memory$^\dagger$, S1--S3 & $\geq +5$ & $+15.8$ & [14.1, 17.5] & 180 & pass \\
  L1--L5 & the same on the 17 seeds that the floor pilot did not use & as above & L1 $+33.2$ & & 17 s & pass (all) \\
  \midrule
  \multicolumn{7}{@{}l}{\emph{The floor: pilot, then fresh seeds}} \\
  FP1, FC1 & PushT, no change, AUC: ours minus frozen & $\geq -2$ & $-0.8$, $-0.2$ & & 3 s each & pass \\
  FP2, FC2 & PushT, rotation, Best: ours minus JEPA-TTT & $\geq +10$ & $+14.3$, $+18.7$ & & 3 s each & pass \\
  FP3, FC3 & our schedules: with minus without the floor & $\geq -1$ & $+0.6$, $-0.1$ & & 45, 81 & pass \\
  FP4 & S4, each task: with minus without the floor & $\geq -2$ & $0.0$ to $+0.6$ & & 3 s each & pass \\
  FC4 & S4, each task: floor minus frozen & $\geq -2$ & $-0.6$ to $-0.4$ & & 9 s each & pass \\
  \midrule
  \multicolumn{7}{@{}l}{\emph{JEPA-TTT protocol}} \\
  T1 & action rotation, Best: ours minus JEPA-TTT & $\geq +10$ & $+14.3$ & & 3 s & pass \\
  T2 & mean AUC over six shifts: ours minus JEPA-TTT & $\geq 0$ & $+2.8$ & & 18 & pass \\
  T3 & held-out success after 50 episodes: ours minus JEPA-TTT & $\geq 0$ & $+13.9$ & & 18 & pass \\
  T4 & mean Best over eight shifts: JEPA-TTT minus frozen & $\geq +5$ & $+17.1$ & & 24 & pass \\
  \midrule
  \multicolumn{7}{@{}l}{\emph{Other world models}} \\
  DW1, JW1 & six cells, S1: ours minus frozen, after returns & $\geq +10$ & $+31.3$ to $+64.6$ & lower $\geq +18.8$ & 6 s each & pass (6 of 6) \\
  DW2, JW2 & six cells, S4: ours minus frozen & $\geq -2$ & $-2.5$ to $+0.8$ & & 6 s each & fail (5 of 6) \\
  DW3 & five cells, S2: ours minus frozen, after returns & $\geq +10$ & $+26.3$ to $+63.1$ & lower $\geq +16.4$ & 6--10 s & pass (5 of 5) \\
  \midrule
  \multicolumn{7}{@{}l}{\emph{Ablation}} \\
  F1a, F1b & full minus earlier sensor, sticky prior; Fixed Share (no floor) & $\geq +3$ & $+5.9$, $+6.2$ & & 48 & pass \\
  F2 & ours minus no memory, same sensor, after returns & $\geq +5$ & $+16.6$ & [13.5, 19.7] & 48 & pass \\
  X1 & ours minus one model (linear, no delay), after returns & $> 0$, CI & $+9.1$ & [6.9, 11.4] & 48 & pass \\
  & \quad 12 further seeds & & $+8.3$ & [6.5, 10.1] & 72 & pass \\
  X2 & ours minus five classes with no delay, hand C & $> 0$, CI & $+25.1$ & [20.6, 29.6] & 48 & pass \\
  & \quad 12 further seeds & & $+21.4$ & [18.1, 24.8] & 72 & pass \\
  \midrule
  \multicolumn{7}{@{}l}{\emph{Offset hand (S9)}} \\
  OG & true hand minus frozen, hand O (fresh: second value) & $\geq +5$ & $+41.8$, $+42.2$ & & 24, 36 & pass \\
  OF1 & ours minus frozen, after returns (fresh: second value) & $\geq +10$ & $+37.1$, $+34.0$ & & 24, 36 & pass \\
  OF2 & ours minus frozen, hand O (fresh: second value) & $> 0$, CI & $+23.4$, $+21.8$ & lower $\geq +14.0$ & 24, 36 & pass \\
  O-wm & the same on DINO-WM and JEPA-WM (rules need 18 runs) & as above & \cref{tab:further} & & 15, 13 & not read \\
  \midrule
  \multicolumn{7}{@{}l}{\emph{Long stays}} \\
  LS1 & one hand: ours minus true hand, episodes 5--139 & $\geq -5$ & $-0.3$; TR $-0.3$ & & 16; 8 & pass \\
  LS2 & long stay, then N: ours minus frozen, first 5 on N & $\geq -3$ & $-10.0$; TR $-12.5$ & & 16; 8 & fail \\
  LS3 & no change, then B: ours minus frozen, after the switch & $> 0$, CI & $+76.6$; TR $+55.4$ & & 16; 8 & pass \\
  \midrule
  \multicolumn{7}{@{}l}{\emph{Many hands, forgetting, near hands}} \\
  MK1 & (ours minus one memory$^\dagger$) with 8 hands minus with 2 & $> 0$, CI & $+5.8$ & [2.0, 9.6] & 16 & pass \\
  MK1-fair & the same with the same sensor & $> 0$, CI & $+3.8$ & [1.3, 6.3] & 16 & pass \\
  MK2 & ours minus Fixed Share$^\dagger$, 8 hands & $\geq 0$ & $+3.9$ & [0.2, 7.7] & 16 & pass \\
  MH1 & K16 and K24: ours minus one memory, after returns & $> 0$, CI & $+6.1$ & [4.5, 7.8] & 32 & pass \\
  & \quad 8 fresh seeds & & $+6.2$ & [4.3, 8.2] & 32 & pass \\
  & \quad TR & & $+6.4$ & [5.0, 8.0] & 16 & pass \\
  MH2 & K24: ours minus Fixed Share, after returns & $> 0$, CI & $0.0$ & [$-1.5$, $+1.6$] & 16 & fail \\
  & \quad 8 fresh seeds & & $+1.4$ & [$-1.0$, $+3.8$] & 16 & fail \\
  & \quad TR & & $+0.6$ & [$-1.6$, $+3.0$] & 8 & fail \\
  LR1 & ours, episodes 300--399 minus 100--199 & $\geq -3$ & $-2.8$ & [$-8.5$, $+3.1$] & 8 & pass \\
  FG1 & return after 50 or 100: ours minus one memory, first 5 & $> 0$, CI & $+3.8$; TR $+7.5$ & lower $\leq 0$ & 32; 16 & fail \\
  RS1 & near hand, $\delta \leq 5\degr{}$: true hand minus ours & $\leq 5$ & $+3.5$, $+3.2$; TR $+2.5$ & & 48; 24 & pass \\
  \midrule
  \multicolumn{7}{@{}l}{\emph{Other tests}} \\
  MS1 & switch inside an episode: ours minus frozen & $\geq +10$ & $+6.6$ & [4.2, 9.4] & 24 & fail \\
  MS2 & switch inside an episode: ours minus no memory & $\geq -2$ & $+5.7$ & [3.7, 7.7] & 24 & pass \\
  CX1 & with minus without the cue, after returns & $> 0$, CI & $+6.3$ & [4.5, 8.3] & 72 & pass \\
  CX2a & the cue, misleading cues, all episodes & $\geq -3$ & $+0.4$ & [$-0.4$, $+1.3$] & 72 & pass \\
  CX2b & the cue, one constant cue, all episodes & $\geq -2$ & $+0.8$ & [$-0.3$, $+1.9$] & 36 & pass \\
  CX2c & the cue, new hands only, all episodes & $\geq -2$ & $-3.3$ & [$-4.5$, $-2.0$] & 36 & fail \\
  CB1 & Cube: ours minus frozen, after returns & $\geq +10$ & $+4.6$ & [2.8, 6.4] & 24 & fail \\
  CB2 & Cube: ours minus ShiftWM-style, S1--S3 & $\geq +5$ & $+4.0$ & [2.5, 5.6] & 36 & fail \\
  CB3 & Cube: ours minus frozen, S4 & $\geq -1$ & $+1.1$ & [$-0.3$, $+2.4$] & 12 & pass \\
  B1 & budget 100: ours minus one memory, after returns & $\geq +5$ & $+5.3$ & [3.1, 7.4] & 48 & pass \\
  BU1 & (ours minus one memory) at budget 25 minus at 100 & $> 0$, CI & $-0.7$ & [$-2.7$, $+1.3$] & 48 & fail \\
  SQ1 & library value at $\sigma = 0.6$ minus at 0, PushT and Reacher & $> 0$, CI & $+0.2$ & [$-3.9$, $+4.4$] & 32 & fail \\
  SQ2 & gain at $\sigma = 1.0$ minus at $2.5$, PushT and Reacher & $\geq +10$ & $+19.3$ & [15.6, 23.0] & 32 & pass \\
  SQ3 & Two-Room: gain at $\sigma = 0$ minus at $2.5$ & $> 0$, CI & $+21.4$ & [17.8, 25.0] & 16 & pass \\
  SQ4 & DINO-WM: gain at $\sigma = 0$ minus at $2.5$ & $> 0$, CI & $+47.9$ & [35.4, 58.3] & 6 s & pass \\
  EB1--EB3 & calibrated noise prior: S4; after returns; S3 & $-0.2$; $-1$; $-2$ & $-0.19$; $+1.44$; $+0.46$ & & 36; 72; 36 & pass \\
  EBc1 & the same on 12 fresh seeds: minus frozen, S4 & $\geq -0.2$ & $-0.23$ & [$-0.42$, $-0.05$] & 36 & fail \\
  EBc2, EBc3 & fresh seeds: minus ours, after returns; S3 & $-1$; $-2$ & $-0.13$; $+0.93$ & & 72; 36 & pass \\
  EBc4 & fresh seeds: minus ours, after returns & $> 0$, CI & $-0.13$ & [$-1.62$, $+1.35$] & 72 & fail \\
  \midrule
  \multicolumn{7}{@{}l}{\emph{RoboCasa, place: DINO-WM, JEPA-WM, \mbox{V-JEPA~2-AC} (\cref{tab:rc-results})}} \\
  Gate & frozen planner, no hand minus a turn of 90\degr{} & $\geq +15$ & $+21.9$ to $+46.9$ & & 16--32 ep & pass (3 of 3) \\
  P0 & sensor $R^2$ of the calibration & $\geq 0.5$ & 0.68 to 0.80 & & 8 ep & pass (3 of 3) \\
  P1 & S1: ours minus frozen, after returns (JEPA-WM $+7.8$) & $\geq +10$ & $+7.8$ to $+18.8$ & & 8 s each & fail (2 of 3) \\
  P2 & S4: ours minus frozen (JEPA-WM $-6.2$) & $\geq -2$ & $-6.2$ to $+8.8$ & & 4 s each & fail (2 of 3) \\
  P1-S2 & DINO-WM, S2: ours minus frozen, after returns & $\geq +10$ & $+11.4$ & [5.3, 17.1] & 6 s & pass \\
  K0 & camera sweep: sensor $R^2$ at each yaw & $\geq 0.5$ & 14 of 21 & & 8 ep & fail (14 of 21) \\
  K1 & slope of the turn on the yaw (JEPA-WM $-0.32$; bar: in $[-1.2, -0.3]$, CI) & in range & $-0.85$ to $-0.32$ & & 7 yaws & fail (2 of 3) \\
  K1-DROID & the K1 interval overlaps the DROID slope & overlap & as K1 & & 7 yaws & pass (3 of 3) \\
  K2 & frozen planner, home minus moved camera, largest drop & $\geq +15$ & $-6.3$ to $+25.0$ & & 16 ep & fail (1 of 3) \\
  K3 & moved camera minus a hand of the same frame error & $\geq -10$ & $-19.9$ to $-3.4$ & & 3--4 yaws & fail (2 of 3) \\
  K4 & at home: planner told its frame minus frozen & $> 0$ & $-12.5$ to $+18.8$ & & 16 ep & fail (2 of 3) \\
  K5 & mean of 3 models: frozen at $+30\degr{}$, $+60\degr{}$ minus at home & $> 0$ & $+9.9$ & & 3 models & pass \\
\end{longtable}
}


\begin{thebibliography}{48}
\providecommand{\natexlab}[1]{#1}
\providecommand{\url}[1]{\texttt{#1}}
\expandafter\ifx\csname urlstyle\endcsname\relax
  \providecommand{\doi}[1]{doi: #1}\else
  \providecommand{\doi}{doi: \begingroup \urlstyle{rm}\Url}\fi

\bibitem[Adams \& MacKay(2007)Adams and MacKay]{adams2007bocpd}
Adams, R.~P. and MacKay, D. J.~C.
\newblock {Bayesian} online changepoint detection.
\newblock \emph{arXiv preprint arXiv:0710.3742}, 2007.

\bibitem[Alegre et~al.(2021)Alegre, Bazzan, and da~Silva]{mbcd}
Alegre, L.~N., Bazzan, A. L.~C., and da~Silva, B.~C.
\newblock Minimum-delay adaptation in non-stationary reinforcement learning via
  online high-confidence change-point detection.
\newblock In \emph{Proceedings of the 20th International Conference on
  Autonomous Agents and Multiagent Systems (AAMAS)}, pp.\  97--105, 2021.

\bibitem[{Anonymous}(2026{\natexlab{a}})]{fssl}
{Anonymous}.
\newblock First visits and revisits: Learning from the deployment stream in
  model-based control.
\newblock Under review at the International Conference on Learning
  Representations (ICLR 2027), 2026{\natexlab{a}}.
\newblock URL \url{https://openreview.net/forum?id=Jv0qc3b9nq}.

\bibitem[{Anonymous}(2026{\natexlab{b}})]{shiftwm}
{Anonymous}.
\newblock {ShiftWM-Bench}: Planning with frozen latent world models under
  hidden within-episode dynamics shifts.
\newblock Under review at the International Conference on Learning
  Representations (ICLR 2027), 2026{\natexlab{b}}.
\newblock URL \url{https://openreview.net/forum?id=qu42pOd1Gy}.

\bibitem[Assran et~al.(2025)Assran, Bardes, Fan, Garrido, Howes, Komeili,
  Muckley, Rizvi, Roberts, Sinha, Zholus, Arnaud, Gejji, Martin, Hogan, Dugas,
  Bojanowski, Khalidov, Labatut, Massa, Szafraniec, Krishnakumar, Li, Ma,
  Chandar, Meier, LeCun, Rabbat, and Ballas]{vjepa2}
Assran, M., Bardes, A., Fan, D., Garrido, Q., Howes, R., Komeili, M., Muckley,
  M., Rizvi, A., Roberts, C., Sinha, K., Zholus, A., Arnaud, S., Gejji, A.,
  Martin, A., Hogan, F.~R., Dugas, D., Bojanowski, P., Khalidov, V., Labatut,
  P., Massa, F., Szafraniec, M., Krishnakumar, K., Li, Y., Ma, X., Chandar, S.,
  Meier, F., LeCun, Y., Rabbat, M., and Ballas, N.
\newblock {V-JEPA} 2: Self-supervised video models enable understanding,
  prediction and planning.
\newblock \emph{arXiv preprint arXiv:2506.09985}, 2025.

\bibitem[Athans et~al.(1977)Athans, Castanon, Dunn, Greene, Lee, Sandell, and
  Willsky]{athans1977mmac}
Athans, M., Castanon, D., Dunn, K., Greene, C., Lee, W., Sandell, N., and
  Willsky, A.
\newblock The stochastic control of the {F-8C} aircraft using a multiple model
  adaptive control ({MMAC}) method--{Part I}: Equilibrium flight.
\newblock \emph{IEEE Transactions on Automatic Control}, 22\penalty0
  (5):\penalty0 768--780, 1977.
\newblock \doi{10.1109/TAC.1977.1101599}.

\bibitem[Bousquet \& Warmuth(2002)Bousquet and Warmuth]{bousquet2002mpp}
Bousquet, O. and Warmuth, M.~K.
\newblock Tracking a small set of experts by mixing past posteriors.
\newblock \emph{Journal of Machine Learning Research}, 3:\penalty0 363--396,
  2002.

\bibitem[Brito \& McNamee(2025)Brito and McNamee]{rwm}
Brito, C.~S. and McNamee, D.
\newblock World models as reference trajectories for rapid motor adaptation.
\newblock \emph{arXiv preprint arXiv:2505.15589}, 2025.

\bibitem[Chi et~al.(2023)Chi, Feng, Du, Xu, Cousineau, Burchfiel, and
  Song]{diffusionpolicy}
Chi, C., Feng, S., Du, Y., Xu, Z., Cousineau, E., Burchfiel, B., and Song, S.
\newblock Diffusion policy: Visuomotor policy learning via action diffusion.
\newblock In \emph{Robotics: Science and Systems}, 2023.
\newblock \doi{10.15607/RSS.2023.XIX.026}.

\bibitem[Florence et~al.(2021)Florence, Lynch, Zeng, Ramirez, Wahid, Downs,
  Wong, Lee, Mordatch, and Tompson]{ibc}
Florence, P., Lynch, C., Zeng, A., Ramirez, O.~A., Wahid, A., Downs, L., Wong,
  A., Lee, J., Mordatch, I., and Tompson, J.
\newblock Implicit behavioral cloning.
\newblock In \emph{Conference on Robot Learning}, volume 164 of
  \emph{Proceedings of Machine Learning Research}, pp.\  158--168. PMLR, 2021.

\bibitem[Garrido et~al.(2026)Garrido, Nagarajan, Terver, Ballas, LeCun, and
  Rabbat]{garrido}
Garrido, Q., Nagarajan, T., Terver, B., Ballas, N., LeCun, Y., and Rabbat, M.
\newblock Learning latent action world models in the wild.
\newblock In \emph{Proceedings of the 43rd International Conference on Machine
  Learning}, 2026.

\bibitem[Hanna \& Stone(2017)Hanna and Stone]{gat}
Hanna, J.~P. and Stone, P.
\newblock Grounded action transformation for robot learning in simulation.
\newblock In \emph{Proceedings of the AAAI Conference on Artificial
  Intelligence}, volume~31, 2017.
\newblock \doi{10.1609/aaai.v31i1.11044}.

\bibitem[Heald et~al.(2021)Heald, Lengyel, and Wolpert]{coin}
Heald, J.~B., Lengyel, M., and Wolpert, D.~M.
\newblock Contextual inference underlies the learning of sensorimotor
  repertoires.
\newblock \emph{Nature}, 600\penalty0 (7889):\penalty0 489--493, 2021.
\newblock \doi{10.1038/s41586-021-04129-3}.

\bibitem[Herbster \& Warmuth(1998)Herbster and Warmuth]{herbster1998fixedshare}
Herbster, M. and Warmuth, M.~K.
\newblock Tracking the best expert.
\newblock \emph{Machine Learning}, 32\penalty0 (2):\penalty0 151--178, 1998.
\newblock \doi{10.1023/A:1007424614876}.

\bibitem[Hovakimyan \& Cao(2010)Hovakimyan and Cao]{l1book}
Hovakimyan, N. and Cao, C.
\newblock \emph{{$\mathcal{L}_1$} Adaptive Control Theory: Guaranteed
  Robustness with Fast Adaptation}.
\newblock Society for Industrial and Applied Mathematics, 2010.
\newblock \doi{10.1137/1.9780898719376}.

\bibitem[Jang et~al.(2026)Jang, Sung, Joung, Chae, Lee, Kim, Kim, and
  Lee]{axisguide}
Jang, J., Sung, Y., Joung, W., Chae, D., Lee, S., Kim, S., Kim, J., and Lee, J.
\newblock {AxisGuide}: Grounding robot action coordinate system in {RGB}
  observations for robust visuomotor manipulation.
\newblock In \emph{Robotics: Science and Systems}, 2026.
\newblock \doi{10.15607/RSS.2026.XXII.125}.

\bibitem[Kass \& Raftery(1995)Kass and Raftery]{kass1995bayes}
Kass, R.~E. and Raftery, A.~E.
\newblock {Bayes} factors.
\newblock \emph{Journal of the American Statistical Association}, 90\penalty0
  (430):\penalty0 773--795, 1995.
\newblock \doi{10.1080/01621459.1995.10476572}.

\bibitem[Khazatsky et~al.(2024)Khazatsky, Pertsch, Nair, Balakrishna, Dasari,
  Karamcheti, Nasiriany, Srirama, Chen, Ellis, Fagan, Hejna, Itkina, Lepert,
  Ma, Miller, Wu, Belkhale, Dass, Ha, Jain, Lee, Lee, Memmel, Park,
  Radosavovic, Wang, Zhan, Black, Chi, Hatch, Lin, Lu, Mercat, Rehman, Sanketi,
  Sharma, Simpson, Vuong, Walke, Wulfe, Xiao, Yang, Yavary, Zhao, Agia, Baijal,
  Guaman~Castro, Chen, Chen, Chung, Drake, Foster, Gao, Herrera, Heo, Hsu, Hu,
  Jackson, Le, Li, Lin, Ma, Maddukuri, Mirchandani, Morton, Nguyen, O'Neill,
  Scalise, Seale, Son, Tian, Tran, Wang, Wu, Xie, Yang, Yin, Zhang, Bastani,
  Berseth, Bohg, Goldberg, Gupta, Gupta, Jayaraman, Lim, Malik,
  Mart{\'\i}n-Mart{\'\i}n, Ramamoorthy, Sadigh, Song, Wu, Yip, Zhu, Kollar,
  Levine, and Finn]{droid}
Khazatsky, A., Pertsch, K., Nair, S., Balakrishna, A., Dasari, S., Karamcheti,
  S., Nasiriany, S., Srirama, M.~K., Chen, L.~Y., Ellis, K., Fagan, P.~D.,
  Hejna, J., Itkina, M., Lepert, M., Ma, Y.~J., Miller, P.~T., Wu, J.,
  Belkhale, S., Dass, S., Ha, H., Jain, A., Lee, A., Lee, Y., Memmel, M., Park,
  S., Radosavovic, I., Wang, K., Zhan, A., Black, K., Chi, C., Hatch, K.~B.,
  Lin, S., Lu, J., Mercat, J., Rehman, A., Sanketi, P.~R., Sharma, A., Simpson,
  C., Vuong, Q., Walke, H.~R., Wulfe, B., Xiao, T., Yang, J.~H., Yavary, A.,
  Zhao, T.~Z., Agia, C., Baijal, R., Guaman~Castro, M., Chen, D., Chen, Q.,
  Chung, T., Drake, J., Foster, E.~P., Gao, J., Herrera, D.~A., Heo, M., Hsu,
  K., Hu, J., Jackson, D., Le, C., Li, Y., Lin, R., Ma, Z., Maddukuri, A.,
  Mirchandani, S., Morton, D., Nguyen, T., O'Neill, A., Scalise, R., Seale, D.,
  Son, V., Tian, S., Tran, E., Wang, A.~E., Wu, Y., Xie, A., Yang, J., Yin, P.,
  Zhang, Y., Bastani, O., Berseth, G., Bohg, J., Goldberg, K., Gupta, A.,
  Gupta, A., Jayaraman, D., Lim, J.~J., Malik, J., Mart{\'\i}n-Mart{\'\i}n, R.,
  Ramamoorthy, S., Sadigh, D., Song, S., Wu, J., Yip, M.~C., Zhu, Y., Kollar,
  T., Levine, S., and Finn, C.
\newblock {DROID}: A large-scale in-the-wild robot manipulation dataset.
\newblock In \emph{Robotics: Science and Systems}, 2024.
\newblock \doi{10.15607/RSS.2024.XX.120}.

\bibitem[Koolen et~al.(2012)Koolen, Adamskiy, and Warmuth]{koolen2012sleep}
Koolen, W.~M., Adamskiy, D., and Warmuth, M.~K.
\newblock Putting {Bayes} to sleep.
\newblock In \emph{Advances in Neural Information Processing Systems},
  volume~25, 2012.

\bibitem[Kumar et~al.(2021)Kumar, Fu, Pathak, and Malik]{rma}
Kumar, A., Fu, Z., Pathak, D., and Malik, J.
\newblock {RMA}: Rapid motor adaptation for legged robots.
\newblock In \emph{Robotics: Science and Systems}, 2021.
\newblock \doi{10.15607/RSS.2021.XVII.011}.

\bibitem[Lai \& Wei(1982)Lai and Wei]{laiwei1982}
Lai, T.~L. and Wei, C.~Z.
\newblock Least squares estimates in stochastic regression models with
  applications to identification and control of dynamic systems.
\newblock \emph{The Annals of Statistics}, 10\penalty0 (1):\penalty0 154--166,
  1982.
\newblock \doi{10.1214/aos/1176345697}.

\bibitem[Li et~al.(2026)Li, Geng, Mao, Li, and Held]{scout}
Li, Y., Geng, L., Mao, X., Li, A., and Held, D.
\newblock Scouting the dynamics gap: Test-time policy adaptation via
  action-outcome feedback.
\newblock In \emph{Conference on Robot Learning}, 2026.
\newblock arXiv:2609.36107.

\bibitem[Li \& Liu(2026)Li and Liu]{eca}
Li, Z. and Liu, T.
\newblock {ECA}: Reward-free online efference-copy adaptation.
\newblock Withdrawn submission to the International Conference on Learning
  Representations (ICLR 2027), 2026.
\newblock URL \url{https://openreview.net/forum?id=zfURMTVbj8}.

\bibitem[Liu et~al.(2023)Liu, Zhu, Gao, Feng, Liu, Zhu, and Stone]{libero}
Liu, B., Zhu, Y., Gao, C., Feng, Y., Liu, Q., Zhu, Y., and Stone, P.
\newblock {LIBERO}: Benchmarking knowledge transfer for lifelong robot
  learning.
\newblock In \emph{Advances in Neural Information Processing Systems},
  volume~36, pp.\  44776--44791, 2023.
\newblock \doi{10.52202/075280-1939}.

\bibitem[MacKay(1992{\natexlab{a}})]{mackay1992bayesian}
MacKay, D. J.~C.
\newblock {Bayesian} interpolation.
\newblock \emph{Neural Computation}, 4\penalty0 (3):\penalty0 415--447,
  1992{\natexlab{a}}.
\newblock \doi{10.1162/neco.1992.4.3.415}.

\bibitem[MacKay(1992{\natexlab{b}})]{mackay1992practical}
MacKay, D. J.~C.
\newblock A practical {Bayesian} framework for backpropagation networks.
\newblock \emph{Neural Computation}, 4\penalty0 (3):\penalty0 448--472,
  1992{\natexlab{b}}.
\newblock \doi{10.1162/neco.1992.4.3.448}.

\bibitem[Maes et~al.(2026)Maes, Le~Lidec, Scieur, LeCun, and Balestriero]{lewm}
Maes, L., Le~Lidec, Q., Scieur, D., LeCun, Y., and Balestriero, R.
\newblock {LeWorldModel}: Stable end-to-end joint-embedding predictive
  architecture from pixels.
\newblock \emph{arXiv preprint arXiv:2603.19312}, 2026.

\bibitem[Nagabandi et~al.(2019)Nagabandi, Finn, and Levine]{mole}
Nagabandi, A., Finn, C., and Levine, S.
\newblock Deep online learning via meta-learning: Continual adaptation for
  model-based {RL}.
\newblock In \emph{International Conference on Learning Representations}, 2019.

\bibitem[Nasiriany et~al.(2024)Nasiriany, Maddukuri, Zhang, Parikh, Lo, Joshi,
  Mandlekar, and Zhu]{robocasa}
Nasiriany, S., Maddukuri, A., Zhang, L., Parikh, A., Lo, A., Joshi, A.,
  Mandlekar, A., and Zhu, Y.
\newblock {RoboCasa}: Large-scale simulation of everyday tasks for generalist
  robots.
\newblock In \emph{Robotics: Science and Systems}, 2024.

\bibitem[Oquab et~al.(2024)Oquab, Darcet, Moutakanni, Vo, Szafraniec, Khalidov,
  Fernandez, Haziza, Massa, El-Nouby, Assran, Ballas, Galuba, Howes, Huang, Li,
  Misra, Rabbat, Sharma, Synnaeve, Xu, Jegou, Mairal, Labatut, Joulin, and
  Bojanowski]{oquab2024dinov2}
Oquab, M., Darcet, T., Moutakanni, T., Vo, H., Szafraniec, M., Khalidov, V.,
  Fernandez, P., Haziza, D., Massa, F., El-Nouby, A., Assran, M., Ballas, N.,
  Galuba, W., Howes, R., Huang, P.-Y., Li, S.-W., Misra, I., Rabbat, M.,
  Sharma, V., Synnaeve, G., Xu, H., Jegou, H., Mairal, J., Labatut, P.,
  Joulin, A., and Bojanowski, P.
\newblock {DINOv2}: Learning robust visual features without supervision.
\newblock \emph{Transactions on Machine Learning Research}, 2024.

\bibitem[Page(1954)]{page1954cusum}
Page, E.~S.
\newblock Continuous inspection schemes.
\newblock \emph{Biometrika}, 41\penalty0 (1--2):\penalty0 100--115, 1954.
\newblock \doi{10.1093/biomet/41.1-2.100}.

\bibitem[Rothenberg(1971)]{rothenberg1971}
Rothenberg, T.~J.
\newblock Identification in parametric models.
\newblock \emph{Econometrica}, 39\penalty0 (3):\penalty0 577--591, 1971.
\newblock \doi{10.2307/1913267}.

\bibitem[Rubinstein(1999)]{cem}
Rubinstein, R.
\newblock The cross-entropy method for combinatorial and continuous
  optimization.
\newblock \emph{Methodology and Computing in Applied Probability}, 1\penalty0
  (2):\penalty0 127--190, 1999.
\newblock \doi{10.1023/A:1010091220143}.

\bibitem[Sobal et~al.(2025)Sobal, Zhang, Cho, Balestriero, Rudner, and
  LeCun]{pldm}
Sobal, V., Zhang, W., Cho, K., Balestriero, R., Rudner, T. G.~J., and LeCun, Y.
\newblock Learning from reward-free offline data: A case for planning with
  latent dynamics models.
\newblock In \emph{Advances in Neural Information Processing Systems},
  volume~38, 2025.

\bibitem[Soni et~al.(2026)Soni, Sehgal, Dave, and Rueckert]{sandwich}
Soni, K., Sehgal, A., Dave, V., and Rueckert, E.
\newblock {Sandwich-Residuals}: Parameter-efficient test-time adaptation of
  world models.
\newblock \emph{arXiv preprint arXiv:2609.21740}, 2026.

\bibitem[Sun et~al.(2020)Sun, Wang, Liu, Miller, Efros, and Hardt]{sun2020ttt}
Sun, Y., Wang, X., Liu, Z., Miller, J., Efros, A., and Hardt, M.
\newblock Test-time training with self-supervision for generalization under
  distribution shifts.
\newblock In \emph{Proceedings of the 37th International Conference on Machine
  Learning}, volume 119 of \emph{Proceedings of Machine Learning Research},
  pp.\  9229--9248. PMLR, 2020.

\bibitem[Tassa et~al.(2018)Tassa, Doron, Muldal, Erez, Li, de~Las~Casas,
  Budden, Abdolmaleki, Merel, Lefrancq, Lillicrap, and Riedmiller]{dmc}
Tassa, Y., Doron, Y., Muldal, A., Erez, T., Li, Y., de~Las~Casas, D., Budden,
  D., Abdolmaleki, A., Merel, J., Lefrancq, A., Lillicrap, T., and Riedmiller,
  M.
\newblock {DeepMind} control suite.
\newblock \emph{arXiv preprint arXiv:1801.00690}, 2018.

\bibitem[Terver et~al.(2025)Terver, Yang, Ponce, Bardes, and LeCun]{jepawms}
Terver, B., Yang, T.-Y., Ponce, J., Bardes, A., and LeCun, Y.
\newblock What drives success in physical planning with joint-embedding
  predictive world models?
\newblock \emph{arXiv preprint arXiv:2512.24497}, 2025.
\newblock Accepted at Transactions on Machine Learning Research (arXiv comment,
  version 3).

\bibitem[Wang et~al.(2026{\natexlab{a}})Wang, Shi, Fei, Fu, Ji, Gong, and
  Qiu]{icwm}
Wang, S., Shi, J., Fei, S., Fu, Z., Ji, L., Gong, J., and Qiu, X.
\newblock In-context world modeling for robotic control.
\newblock \emph{arXiv preprint arXiv:2606.26025}, 2026{\natexlab{a}}.

\bibitem[Wang et~al.(2026{\natexlab{b}})Wang, Bounou, LeCun, and Ren]{adajepa}
Wang, Y., Bounou, O., LeCun, Y., and Ren, M.
\newblock {AdaJEPA}: An adaptive latent world model.
\newblock \emph{arXiv preprint arXiv:2606.32026}, 2026{\natexlab{b}}.

\bibitem[Wang et~al.(2022)Wang, Chen, and Dong]{llirl}
Wang, Z., Chen, C., and Dong, D.
\newblock Lifelong incremental reinforcement learning with online {Bayesian}
  inference.
\newblock \emph{IEEE Transactions on Neural Networks and Learning Systems},
  33\penalty0 (8):\penalty0 4003--4016, 2022.
\newblock \doi{10.1109/TNNLS.2021.3055499}.

\bibitem[Wolpert \& Kawato(1998)Wolpert and Kawato]{mosaic}
Wolpert, D.~M. and Kawato, M.
\newblock Multiple paired forward and inverse models for motor control.
\newblock \emph{Neural Networks}, 11\penalty0 (7--8):\penalty0 1317--1329,
  1998.
\newblock \doi{10.1016/S0893-6080(98)00066-5}.

\bibitem[Xu et~al.(2020)Xu, Ding, Zhu, Liu, Chen, and Zhao]{xu2020gpmixture}
Xu, M., Ding, W., Zhu, J., Liu, Z., Chen, B., and Zhao, D.
\newblock Task-agnostic online reinforcement learning with an infinite mixture
  of {Gaussian} processes.
\newblock In \emph{Advances in Neural Information Processing Systems},
  volume~33, pp.\  6429--6440, 2020.

\bibitem[Yang et~al.(2026)Yang, Han, Ozyurt, Yang, Yang, Cao, Zhen, Du, and
  Gan]{syncworld}
Yang, Y., Han, Z., Ozyurt, F., Yang, Z., Yang, H., Cao, J., Zhen, H., Du, Y.,
  and Gan, C.
\newblock {SyncWorld}: Visual calibration enables world models as zero-shot
  simulators.
\newblock \emph{arXiv preprint arXiv:2609.09155}, 2026.

\bibitem[Yu(2026)]{spikeworld}
Yu, Z.
\newblock {SpikeWorld}: Fast-state adaptation for frozen spiking world models.
\newblock \emph{arXiv preprint arXiv:2608.07712}, 2026.

\bibitem[Zhang et~al.(2023)Zhang, Becker-Ehmck, van~der Smagt, and Karl]{aime}
Zhang, X., Becker-Ehmck, P., van~der Smagt, P., and Karl, M.
\newblock Action inference by maximising evidence: Zero-shot imitation from
  observation with world models.
\newblock In \emph{Advances in Neural Information Processing Systems},
  volume~36, pp.\  46284--46303, 2023.
\newblock \doi{10.52202/075280-2007}.

\bibitem[Zhang et~al.(2026)Zhang, Ye, Agarwal, Nourkhiz~Mahjoub, Moradi~Pari,
  Khashabi, Shu, and Tadiparthi]{jepattt}
Zhang, Z., Ye, S., Agarwal, N., Nourkhiz~Mahjoub, H., Moradi~Pari, E.,
  Khashabi, D., Shu, T., and Tadiparthi, V.
\newblock {JEPA-TTT}: Persistent test-time training of latent world models for
  planning under dynamics shifts.
\newblock \emph{arXiv preprint arXiv:2610.00722}, 2026.

\bibitem[Zholus et~al.(2026)Zholus, Beltran-Velez, Yuan, Chandar, Nagarajan,
  Severo, Sinha, Drozdzal, Romero~Soriano, Bohg, Ballas, and Assran]{robojepa}
Zholus, A., Beltran-Velez, N., Yuan, J., Chandar, S., Nagarajan, T., Severo,
  D., Sinha, K., Drozdzal, M., Romero~Soriano, A., Bohg, J., Ballas, N., and
  Assran, M.
\newblock {RoboJEPA}: Scaling robotic latent world models.
\newblock \emph{arXiv preprint arXiv:2610.10515}, 2026.

\bibitem[Zhou et~al.(2025)Zhou, Pan, LeCun, and Pinto]{dinowm}
Zhou, G., Pan, H., LeCun, Y., and Pinto, L.
\newblock {DINO-WM}: World models on pre-trained visual features enable
  zero-shot planning.
\newblock In \emph{Proceedings of the 42nd International Conference on Machine
  Learning}, volume 267 of \emph{Proceedings of Machine Learning Research},
  pp.\  79115--79135. PMLR, 2025.

\end{thebibliography}
\end{document}